\pdfoutput=1
\documentclass{article}

\usepackage[preprint]{neurips_2020}

\usepackage{natbib}

\usepackage[utf8]{inputenc} 
\usepackage[T1]{fontenc}    
\usepackage{hyperref}       
\usepackage{url}            
\usepackage{booktabs}       
\usepackage{amsfonts}       
\usepackage{nicefrac}       
\usepackage{microtype}      
\usepackage{amssymb}
\usepackage{xcolor}
\usepackage{graphicx}
\usepackage{subfig}
\usepackage{wrapfig}

\title{Asynchronous Federated Learning with Reduced Number of Rounds and with Differential Privacy from Less Aggregated Gaussian Noise}

 \author{Marten van Dijk$^{1,2}$, Nhuong V. Nguyen$^{3}$, Toan N. Nguyen$^{3}$, \\ \textbf{Lam M. Nguyen}$^{4}$\textbf{,} \textbf{Quoc Tran-Dinh}$^{5}$\textbf{,}  \textbf{Phuong Ha Nguyen}$^{1}$ \\
$^{1}$ Department of Electrical and Computer Engineering, University of Connecticut, USA\\
$^{2}$ CWI Amsterdam, The Netherlands\\
$^{3}$ Department of Computer Science and Engineering, University of Connecticut, USA \\
$^{4}$ IBM Research, Thomas J. Watson Research Center, Yorktown Heights, USA\\
$^{5}$ Department of Statistics and Operations Research, \\
The University of North Carolina at Chapel Hill, Chapel Hill, NC, USA. \\
\texttt{\{marten.van$\_$dijk, nhuong.nguyen\}@uconn.edu}, \texttt{nntoan2211@gmail.com},  \\
\texttt{LamNguyen.MLTD@ibm.com}, \texttt{quoctd@email.unc.edu}, \texttt{phuongha.ntu@gmail.com}}
  
\begin{document}

\maketitle

\begin{abstract}
The feasibility of federated learning is highly constrained by the server-clients infrastructure in terms of network communication. Most newly launched  smartphones and IoT devices are equipped with GPUs or sufficient computing hardware to run powerful AI models. However, in case of the original synchronous federated learning, client devices suffer waiting times and regular communication between clients and server is required. This implies more sensitivity to local model training times and irregular or missed updates, hence, less or limited scalability to large numbers of clients  and convergence rates measured in real time will suffer. We propose a new algorithm for asynchronous federated learning which eliminates waiting times and reduces overall network communication - we provide rigorous theoretical analysis for strongly convex objective functions and provide simulation results. By adding Gaussian noise we show how our algorithm can be made differentially private -- new theorems show how the aggregated added Gaussian noise is significantly reduced.
\end{abstract}

\section{Introduction}

Federated Learning (FL) is a distributed machine learning approach which enables training on a large corpus of decentralized data located on devices like mobile phones or IoT devices. Federated learning brings the concept of ``bringing the code to the data, instead of the data to the code" \citep{bonawitz2019towards}. Google \citep{konevcny2016federated} demonstrated FL for the  first time at a large scale  when they conducted experiments of training a global model across all mobile devices via the Google Keyboard Android application \citep{GoogleAIBlog}.


FL as originally introduced and described in \citep{jianmin,mcmahan} operates as follows:
The participants in FL are a server containing a \textit{global} model and clients $c\in \{1,\ldots, n\}$
having their own data sets ${\cal D}_c$ that are not shared with each other. 
There are many rounds for training a global model $w$. At the $i$-th round, the server propagates the currently computed global model (weight vector) $w_i$ to each client.
Each 
client $c$ takes $w_i$ and performs training/learning based on a subset of samples chosen from its own dataset.  For example, clients use  (mini batch) Stochastic Gradient Descent (SGD)~\citep{robbins1951stochastic} for training and learning. 
This  creates a new local model $w_{c,i}$ based on $w_i$ which is specific to client $c$.
Each client transmits its local model $w_{c,i}$  to the server. A new global model $w_{i+1}$ is then created by the server as an average 
$w_{i+1}=\frac{1}{n}\sum_{c=1..n}  w_{c,i}$ 
of the local models once these  are received. 


Original FL requires synchrony between the server and clients.  
It requires that each client \textit{sends a full model back to the server} in each round and \textit{each client needs to wait for the next computation round}. For large and complicated models, this becomes a main bottleneck  due to the asymmetric property of internet connection and the different computation power of devices \citep{yang,konevcny2016federated,luping,kevin}.

Even though FL has been widely accepted in the machine learning community, there are {\em no convergence proofs} for FL that study how round step sizes, round sample set sizes, and convergence rate relate. Due to lack of convergence analysis, the users and server may not know the most efficient setup for FL. 
How many training data samples should be sampled  at the clients in each round? What are the most optimal step size schemes for SGD? 
And, very important for scalability of FL, how can we minimize network communication (in particular, reduce the number of rounds)?  How does all of this affect our privacy budget when adding differential privacy?


\vspace{3pt}
\noindent
{\bf Challenge.} We recognize the need for asynchronous federated learning for which we can prove convergence results in order to receive guidance to how to optimally set parameters (in terms of number of samples per round and learning rates)  in order to reduce amortized network communication and reduce aggregated added Gaussian noise for differential privacy.

Our approach solves this question for training a global model based on a strongly convex objective function. For diminishing learning rate, we base our analysis on 
%
\citep{nguyen2018sgd,nguyen2018new} which introduced an asynchronous SGD framework  which includes Hogwild! \citep{Hogwild} and for which tight bounds on the convergence rate for strongly convex problems was proven. 
This framework allows analysis of a  general SGD recurrence under inconsistent  weight vector reads from and weight vector writes to shared memory. The framework allows out of order reads and writes within a certain window of time, which we call permissible delay, measured in number of iterations.
By casting asynchronous federated learning in this framework we are able to adopt its theoretical analysis.

%
The main insight is that the asynchronous SGD framework can resist much larger delays than the natural delays caused by the network communication infrastructure. This means that the {\em asynchronous} FL algorithm itself is allowed to introduce much more asynchronous behavior  by design  -- this is exploited by having 
the number of SGD iterations performed locally at each client 
increase from round to round and this {\em reduces the amount of network communication}. This reduction also turns out to {\em reduce the aggregated added  Gaussian noise for differential privacy} -- even if per round the Gaussian noise may need to be larger.


Related work and discussion on  horizontal FL is in Supplemental Material \ref{app-alg}.
Our new asynchronous distributed SGD has the following novelties/contributions:
    \textbf{Asynchronous communication and flexible learning at clients:} 
     The clients and server can work together to create a global model 
     in asynchronous fashion, see Theorem \ref{thmalg}. 
      We assume that 
messages/packets never drop; they will be resend but can arrive out of order. We are robust against this kind of asynchronous behavior.
    The clients can perform asynchronous SGD (i.e., Hogwild!) themselves to locally optimize the power consumption and required computational resources.
    
    \textbf{Rigorous convergence rate analysis:} We create a mathematical framework to describe our asynchronous FL. Given the delay function, we provide general recipes for constructing increasing sample size sequences and diminishing round step size sequences. For strong convex problems, we show that this attains the best possible convergence rate within a constant factor ($\leq 16\cdot 36^2$), see Theorem \ref{thmsc2}. 

    \textbf{Reduced communication cost:} 
    The more gradient computations $K$ are needed for the desired accuracy, the more we reduce communication rounds $T$ and overall network communication between the clients and the server relative to original FL. For strong convex problems  $T\sim \sqrt{K}$ (up to a factor) -- as opposed to $K$ for original asynchronous FL which uses a constant sample size sequence.
    
    \textbf{Reduced aggregated added Gaussian noise for differential privacy:} We generalize the Gaussian differential privacy framework of \citep{abadi2016deep} by proving a relationship between its constants, see Theorem \ref{thm1}. This allows us to analyse, given $K$ and a sample size sequence, which Gaussian noise level $\sigma$ to use during each local round of computation. This proves that, given a privacy budget, the aggregated added noise over all local rounds reduces as a function of $K$ when sample size sequences increase more, see Theorem \ref{thmmain} (which also describes a phase transition between two ranges of $K$ relative to the size of the data set). This positively correlates with the accuracy of the local and global models.

From our theory 
we expect to see that in order to bootstrap convergence it is best  to start with larger step sizes and start with  rounds of small size (measured in the number of local SGD updates within a round)  after which these rounds should start  increasing in  size and should start using smaller and smaller step sizes (learning rate) for best performance (in terms of minimizing communication and minimizing aggregated added differential privacy noise).
Experiments 
confirm our expectations. 

\section{Asynchronous distributed SGD}

The optimization problem for training many Machine Learning (ML) models using a training set $\{\xi_i\}_{i=1}^m$ of $m$ samples can be formulated as a finite-sum minimization problem as follows
$$ 
\min_{w \in \mathbb{R}^d} \left\{ F(w) = \frac{1}{m}
\sum_{i=1}^m f(w; \xi_i) \right\}.
$$
The objective is to minimize a loss function with respect to model parameters $w$. This problem is known as empirical risk minimization and it covers a wide range of convex and non-convex problems from the ML domain, including, but not limited to, logistic regression, multi-kernel learning, conditional random fields and neural networks.
We are interested in solving the following more general stochastic optimization problem with respect to some distribution $\mathcal{D}$:
\begin{align}
\min_{w \in \mathbb{R}^d} \left\{ F(w) = \mathbb{E}_{\xi \sim \mathcal{D}} [ f(w;\xi) ] \right\},  \label{eqObj}  
\end{align}
where $F$ has a Lipschitz continuous gradient and $f$ is bounded from below for every $\xi$.

Our approach is based on the Hogwild!~\citep{nguyen2018sgd} recursion 
\begin{equation}
 w_{t+1} = w_t - \eta_t  \nabla f(\hat{w}_t;\xi_t),\label{eqwM2a}
 \end{equation}
 where $\hat{w}_t$ represents the vector used in computing the gradient $\nabla f(\hat{w}_t;\xi_t)$ and whose vector entries have been read (one by one)  from  an aggregate of a mix of  previous updates that led to $w_{j}$, $j\leq t$.
 In a single-thread setting where updates are done in a fully consistent way, i.e. $\hat{w}_t=w_t$, yields SGD with diminishing step sizes $\{\eta_t\}$. 
Recursion (\ref{eqwM2a}) models asynchronous SGD. We define the amount of asynchronous behavior by a function $\tau(t)$:

\begin{defn}
We say that the sequence $\{w_t\}$  is {\em consistent with delay function $\tau$}  
if, for all $t$, vector $\hat{w}_t$ includes the aggregate 
of the updates up to and including those made during the $(t-\tau(t))$-th iteration\footnote{ (\ref{eqwM2a}) defines the $(t+1)$-th iteration, where $\eta_t\nabla f(\hat{w}_t;\xi_t)$ represents the $(t+1)$-th update.}, i.e., $\hat{w}_t = w_0 - \sum_{j\in {\cal U}} \eta_j \nabla f(\hat{w}_j;\xi_j)$ for some ${\cal U}$ with $\{0,1,\ldots, t-\tau(t)-1\}\subseteq {\cal U}$.
\end{defn}

It turns out that for strong convex problems an optimal convergence rate for diminishing step sizes can be achieved for $\tau(t)$ as large as $\approx \sqrt{t/\ln t}$ (for a precise formulation see
Supplemental Material \ref{app-sc-dim}).
In order words, the recursion can resist large amounts of asynchronous behavior. In practice such amounts of asynchronous behavior will not occur. Our key idea is to  design {\em a distributed SGD algorithm which introduces additional asynchronous behavior} (because we can resist it anyway) by having the algorithm communicate less with the central server. This has as main advantage less communication and when combined with Differential Privacy (DP) less aggregated added DP noise is needed (leading to faster convergence to a model with desired accuracy). For the same DP guarantees and the same number of iterations (grad computations) we obtain better accuracy and have less communication.

\subsection{Asynchronous solution}

\begin{algorithm}[!t]
\caption{Client$_c$ -- Local Model with Differential Privacy}
\label{alg:DP}

\begin{algorithmic}[1]
\Procedure{Setup}{$n$}: Initialize  increasing sample size sequences $\{s_i\}_{i\geq 0}$ and $\{s_{i,c}\approx p_cs_i\}_{i\geq 0}$ for each client $c\in \{1,\ldots, n\}$, where $p_c$ scales the importance of client $c$. Initialize diminishing round step sizes $\{\bar{\eta}_i\}_{i\geq 0}$, a permissible delay function $\tau(\cdot)$ with $t-\tau(t)$ increasing in $t$, and a default global model for each client to start with.
\EndProcedure   
\State

\Procedure{ISRReceive}{$\hat{v},k$}: This Interrupt Service Routine is called whenever a new broadcast global model $\hat{v}$ is received from the server. Once received the client's local model $\hat{w}$ is replaced with $\hat{v}$
from which the latest updates $\bar{\eta}_i \cdot U$ are subtracted. The server broadcasts a new global model once the updates for a next, say $k$th, sample round from {\em all} clients have been received. This means that the received $\hat{v}$ includes all the updates up to and including round $k$.
\EndProcedure   

\State

\Procedure{MainClientDP}{$\mathcal{D}_c$} 
\State $i=0$, $\hat{w}=\hat{w}_{c,0,0}$ 
 
\While{\textbf{True}}
    \State $h=0$, $U=0$ 
    \While{$h< s_{i,c}$}
    
              \State $t_{glob}= s_0+\ldots+s_{i} - (s_{i,c}-h)-1$
       \State $t_{delay} = s_{k}+\ldots+s_{i} - (s_{i,c}-h) $
       \State {\bf while} 
       $\tau(t_{glob}) = t_{delay}$ 
       {\bf do} nothing {\bf end while}

        \State Sample uniformly at random $\xi$ from $\mathcal{D}_c$ 
        
        \State $g = \nabla f(\hat{w}, \xi)$ 
        
        \State $g = g / \max \{1, \|g\|/C\ \}$ \Comment{NEW: Clip gradient}
        
        \State ${U} = {U} + g$  
        
        \State Update model $w = \hat{w} - \bar{\eta}_{i} \cdot g$ \Comment{$w$ represents $ w_{c,i,h+1}$} 
         \State Update model $\hat{w} = w$ 


        \State $h$++
    \EndWhile
    
    \State $n\leftarrow N(0,C^2 \sigma_i^2\textbf{I})$ \Comment{NEW: Draw batch noise}
    \State $\hat{w}= \hat{w}+\bar{\eta}_i\cdot n, U=U+n$ \Comment{NEW: Inject batch noise} 
    
    \State Send $(i,c, U)$ to the Server. 
    \State $i$++
\EndWhile
   
\EndProcedure


\end{algorithmic}
\end{algorithm}

Clients update their local models according to
Algorithm \ref{alg:DP}, where we discard lines 17, 23, and 24 indicated by NEW; these additions are analysed in  the next section.
Supplemental Material \ref{app-alg} has all the detailed pseudo code with annotated invariants.



The clients in the distributed computation  apply recursion (\ref{eqwM2a}) in lines 16, 19, 20. We want to label each recursion with an iteration count $t$; this can then be used to compute with which delay function the labeled sequence $\{w_t \}$ is consistent.
In  order to find an ordering based on $t$ we define in Supplemental Material \ref{appmain} a mapping $\rho$ from the annotated labels $(c,i,h)$ in \Call{MainClient}{} to $t$ and use this to prove:

\begin{thm} \label{thmalg}
Our setup, client, and server algorithms produce a sequence $\{w_t\}$ according to recursion (\ref{eqwM2a}) where $\{\xi_t\}$ are selected from distribution ${\cal D}=\sum_{c=1}^n p_c {\cal D}_c$. Sequence $\{w_t\}$ is consistent with delay function $\tau$ as defined in \Call{Setup}{}.
\end{thm}





We assume that 
messages/packets never drop. They will be resent 
but can arrive out of order. We are robust against this kind of asynchronous behavior:
The amount of asynchronous behavior is limited by $\tau(\cdot)$; when the delay is getting too large, then the client enters a wait loop which terminates only when \Call{ISRReceive}{} receives a more recent global model $\hat{v}$ with higher $k$ (making $t_{delay}$ smaller). Since $\tau(t)$ increases in $t$ and is much larger than the delays caused by network latency and retransmission of dropped packets, asynchronous behavior due to such effects will not cause clients to get stuck in a wait loop. We assume different clients have approximately the same speed of computation which implies that this will not cause fast clients having to wait for long bursts of time.\footnote{When entering a wait loop, the client's operating system should context switch out and resume other computation. 
If a client $c$ is an outlier with slow computation speed, then we can adjust $p_c$ to be smaller in order to have its minibatch/sample size $s_{i,c}$ be proportionally smaller; this will change distribution ${\cal D}$ and therefore change objective function (\ref{eqObj}).
} 

We exploit the algorithm's resistance against delays $\tau(t)$ by using increasing sample size sequences $\{s_{i,c}\}$. Since the server only broadcasts when all clients have communicated their updates for a 'round' $k$, increasing sample sizes implies that $t_{delay}$ can get closer to $\tau(t_{glob})$. So, sample size sequences should not increase too much: We require the property that there exists a threshold $d$ such that for all $i\geq d+1$,
\vspace{-.2cm}
\begin{equation} \tau(\sum_{j=0}^i s_j)\geq \sum_{j=i-d}^i s_j. \label{eqtausample}
\end{equation}
In Supplemental Material \ref{appinv} we show that this allows us to replace  condition $\tau(t_{glob})=t_{delay}$ of the wait loop by 
$i = k + d$ when $i\geq d+1$  while still guaranteeing $t_{delay}\leq \tau(t_{glob})$ as an invariant. In practice, since sample sizes increase, we only need to require (\ref{eqtausample}) for $d=1$ in order to resist asynchronous behavior due to network latency.







We remark that our asynchronous distributed SGD is compatible with the more general recursion mentioned in \citep{nguyen2018sgd,nguyen2018new} and explained in Supplemental Material \ref{app-rec}. In this recursion each client can apply a "mask" which indicates the entries of the local model that will be considered. This allows each client to only transmit the local model entries corresponding to the mask and this will reduce communication per round.

Given a fixed budget/number of grad computations $K$ which each client needs to perform, an increasing sample size sequence $\{s_i\}$ reduces the number $T$ of communication rounds. This is an advantage as long as convergence to an accurate solution is guaranteed, that is, $\{s_i\}$ has to satisfy (\ref{eqtausample}). 


\subsection{Increasing sample size sequences}

Supplemental Material \ref{appinc} explains how a general formula for function $\tau$ translates into an increasing sample size sequence $\{s_i\}$ that satisfies (\ref{eqtausample}).
As soon as we have selected an increasing sample size sequence based on $\tau(\cdot)$, Supplemental Material \ref{appstep} explains how we can translate the diminishing step size sequence $\{\eta_t\}$ to a diminishing round step size sequence $\{\bar{\eta}_i\}$ that only diminish with every mini batch $s_i$. In Supplemental Material \ref{app-sc-dim} (with definitions for smooth, convex, and strong convex) we apply these techniques to existing theory from \citep{nguyen2018sgd,nguyen2018new,nguyen2019tight} for strong convex problems:\footnote{We remark that this theorem also holds for our algorithm where the clients compute batch SGD for the sample set and it can be adapted for the more general recursion with masks as explained in Supplemental Material \ref{app-rec} (with "$D>1$").}

\begin{thm} \label{thmsc2}
Let $f$ be $L$-smooth, convex, and let the objective function $F(w)=\mathbb{E}_{\xi\sim {\cal D}}[f(w;\xi)]$ be $\mu$-strongly convex with finite $N = 2 \mathbb{E}[ \|\nabla f(w_{*}; \xi)\|^2 ]$ where $w_{*} = \arg \min_w F(w)$.
We have  recipes  for (a) constructing, given parameter $d$,  increasing sample size sequences $\{s_i\}$ with $s_i = \Theta(\frac{i}{\ln i})$ which satisfy (\ref{eqtausample}),
and for (b)
constructing a corresponding  diminishing round step size sequence $\{\bar{\eta}_i\}$ with $\bar{\eta}_i= O(\frac{\ln i}{i^2})$ such that for large enough $t$
the convergence rate 
$$\mathbb{E}[\|w_{t} - w_* \|^2]\leq \frac{8\cdot 36^2 \cdot N}{\mu^2} \frac{1}{t}(1- O(1/t)).$$
By using the lower bound from \citep{nguyen2019tight} this is optimal up to a constant factor $\leq 16\cdot 36^2$ (independent of any parameters like $L$, $\mu$, sparsity, or dimension of the model).
\end{thm}


For a fixed number of grad computations $K$, the number $T$ of communication rounds satisfies $K=\sum_{j=0}^T s_j$. When forgetting the $\ln i$ component, this makes $T$ proportional to $\sqrt{K}$ -- rather than proportional to $K$ for original FL which uses a constant sample size sequence. This gives a significant reduction in communication rounds and network communication overall. 

We do not know whether the theory for $\tau(\cdot)$ for strong convex problems gives a tight bound and for this reason we also experiment with linear $s_i =\Theta(i)$ in the strongly convex case. 
For $s_i=\Theta(i)$ we have $\bar{\eta}_i= O(i^{-2})$.
As one benchmark we compare to using a constant step size $\eta=\bar{\eta}_i$. 
Supplemental Material \ref{appconst} analyses this case and shows how to choose the constant sample size $s=s_i$ (as large as $\frac{a}{ L\mu (d+1)}$ for a well defined constant $a$) in order to achieve the best convergence rate.


In Supplemental Material \ref{app-nonconvex} we discuss plain and non-convex problems. We argue  in both cases to choose  a  diminishing step size sequence of $O(t^{-1/2})$ and to experiment with different increasing sample size sequences to determine into what extent the asynchronous distributed SGD  is robust against delays. Since strong convex problems have best convergence and therefore best robustness against delays, we expect a suitable increasing sample size sequence $\Theta(i^p)$ for some $0\leq p\leq 1$.

\section{Differential privacy} \label{sec:DP}

Differential privacy \citep{dwork2006calibrating, dwork2011firm,dwork2014algorithmic,dwork2006our} defines privacy guarantees for algorithms on databases, in our case a client's sequence of mini-batch gradient computations on his/her training data set. The guarantee quantifies into what extent the output  of a client (the collection of updates communicated to the server) can be used to differentiate among two adjacent training data sets $d$ and $d'$ (i.e., where one set has one extra element compared to the other set).

\begin{defn} \label{defDP} A randomized mechanism ${\cal M}: D \rightarrow R$ is $(\epsilon, \delta)$-differentially private if for any adjacent $d$ and $d'$ in $D$ and for any subset $S\subseteq R$ of outputs,
$$ Pr[{\cal M}(d)\in S]\leq e^{\epsilon} Pr[{\cal M}(d')\in S] + \delta.$$
\end{defn}


Algorithm \ref{alg:DP} merges the Gaussian Differential Privacy (DP) algorithm of \citep{abadi2016deep} with \Call{MainClient}{}. Gaussian DP assumes that all gradients are bounded by some constant $C$ (this is needed in the DP proofs of \citep{abadi2016deep}). However, in general such a bound cannot be assumed (for example, the bounded gradient assumption  is in conflict with strong convexity \citep{nguyen2018sgd}). For this reason a constant $C$ is used to clip computed gradients. Once a batch $U$ of gradients is computed, Gaussian noise $n$ is added, after which the result is multiplied by the step size $\bar{\eta}_i$ (and added to the local model $\hat{w}$).

Gaussian DP introduces clipping and adds Gaussian noise. Experiments in \citep{abadi2016deep} show that such an adapted version of mini-batch SGD  still leads to acceptable convergence to acceptable accuracy. Note however, Theorem \ref{thmsc2} on the convergence rate for strong convex problems 
does not hold for the adapted version. Nevertheless experiments in Section \ref{sec:experiment} show that copying over the suggested step size sequences and sample size sequences lead to proper convergence and accuracy. 

%
%
%
%
%
%
%
%
%
%
%
%
%
%

In order to obtain privacy guarantees, we want to apply the DP theory from \citep{abadi2016deep} for diminishing step size sequence $\{\bar{\eta}_i\}$ and increasing sample size sequence $\{s_{i,c}\}$. 
Next theorems show how this is done (their proofs are in Supplemental Material \ref{appDP}) -- it generalizes the results of \citep{abadi2016deep} in a non-trivial way by analysing increasing sample size sequences, by making explicit the higher order error term in \citep{abadi2016deep},  and by providing a precise relationship among the constants used in the DP theory of \citep{abadi2016deep}.
We assume finite training data sets $|{\cal D}_c|=N_c$.

\begin{thm} \label{thm1}
%
We assume that $\sigma = \sigma_i$ for all rounds $i$. 
We assume there exists a constant $r_0<1/e$  such that $s_{i,c}/N_c\leq r_0/\sigma$. 
Then there exists\footnote{A precise characterization of $r$ is in (\ref{constr}) in Supplemental Material \ref{app-dp1}.} a constant $r$ as a function of $r_0$ and $\sigma$ used in the definition of $c(x)$ below.
For $j=1,2,3$ we define $\hat{S}_j$ (resembling an average over the sum of $j$-th powers of $s_{i,c}/N_c$) 
with related constants $\rho$ and $\hat{\rho}$:
$$\hat{S}_j = \frac{1}{T} \sum_{i=1}^T \frac{s_{i,c}^j }{N_c(N_c-s_{i,c})^{j-1}}, \ \
 \frac{\hat{S}_1\hat{S}_3}{\hat{S}_2^2}\leq \rho
 \mbox{ and }
 \frac{\hat{S}_1^2}{\hat{S}_2}\leq \hat{\rho} .$$ 

Let $\epsilon = c_1 T \hat{S}_1^2$.
Then, Algorithm \ref{alg:DP} is $(\epsilon, \delta)$-differentially private if
$$\sigma \geq \frac{2}{\sqrt{c_0}} \frac{\sqrt{ \hat{S}_2 T \ln (1/\delta) }}{\epsilon}  \mbox{ where } c_0=c(c_1)
\mbox{ with } c(x) = \min \{ \frac{\sqrt{2r\rho x+1} -1}{r \rho x}, \frac{2}{\hat{\rho} x}  
\}.
$$
\end{thm}

We notice that this generalizes Theorem 2 of \citep{abadi2016deep} where all $s_{i,c}/N_c=q$. 


By having characterized  $c_0=c(c_1)$ in the above theorem, we are able to analyse the effect of increasing sample sizes $s_{i,c}$ in our main DP theorem:

\begin{thm} \label{thmmain} Let $q_i=s_{i,c}/N_c
= q \cdot (i+m)^p$ with $m\geq 0$ and $p\geq 0$. Let $T$ be the total number of rounds and 
let $K$ be the total number of grad recursions/computations over all $T$ rounds.  Let $\epsilon>0$ and $\sigma\geq 1.137$. There exists\footnote{A precise description of how to calculate $r_0(\sigma)$  is in the proof of Theorem \ref{thm-mainp} in Supplemental Material \ref{app-main}.} 
a function $r_0(\sigma)$ with the following properties:
Define 
$$B = \frac{1}{1+p} \cdot (\frac{\sqrt{3}-1}{2}(2p+1))^{\frac{1+p}{1+2p}}, \ A = B\cdot (1+\frac{m}{T})^{\frac{(3+4p)(1+p)}{(1+2p)}}, \mbox{ and } D=\frac{(r_0(\sigma)/\sigma)^{\frac{1+p}{p}}}{p+1}(1+\frac{m}{T})^{1+p},
$$
$$ K^-= B \cdot \epsilon^{\frac{1+p}{1+2p}} q^{-\frac{1}{1+2p}} N_c, \ \
K^+ = A \cdot \epsilon^{\frac{1+p}{1+2p}} q^{-\frac{1}{1+2p}} N_c, \ \mbox{ and } K^*=D\cdot q^{-\frac{1}{p}} N_c.
$$
For fixed $p$, we have, for $K\leq K^*$, Algorithm \ref{alg:DP} is $(\epsilon,\delta)$-differential private if
\begin{eqnarray*}
\sigma &\geq& \sqrt{\frac{2\ln (1/\delta)}{\epsilon}} \frac{(1+\frac{m}{T})^{2+3p}}{\sqrt{1-r_0(\sigma)/\sigma}} \hspace{3.7cm} \mbox{ if } K\leq K^-, \\
\sigma &\geq& (\frac{K}{K^+})^\frac{1+2p}{2+2p}\cdot 1.21 \cdot  \sqrt{\frac{2\ln (1/\delta)}{\epsilon}}
\frac{(1+\frac{m}{T})^{2+3p}}{\sqrt{1-r_0(\sigma)/\sigma}}\hspace{1cm}
\mbox{ if } K^+\leq K.
\end{eqnarray*}
For fixed $K$, define $p^-$ as the solution of $p$ for which $K=K^-$ and define $p^+$ as the solution of $p$ for which $K=K^+$: We have $p^+-p^-=O(m/T)$. The smallest lower bound on $\sqrt{T}\sigma$ resulting from our analysis corresponding to the inequalities above varies with $((p+1)K/(q N_c))^{1/(1+2p)}$ as a function of $p\leq p^-$, after which $\sqrt{T}\sigma$ jumps by a factor\footnote{As an artifact of our bounding techniques.} $1.21$ when transitioning from $p=p^-$ to $p=p^+$, after which $\sqrt{T}\sigma$ varies with $\frac{p+1}{\sqrt{2p+1}}$ for $p\geq p^+$.
\end{thm}

We  consider $0\leq p\leq 1$, $\delta$ preferably smaller than $1/|N_c|$, and $\epsilon$ values like 2, 4, 8 as used in literature (e.g., \citep{abadi2016deep}); this makes the lower bound on $\sigma$ not 'too' large for the algorithm to converge to a solution with good/proper test accuracy.

The noise added {\em per round} is drawn from ${\cal N}(0,\sigma^2 C^2{\bf I})$. 
The theorem shows that for fixed $p$ and increasing $K$, 
the lower bound on $\sigma$ remains constant up to $K\leq K^-$ after which, for $K\geq K^+$,  the lower bound increases as $(K/K^+)^{(1+2p)/(2+2p)}$. Notice that such an increased lower bound must hold for the $\sigma$ chosen for {\em all} rounds. Notice that for larger $p$, $(K/K^+)^{(1+2p)/(2+2p)}$ increases more fast and this suggests that an increasing sample size sequence for larger $p$ may not be a good choice (if $K$ must be $>K^+$).

However, when looking at the {\em aggregated added noise over all rounds}, the picture is different because the number of rounds $T$ for fixed $K$ increases for smaller $p$: The aggregated added noise is distributed as ${\cal N}(0,T\sigma^2C^2\textbf{I})$. The theorem shows that the lower bound for $\sqrt{T}\sigma$ decreases to some value $B$ for increasing $p$ after which $B$ only slightly fluctuates with $p$. So, we can choose $\sigma$ such that the aggregated added noise reduces for increasing $p$. This shows that an increasing sample size sequence performs better than a constant sample size sequence ($p=0$) in this sense.

We conclude that the same round noise leads to the same DP guarantees for $p\leq p^-$. For increasing $p$ we have less round communication and the aggregated added noise reduces. For $p\geq p^+$ the aggregated added noise needs to remain constant in order to offer the same DP guarantees; this is achieved by increasing the round noise in exchange for larger sample sizes. 

\noindent
{\bf Parameter selection.} Theorem \ref{thmmain} can be used used as follows:
We assume we know the initial sample size $s_{0,c}$,  data set size $N_c$, value for $p$, and differential privacy budget ${\cal B}= \sqrt{2\ln (1/\delta)}/\sqrt{\epsilon}$ for some $\epsilon$ which we wish to achieve. We estimate a total number of grad computations $K$ needed for an anticipated $\sigma \geq {\cal B}$ based on the lower bound of $\sigma$ for $K\leq K^-$.
We apply Theorem \ref{thmmain} to find concrete parameter settings: We compute $r_0(\sigma)$. Next (case 1 -- aka $K\leq K^-$), we compute $q$ small enough such that $K\leq K^-$ and $K\leq K^*$. Given the initial sample size $s_{0,c}$ and data set size $N_c$ we compute $m$. This yields a formula for $q_i$. We compute $T$ (notice that $K/N_c = \sum_{i=0..T} q_i$) and check the value for $m/T$. Factor $m/T$ indicates how much alike the sample size sequence is to the constant sample size sequence; we will need to correct our calculations based on the calculated $m/T$ and possibly iterate. This leads to a lower bound on $\sigma$ from which we obtain an upper bound on ${\cal B}$; if acceptable, then we are done. If not, then we try a larger $\sigma$, recompute $r_0(\sigma)$, and again start with (case 1) as explained above. At the end we have a parameter setting for which we can compute the overall reduction in communication rounds and reduction in aggregated added DP noise -- this is compared to parameter settings found for other values of $p$ (in particular, $p=0$) that achieve ${\cal B}$. The sample size sequence is computed as $s_{i,c}=\lceil N_c q_i\rceil$. 

After all these investigations, we may also decide to (case 2 -- aka $K\geq K^+$) compute $q$ small enough such that $K\leq K^*$ and $K\leq k\cdot K^+$ for some small enough factor $k>1$. Based on $k$ and ${\cal B}$  we anticipate $\sigma$ based on the lower bound of $\sigma$ for $K\geq K^+$. As in case 1, we use $q$ to compute $m$ and $T$. We again check factor $m/T$, correct calculations, possibly iterate, and finally leading to a lower bound on $\sigma$ which gives an upper bound on ${\cal B}$. We may need to try a larger $\sigma$, recompute $r_0(\sigma)$ and repeat (case 2) until satisfied. Our general theory in Supplemental Material \ref{app-main} allows other larger settings for $r_0$ (other than $r_0=r_0(\sigma)$), especially in case 2, this may lead to improved results. 

Supplemental Material \ref{app:app} shows various examples. One example with $s_{0,c}=16$, $N_c=10,000$, $K=25,000$ (this assumes a `good' data set),  $\sigma=8$, and $\epsilon=1$ 
gives  for $s_{i,c}=16+\lceil 1.322\cdot i\rceil$ the values $T\approx 195$ and privacy budget ${\cal B}=5.78$ implying $\delta = 5.5\cdot 10^{-8}$. Compared to a constant sample size of 16 which has 1563 rounds, we have a reduction in communication of $195/1563=0.12$ (a factor 8 smaller) and the aggregated added noise reduces from $\sqrt{1563}\cdot 5.78=229$ to $\sqrt{195}\cdot 8=112$. This increasing sample size sequence is used in our DP experiments of the next section.

\section{Experiments}
\label{sec:experiment}

We summarize experimental results on strong convex problems. Supplemental Material \ref{app:experiment} shows additional results for plain and non-convex problems.

\begin{figure}[ht!]
  \centering
   \vspace{-.0cm}
  \subfloat[][Convergence rate with difference step sizes\\ (strong convex, phishing data set).]
  {\includegraphics[width=0.5\textwidth]{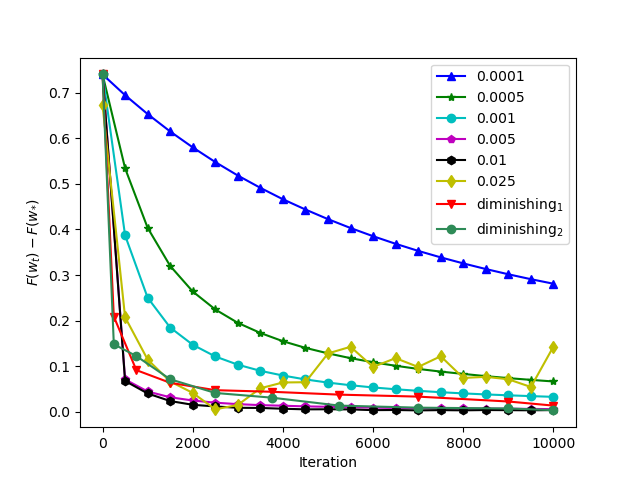}\label{fig:async_fl_main_strongly_convex}}
  \hfill
  \subfloat[][DP with asyn-FL, $K=25000, \sigma=8.0$\\ (strong convex, w8a data set).]
  {\includegraphics[width=0.5\textwidth]{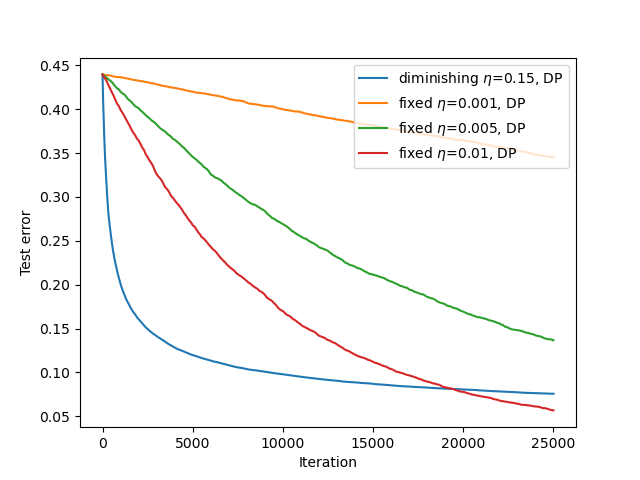}\label{fig:tbl_async_fl_main_convex_dp_w8a}}
  \caption{Asynchronous FL with (a) different step size schemes and (b) differential privacy.}
  \label{fig:asyn_fl_main_figure}
\end{figure}

\textbf{Asynchronous FL without differential privacy:} Figure~\ref{fig:async_fl_main_strongly_convex} shows that our proposed asynchronous FL with diminishing step sizes and increasing sample size sequence achieves the  same or better accuracy when compared to FL with constant step sizes and constant sample sizes.
The figure depicts a strong convex problem with diminishing step size scheme $\eta_t=\frac{\eta_0}{1 + \beta \cdot t}$ for an initial step size $\eta_{0}=0.1$ with a linearly increasing sample size sequence (see Supplemental Material~ \ref{subsec:samplesize_FL_simu} for details); diminishing$_1$ uses a more fine tuned  $\eta_t$ locally at the clients and diminishing$_2$ uses the transformation to round step sizes $\bar{\eta}_i$.
The number of communication rounds for constant step and sample sizes is $20$ rounds, while  the diminishing step size with increasing sample size setting only needs  $9$ communication rounds. 

\vspace{-0.5cm}


\textbf{Asynchronous FL with differential privacy:} 
In Figure~$\ref{fig:tbl_async_fl_main_convex_dp_w8a}$
we depict the performance of our diminishing step size scheme ($\eta_0=0.15$) with linearly increasing sample size sequence $s_{i,c}=16+\lceil 1.348\cdot i\rceil$ and with added DP noise ($\sigma=8$; see Supplemental Material~\ref{subsec:exp_asynFL_DP} for details).
Experiments show that increasing the sample sizes (up to linear for strong convex problems) without DP does not affect accuracy too much. Therefore we expect that with DP increasing sample size sequences improve accuracy over the constant sample size  sequence (due to less aggregated added noise while still achieving the same DP guarantees). This is confirmed in Figure~$\ref{fig:tbl_async_fl_main_convex_dp_w8a}$.


\begin{figure}[ht!]
\begin{center}
\includegraphics[width=0.6\textwidth]{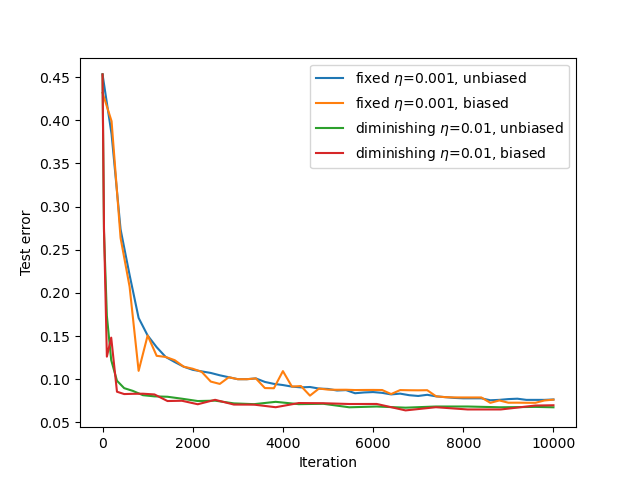}
\end{center}
\caption{Asynchronous FL with biased and unbiased dataset (strongly convex, MNIST subsets)}
\label{fig:asyn_fl_biased_strongly_convex_main_mnist}
\end{figure}

\textbf{Asynchronous FL with  biased data set:}  The goal of this experiment is to show that our FL framework can work well with biased data sets meaning that different clients use different distributions ${\cal D}_c$. We continue with the setting as mentioned above, now with $\eta_{0}=0.01$
(see Supplemental Material~\ref{subsec:dataset_FL_simu} for details).  Figure~$\ref{fig:asyn_fl_biased_strongly_convex_main_mnist}$ shows no significant difference when  clients run  biased or unbiased data sets. We conclude that our proposed FL framework can tolerate the issue of biased data sets, which is common in practice.

\section{Conclusion}
We have introduced asynchronous FL based on SGD with diminishing step size schemes and increasing sample size sequences. Our algorithm achieves a significant  reduction in communication between clients and server and  reduction in aggregated DP noise. We generalized Gaussian DP theory and proved optimal (up to a constant factor) convergence rate for strong convex problems. 

\bibliography{reference}

\begin{thebibliography}{37}
\providecommand{\natexlab}[1]{#1}
\providecommand{\url}[1]{\texttt{#1}}
\expandafter\ifx\csname urlstyle\endcsname\relax
  \providecommand{\doi}[1]{doi: #1}\else
  \providecommand{\doi}{doi: \begingroup \urlstyle{rm}\Url}\fi

\bibitem[Abadi et~al.(2016)Abadi, Chu, Goodfellow, McMahan, Mironov, Talwar,
  and Zhang]{abadi2016deep}
Martin Abadi, Andy Chu, Ian Goodfellow, H~Brendan McMahan, Ilya Mironov, Kunal
  Talwar, and Li~Zhang.
\newblock Deep learning with differential privacy.
\newblock In \emph{Proceedings of the 2016 ACM SIGSAC Conference on Computer
  and Communications Security}, pages 308--318. ACM, 2016.

\bibitem[Bonawitz et~al.(2019)Bonawitz, Eichner, Grieskamp, Huba, Ingerman,
  Ivanov, Kiddon, Konecny, Mazzocchi, McMahan, et~al.]{bonawitz2019towards}
Keith Bonawitz, Hubert Eichner, Wolfgang Grieskamp, Dzmitry Huba, Alex
  Ingerman, Vladimir Ivanov, Chloe Kiddon, Jakub Konecny, Stefano Mazzocchi,
  H~Brendan McMahan, et~al.
\newblock Towards federated learning at scale: System design.
\newblock \emph{arXiv preprint arXiv:1902.01046}, 2019.

\bibitem[Bottou et~al.(2018)Bottou, Curtis, and Nocedal]{BottouCN18}
L{\'{e}}on Bottou, Frank~E. Curtis, and Jorge Nocedal.
\newblock Optimization methods for large-scale machine learning.
\newblock \emph{{SIAM} Review}, 60\penalty0 (2):\penalty0 223--311, 2018.

\bibitem[Chee and Toulis(2018)]{CheeT18}
Jerry Chee and Panos Toulis.
\newblock Convergence diagnostics for stochastic gradient descent with constant
  learning rate.
\newblock In \emph{International Conference on Artificial Intelligence and
  Statistics, {AISTATS} 2018}, pages 1476--1485, 2018.

\bibitem[Chen et~al.(2016)Chen, Monga, Bengio, and Jozefowicz]{jianmin}
Jianmin Chen, Rajat Monga, Samy Bengio, and Rafal Jozefowicz.
\newblock Revisiting distributed synchronous sgd.
\newblock \emph{ICLR Workshop Track}, 2016.

\bibitem[Chen et~al.(2019{\natexlab{a}})Chen, Sun, and Jin]{yang}
Yang Chen, Xiaoyan Sun, and Yaochu Jin.
\newblock Communication-efficient federated deep learning with asynchronous
  model update and temporally weighted aggregation.
\newblock \emph{arXiv preprint}, 2019{\natexlab{a}}.
\newblock URL \url{https://arxiv.org/pdf/1903.07424.pdf}.

\bibitem[Chen et~al.(2019{\natexlab{b}})Chen, Sun, and Jin]{yang2019}
Yang Chen, Xiaoyan Sun, and Yaochu Jin.
\newblock Communication-efficient federated deep learning with asynchronous
  model update and temporally weighted aggregation.
\newblock \emph{arXiv preprint}, 2019{\natexlab{b}}.
\newblock URL \url{https://arxiv.org/pdf/1903.07424.pdf}.

\bibitem[Dwork(2011)]{dwork2011firm}
Cynthia Dwork.
\newblock A firm foundation for private data analysis.
\newblock \emph{Communications of the ACM}, 54\penalty0 (1):\penalty0 86--95,
  2011.

\bibitem[Dwork et~al.(2006{\natexlab{a}})Dwork, Kenthapadi, McSherry, Mironov,
  and Naor]{dwork2006our}
Cynthia Dwork, Krishnaram Kenthapadi, Frank McSherry, Ilya Mironov, and Moni
  Naor.
\newblock Our data, ourselves: Privacy via distributed noise generation.
\newblock In \emph{Annual International Conference on the Theory and
  Applications of Cryptographic Techniques}, pages 486--503. Springer,
  2006{\natexlab{a}}.

\bibitem[Dwork et~al.(2006{\natexlab{b}})Dwork, McSherry, Nissim, and
  Smith]{dwork2006calibrating}
Cynthia Dwork, Frank McSherry, Kobbi Nissim, and Adam Smith.
\newblock Calibrating noise to sensitivity in private data analysis.
\newblock In \emph{Theory of cryptography conference}, pages 265--284.
  Springer, 2006{\natexlab{b}}.

\bibitem[Dwork et~al.(2014)Dwork, Roth, et~al.]{dwork2014algorithmic}
Cynthia Dwork, Aaron Roth, et~al.
\newblock The algorithmic foundations of differential privacy.
\newblock \emph{Foundations and Trends{\textregistered} in Theoretical Computer
  Science}, 9\penalty0 (3--4):\penalty0 211--407, 2014.

\bibitem[Evans et~al.(2017)Evans, Kolesnikov, and Rosulek]{evans2017pragmatic}
David Evans, Vladimir Kolesnikov, and Mike Rosulek.
\newblock A pragmatic introduction to secure multi-party computation.
\newblock \emph{Foundations and Trends{\textregistered} in Privacy and
  Security}, 2\penalty0 (2-3), 2017.

\bibitem[Hsieh et~al.(2017)Hsieh, Harlap, Vijaykumar, Konomis, Ganger, Gibbons,
  and Mutlu]{kevin}
Kevin Hsieh, Aaron Harlap, Nandita Vijaykumar, Dimitris Konomis, Gregory~R.
  Ganger, Phillip~B. Gibbons, and Onur Mutlu.
\newblock Gaia: Geo-distributed machine learning approaching {LAN} speeds.
\newblock \emph{14th {USENIX} Symposium on Networked Systems Design and
  Implementation ({NSDI} 17).}, 2017.

\bibitem[Kone{\v{c}}n{\`y} et~al.(2016{\natexlab{a}})Kone{\v{c}}n{\`y},
  McMahan, Ramage, and Richt{\'a}rik]{konevcny2016federated}
Jakub Kone{\v{c}}n{\`y}, H~Brendan McMahan, Daniel Ramage, and Peter
  Richt{\'a}rik.
\newblock Federated optimization: Distributed machine learning for on-device
  intelligence.
\newblock \emph{arXiv preprint arXiv:1610.02527}, 2016{\natexlab{a}}.

\bibitem[Kone{\v{c}}n{\`y} et~al.(2016{\natexlab{b}})Kone{\v{c}}n{\`y},
  McMahan, Yu, Richt{\'a}rik, Suresh, and Bacon]{konevcny2016federated_com}
Jakub Kone{\v{c}}n{\`y}, H~Brendan McMahan, Felix~X Yu, Peter Richt{\'a}rik,
  Ananda~Theertha Suresh, and Dave Bacon.
\newblock Federated learning: Strategies for improving communication
  efficiency.
\newblock \emph{arXiv preprint arXiv:1610.05492}, 2016{\natexlab{b}}.

\bibitem[Leblond et~al.(2018)Leblond, Pedregosa, and
  Lacoste-Julien]{Leblond2018}
R{\'e}mi Leblond, Fabian Pedregosa, and Simon Lacoste-Julien.
\newblock Improved asynchronous parallel optimization analysis for stochastic
  incremental methods.
\newblock \emph{JMLR}, 19\penalty0 (1):\penalty0 3140--3207, 2018.

\bibitem[Li et~al.(2019{\natexlab{a}})Li, Sahu, Talwalkar, and
  Smith]{tianli2019federated}
Tian Li, Anit~Kumar Sahu, Ameet Talwalkar, and Virginia Smith.
\newblock Federated learning:challenges, methods, and future directions.
\newblock \emph{arXiv preprint arXiv:1908.07873}, 2019{\natexlab{a}}.

\bibitem[Li et~al.(2019{\natexlab{b}})Li, Sahu, Zaheer, Sanjabi, Talwalkar, and
  Smith]{tianli2019hetero}
Tian Li, Anit~Kumar Sahu, Manzil Zaheer, Maziar Sanjabi, Ameet Talwalkar, and
  Virginia Smith.
\newblock Federated optimization for heterogeneous networks.
\newblock \emph{arXiv preprint}, 2019{\natexlab{b}}.
\newblock URL \url{https://arxiv.org/pdf/1812.06127.pdf}.

\bibitem[Lian et~al.(2015)Lian, Huang, Li, and Liu]{lian2015asynchronous}
Xiangru Lian, Yijun Huang, Yuncheng Li, and Ji~Liu.
\newblock Asynchronous parallel stochastic gradient for nonconvex optimization.
\newblock In \emph{Advances in Neural Information Processing Systems}, pages
  2737--2745, 2015.

\bibitem[Lian et~al.(2017)Lian, Zhang, Zhang, and Liu]{lian2017asynchronous}
Xiangru Lian, Wei Zhang, Ce~Zhang, and Ji~Liu.
\newblock Asynchronous decentralized parallel stochastic gradient descent.
\newblock \emph{arXiv preprint arXiv:1710.06952}, 2017.

\bibitem[McMahan and Ramage(2017)]{GoogleAIBlog}
Brendan McMahan and Daniel Ramage.
\newblock Federated learning: Collaborative machine learning without
  centralized training data, 2017.
\newblock URL
  \url{https://ai.googleblog.com/2017/04/federated-learning-collaborative.html}.
\newblock Last accessed 09/24/2019.

\bibitem[McMahan et~al.(2016)McMahan, Moore, Ramage, and y~Arcas]{mcmahan}
H.~Brendan McMahan, Eider Moore, Daniel Ramage, and Blaise~Agüera y~Arcas.
\newblock Federated learning of deep networks using model averaging.
\newblock \emph{ICLR Workshop Track}, 2016.

\bibitem[Meng et~al.(2017)Meng, Chen, Yu, Wang, Ma, and
  Liu]{meng2017asynchronous}
Qi~Meng, Wei Chen, Jingcheng Yu, Taifeng Wang, Zhi-Ming Ma, and Tie-Yan Liu.
\newblock Asynchronous stochastic proximal optimization algorithms with
  variance reduction.
\newblock In \emph{Thirty-First AAAI Conference on Artificial Intelligence},
  2017.

\bibitem[Nguyen et~al.(2018)Nguyen, Nguyen, van Dijk, Richt{\'a}rik,
  Scheinberg, and Tak{\'a}{\v{c}}]{nguyen2018sgd}
Lam~M Nguyen, Phuong~Ha Nguyen, Marten van Dijk, Peter Richt{\'a}rik, Katya
  Scheinberg, and Martin Tak{\'a}{\v{c}}.
\newblock Sgd and hogwild! convergence without the bounded gradients
  assumption.
\newblock \emph{arXiv preprint arXiv:1802.03801}, 2018.

\bibitem[Nguyen et~al.(2019{\natexlab{a}})Nguyen, Nguyen, Richt{{\'a}}rik,
  Scheinberg, Tak{{\'a}}{\v{c}}, and van Dijk]{nguyen2018new}
Lam~M. Nguyen, Phuong~Ha Nguyen, Peter Richt{{\'a}}rik, Katya Scheinberg,
  Martin Tak{{\'a}}{\v{c}}, and Marten van Dijk.
\newblock New convergence aspects of stochastic gradient algorithms.
\newblock \emph{Journal of Machine Learning Research}, 20\penalty0
  (176):\penalty0 1--49, 2019{\natexlab{a}}.

\bibitem[Nguyen et~al.(2019{\natexlab{b}})Nguyen, Nguyen, and van
  Dijk]{nguyen2019tight}
P.~H. Nguyen, L.~M. Nguyen, and M.~van Dijk.
\newblock Tight dimension independent lower bound on the expected convergence
  rate for diminishing step sizes in {SGD}.
\newblock \emph{The 33th Annual Conference on Neural Information Processing
  Systems (NeurIPS 2019)}, 2019{\natexlab{b}}.

\bibitem[Phong et~al.(2018)Phong, Aono, Hayashi, Wang, and
  Moriai]{phong2018privacy}
Le~Trieu Phong, Yoshinori Aono, Takuya Hayashi, Lihua Wang, and Shiho Moriai.
\newblock Privacy-preserving deep learning via additively homomorphic
  encryption.
\newblock \emph{IEEE Transactions on Information Forensics and Security},
  13\penalty0 (5):\penalty0 1333--1345, 2018.

\bibitem[Recht et~al.(2011)Recht, Re, Wright, and Niu]{Hogwild}
Benjamin Recht, Christopher Re, Stephen Wright, and Feng Niu.
\newblock Hogwild: A lock-free approach to parallelizing stochastic gradient
  descent.
\newblock In \emph{Advances in neural information processing systems}, pages
  693--701, 2011.

\bibitem[Robbins and Monro(1951{\natexlab{a}})]{RM1951}
Herbert Robbins and Sutton Monro.
\newblock A stochastic approximation method.
\newblock \emph{The Annals of Mathematical Statistics}, 22\penalty0
  (3):\penalty0 400--407, 1951{\natexlab{a}}.

\bibitem[Robbins and Monro(1951{\natexlab{b}})]{robbins1951stochastic}
Herbert Robbins and Sutton Monro.
\newblock A stochastic approximation method.
\newblock \emph{The annals of mathematical statistics}, pages 400--407,
  1951{\natexlab{b}}.

\bibitem[Shi et~al.(2019)Shi, Wang, Zhao, Tang, Wang, Huang, and
  Chu]{shi2019distributed}
Shaohuai Shi, Qiang Wang, Kaiyong Zhao, Zhenheng Tang, Yuxin Wang, Xiang Huang,
  and Xiaowen Chu.
\newblock A distributed synchronous sgd algorithm with global top-$ k $
  sparsification for low bandwidth networks.
\newblock \emph{arXiv preprint arXiv:1901.04359}, 2019.

\bibitem[Stich(2018)]{stich2018local}
Sebastian~U Stich.
\newblock Local sgd converges fast and communicates little.
\newblock \emph{arXiv preprint arXiv:1805.09767}, 2018.

\bibitem[Van~Dijk et~al.(2019)Van~Dijk, Nguyen, Nguyen, and
  Phan]{van2019characterization}
Marten Van~Dijk, Lam Nguyen, Phuong~Ha Nguyen, and Dzung Phan.
\newblock Characterization of convex objective functions and optimal expected
  convergence rates for sgd.
\newblock In \emph{International Conference on Machine Learning}, pages
  6392--6400, 2019.

\bibitem[Wang et~al.(2019)Wang, Wang, and Li]{luping}
Luping Wang, Wei Wang, and Bo~Li.
\newblock Cmfl: Mitigating communication overhead for federated learning.
\newblock \emph{IEEE International Conference on Distributed Computing
  Systems.}, 2019.

\bibitem[Xie et~al.(2019)Xie, Koyejo, and Gupta]{cong2019}
Cong Xie, Sanmi Koyejo, and Indranil Gupta.
\newblock Asynchronous federated optimization.
\newblock \emph{arXiv preprint}, 2019.
\newblock URL \url{https://arxiv.org/pdf/1903.03934v1.pdf}.

\bibitem[Zheng et~al.(2017)Zheng, Meng, Wang, Chen, Yu, Ma, and
  Liu]{zheng2017asynchronous}
Shuxin Zheng, Qi~Meng, Taifeng Wang, Wei Chen, Nenghai Yu, Zhi-Ming Ma, and
  Tie-Yan Liu.
\newblock Asynchronous stochastic gradient descent with delay compensation.
\newblock In \emph{Proceedings of the 34th International Conference on Machine
  Learning-Volume 70}, pages 4120--4129. JMLR. org, 2017.

\bibitem[Zinkevich et~al.(2009)Zinkevich, Langford, and
  Smola]{zinkevich2009slow}
Martin Zinkevich, John Langford, and Alex~J Smola.
\newblock Slow learners are fast.
\newblock In \emph{Advances in neural information processing systems}, pages
  2331--2339, 2009.

\end{thebibliography}
\bibliographystyle{plainnat} 

\clearpage
\appendix

    

\vbox{%
\hsize\textwidth
\linewidth\hsize
\vskip 0.1in
\centering
{\LARGE\bf{Appendix} \par}
  \vskip 0.2in
\hrule height 1pt
\vskip 0.09in
}

\section{Algorithms} \label{app-alg}

\begin{algorithm}[!h]
\caption{Initial Setup} 
\label{alg:setup}

\begin{algorithmic}[1]
\Procedure{Setup}{$n$}
    \State {\bf Initialize} global model $\hat{v}_0=\hat{w}_{c,0,0}$ for server and clients $c\in \{1,\ldots, n\}$
    \State {\bf Initialize} diminishing round step size sequence $\{\bar{\eta}_{i}\}_{i\geq 0}$
    \State {\bf Initialize} increasing sample size sequence $\{s_{i}\}_{i\geq 0}$
    \For{$i \geq 0$}
        \For{$t \in \{0,\ldots, s_i-1\}$}
            \State Assign $a(i,t)=c$ with probability $p_c$
        \EndFor
        \For{$c\in \{1,\ldots n\}$}
            \State $s_{i,c}= |\{ t \ : \ a(i,t)=c\}|$
        \EndFor
    \EndFor
    \State \Comment{$\{s_{i,c}\}_{i\geq 0}$ represents the sample size sequence for client $c$; notice that $\mathbb{E}[s_{i,c}]=p_c s_i$}
    \State {\bf Initialize} permissible delay function $\tau(\cdot)$ with $t-\tau(t)$ increasing in $t$
\EndProcedure
\end{algorithmic}
\end{algorithm}

\begin{algorithm}[!t]
\caption{Server -- Global Model}
\label{alg:server}
 
\begin{algorithmic}[1]
\Procedure{ISRReceive}{$message$} \Comment{Interrupt Service Routine}
\If{$message==(i,c,U)$ is from a client $c$} \Comment{$U$ represents $U_{i,c}$}
\State $Q.enqueue(message)$ 
 \Comment{Queue $Q$ maintains aggregate gradients not yet processed}
\EndIf
\EndProcedure
\State
\Procedure{MainServer}{}
\State $k = 0$ \Comment{Represents a broadcast counter}
\State $\hat{v}=\hat{v}_0$
\State Initialize $Q$ and $H$ to empty queues
\While{\textbf{True}}
       
    \If{$Q$ is not empty}
    \State {$(i,c,U) \leftarrow$ $Q.dequeue()$} \Comment{Receive $U_{i,c}$}
            \State $\hat{v} = \hat{v} - \bar{\eta}_{i} \cdot U$
            \State $H.enqueue((i,c))$  
            \If{$H$ has $(k,c)$ for all $c\in \{1,\ldots, n\}$}
                \State $H.dequeue((k,c))$ for all $c \in \{1,\ldots, n\}$ 
             \State $k$++
                \State Broadcast $(\hat{v}, k)$ to all clients \Comment{$\hat{v}=\hat{v}_k$}
            \EndIf
    \EndIf
    \State \Comment{{\em Invariant}: $\hat{v}= \hat{v}_0 -
    \sum_{i=0}^{k-1} \sum_{c=1}^n \bar{\eta}_i U_{i,c}
    - \sum_{(i,c)\in H} \bar{\eta}_i U_{i,c}$}
    \State \Comment{$\hat{v}_k$ 
    includes the aggregate of updates $\sum_{i=0}^{k-1} \sum_{c=1}^n \bar{\eta}_i U_{i,c}$}
    

\EndWhile
\EndProcedure

\end{algorithmic}
\end{algorithm} 

\begin{algorithm}[!t]
\caption{Client$_c$ -- Local Model}
\label{alg:client}

\begin{algorithmic}[1]

\Procedure{ISRReceive}{$message$} \Comment{Interrupt Service Routine}
\If{$message==(\hat{v}, k)$ comes from the server \& $k > kold$} 
\State \Comment{The client will only accept and use a more fresh global model}
\State Replace the variable $k$ as maintained by the client by that of  $message$

            \State $\hat{w} = \hat{v} - \bar{\eta}_i \cdot U$ \Comment{This represents $\hat{w}_{c,i,h}= \hat{v}_k - \bar{\eta}_i \cdot U_{i,c,h}$}
            

\State $kold=k$
\State $t_{delay} = s_{k}+\ldots+s_{i} - (s_{i,c}-h) $
\EndIf
\EndProcedure


\State
\Procedure{MainClient}{$\mathcal{D}_c$} \Comment{$\mathcal{D}_c$ represents the local training data for client $c$}
\State $i=0$, $\hat{w}=\hat{w}_{c,0,0}$ \Comment{Local round counter}
 
 
\While{\textbf{True}}
    \State $h=0$, $U=0$ \Comment{$U_{i,c,h}$ for $h=0$ equals the all-zero vector}
    \While{$h< s_{i,c}$}
       \State $t_{glob}= s_0+\ldots+s_{i} - (s_{i,c}-h)-1$
       \State $t_{delay} = s_{k}+\ldots+s_{i} - (s_{i,c}-h) $
 
        \While{$\tau(t_{glob}) \leq  t_{delay}$}
            \State only keep track the update of $t_{delay}$
            \Comment{{\em Invariant}: $t_{delay}\leq \tau(t_{glob})$}
        \EndWhile
        
        \State Sample uniformly at random $\xi$ from $\mathcal{D}_c$ \Comment{$\xi$ represents $\xi_{c,i,h}$}
        
        \State $g = \nabla f(\hat{w}, \xi)$ \Comment{$g$ represents $g_{c,i,h}= \nabla f(\hat{w}_{c,i,h}, \xi_{c,i,h}) $}
        
        \State ${U} = {U} + g$  \Comment{Represents $U_{i,c,h+1}=U_{i,c,h} + g_{c,i,h}$ implying $U_{i,c,h+1}= \sum_{j=0}^h g_{c,i,j}$}
        
        \State Update model $w = \hat{w} - \bar{\eta}_{i} \cdot g$ \Comment{This represents $ w_{c,i,h+1} = \hat{w}_{c,i,h} - \bar{\eta}_i g_{c,i,h}$}
         \State Update model $\hat{w} = w$ \Comment{This represents $\hat{w}_{c,i,h+1} = w_{c,i,h+1}$}
        
        \State $h$++
    \EndWhile
    
    \State Send $(i,c, U)$ to the Server. \Comment{$U$ represents $U_{i,c} = U_{i,c,s_{i,c}} = \sum_{h=0}^{s_{i,c}-1} g_{c,i,h}$}
    \State $i$++
\EndWhile
   
\EndProcedure

\end{algorithmic}
\end{algorithm} 

We propose to have a server to maintain a global model which is updated according to Algorithm \ref{alg:server}. The server receives updates from clients who work on local models by executing Algorithm \ref{alg:client}. Before any computation starts both server and clients agree on the used diminishing step size sequence, the increasing sample (mini batch) sequences, initial default global model, and permissible delay function $\tau$, see Algorithm \ref{alg:setup}.
Note that the Interrupt Service Routines (ISR) at the client and service side interrupt the main execution as soon as a message is received in which case the ISR routines execute before returning to the main code.

The algorithms have added comments with interpretation of the locally computed variables. These interpretations/annotations are used in next sections when proving properties.


We remark that $s_{i,c}$ are initialized by a coin flipping procedure in \Call{Setup}{}. Since the $s_i$ are increasing, we may approximate $s_{i,c}\approx \mathbb{E}[s_{i,c}]=p_c s_i$ (because of the law of large numbers). This is what we use when adding differential privacy in Section \ref{sec:DP} (it makes $s_{i,c}$ well-defined as an increasing sequence so that our theorems apply).

We also remark that we may think of the server as a number of connected separate aggregators that serve as proxies between the clients and server. This allows a hierarchical communication network where each aggregator only aggregates updates over a subset of clients and the central server aggregates the aggregated updates. This define a communication tree (which can be made more deep if suitable). Extra layers of aggregators allows us to satisfy network throughput constraints (at the price of added communication latency).




\vspace{3pt}
\noindent
{\bf Related work.}
Asynchronous training~\citep{zinkevich2009slow,lian2015asynchronous,lian2017asynchronous,zheng2017asynchronous,meng2017asynchronous,stich2018local,shi2019distributed} is widely used in traditional distributed stochastic gradient descent (SGD). Typically, asynchronous SGD converges faster than synchronous SGD in real time due to parallelism. However, classic asynchronous SGD directly sends gradients to the server after each local update, which is not practical for edge devices (such as mobile or IoT devices) due to the unreliable and slow communication.

In \citep{cong2019,yang2019} asynchronous training combined  with federated optimization is proposed. Specifically, the server and workers conduct updates asynchronously: the server immediately updates the global model whenever it receives a local model from clients. Therefore, the communication between the server and workers is non-blocking and more effective. However, asynchronous FL may be without convergence guarantees and can diverge in practical settings when data are heterogeneous \citep{tianli2019federated}.  In \citep{yang2019} a temporally weighted aggregation strategy by the server is introduced in order to make use of the previously trained local models and to eventually enhance the measured accuracy and convergence of the global model at the server (we notice that diminishing step sizes can also be used to achieve the same effect as a temporally weighted strategy).

In \citep{tianli2019hetero}, the authors introduce FedProx which is a modification of FedAvg (i.e., original FL algorithm of \citep{mcmahan}). In FedProx, the clients solve a proximal minimization problem rather than traditional minimization as in FedAvg. For theory, the authors use $B$-local dissimilarity and bounded dissimilarity assumptions for the global objective function. This implies that there is a bounded gradient assumption applied to the global objective function. Moreover, their proof allows the global objective function to  be strongly convex. However, the bounded gradient assumption is in conflict with assuming strong convexity as explained in \citep{nguyen2018sgd,nguyen2018new}.
Since these two assumptions cannot be used together for a convergence analysis, the convergence analysis of FedProx is not complete.  Our analysis  does not use the bounded gradient assumption for analysing strongly convex objective functions. 

\vspace{3pt}
\noindent
{\bf Horizontal federated learning} or sample-based federated learning 
\citep{mcmahan,tianli2019hetero,bonawitz2019towards,konevcny2016federated,konevcny2016federated_com}
is applied in scenarios where local data sets in clients share the same feature space but are different in their samples. 
In a typical architecture for a horizontal federated learning system  
clients with the same data structure collaboratively train independently a global model with the help of a parameter server. 
A typical assumption \citep{phong2018privacy} is that the participants are honest while the server is honest-but-curious. Ideally,  no leakage of  information from any participants to other participants and the server is allowed. This can be accomplished by secure multi-party computation \citep{evans2017pragmatic} which turns out to be communication heavy. The more practical alternative is to apply differential privacy techniques (which controls the amount of leakage).



\section{Proofs asynchronous distributed SGD} 

\subsection{Proof of Theorem \ref{thmalg}}
\label{appmain}

The clients in the distributed computation apply recursion (\ref{eqwM2a}). We want to label each recursion with an iteration count $t$; this can then be used to compute with which delay function the labeled sequence $\{w_t \}$ is consistent.
In order to find an ordering based on $t$ we first define a mapping $\rho$ from the annotated labels $(c,i,h)$ in \Call{MainClient}{} to $t$:
%
$$\rho(c,i,h)= (\sum_{l < i} s_{l}) + \min\{ t' \ : \ h=|\{ t\leq t' \ : \ a(i,t)=c\}|\} ,$$ 
where sequence sample size sequence $\{s_i\}$ and labelling function $a(.,.)$ are defined in \Call{Setup}{}.

Notice that given $t$, we can compute $i$ as the largest index for which $\sum_{l<i} s_{l} \leq t$, compute $t'=t- \sum_{l<i} s_{l}$ and $c=a(i,t')$, and compute $h=|\{ t\leq t' \ : \ a(i,t)=c\}|$. This procedure inverts $\rho$ since $\rho(c,i,h)=t$, hence, $\rho$ is bijective.  We use $\rho$ to order local models $\hat{w}_{c,i,h}$ by writing $\hat{w}_t=\hat{w}_{c,i,h}$ with $t=\rho(c,i,h)$. Similarly, we write $\xi_t= \xi_{c,i,h}$ for local training data samples. 


From the invariant in \Call{MainServer}{} we infer that  $\hat{v}_k$ includes all the aggregate updates $U_{i,c}$ for $i< k$ and $c\in {1,\ldots n}$. See Algorithm \ref{alg:client} for the clients, 
$$\sum_{c\in \{1,\ldots, n\}} U_{i,c} = \sum_{c\in \{1, \ldots, n\}} \sum_{h=0}^{s_{i,c}-1} g_{c,i,h} \mbox{ with } g_{c,i,h}= \nabla f(\hat{w}_{c,i,h}; \xi_{c,i,h}).$$
By using mapping $\rho$, this is equal to
\begin{eqnarray*}
\sum_{c\in \{1, \ldots, n\}} \sum_{h=0}^{s_{i,c}-1} \nabla f(\hat{w}_{c,i,h}; \xi_{c,i,h}) &=&
\sum_{c\in \{1, \ldots, n\}} \sum_{h=0}^{s_{i,c}-1} \nabla f(\hat{w}_{\rho(c,i,h)}; \xi_{\rho(c,i,h)}) \\
&=& \sum_{t=s_0+\ldots s_{i-1}}^{s_0+\ldots+s_i-1} \nabla f(\hat{w}_{t}; \xi_{t}).
\end{eqnarray*}
This implies that 
$\hat{v}_k$, includes all the gradient updates (across all clients) that correspond to $\hat{w}_t$ for $t\leq s_0+\ldots+s_{k-1}-1= t_{glob}-t_{delay}$ in the notation of \Call{MainClient}{}.

Let $\rho(c,i,h)=t$. Notice that $t= \rho(c,i,h) \leq s_0+ \ldots + s_i -(s_{i,c}-h)-1=t_{glob}$. In \Call{MainClient}{} we wait as long as $\tau(t_{glob})= t_{delay}$. This means that $h$ will not further increase until $k$ increases in client $c$'s \Call{ISRReceive}{} when a new global model is received. This implies $t_{delay}\leq \tau(t_{glob})$ as an invariant of the algorithm. Because $t-\tau(t)$ is an increasing function in $t$, the derived inequality implies  $t-\tau(t)\leq t_{glob}-\tau(t_{glob}) \leq t_{glob}-t_{delay}$. Notice that $\hat{w}_t=\hat{w}_{c,i,h}$  includes the updates aggregated in $\hat{v}_k$ as last received by client $c$'s \Call{ISRReceive}{}. We concluded above that $\hat{v}_k$ includes all gradient updates that correspond to gradient computations up to iteration count $t_{glob}-t_{delay}$. This includes gradient computations up to iteration count $t-\tau(t)$. Therefore, the computed sequence $\{w_t\}$ is consistent with delay function $\tau$.

As a second consequence of mapping $\rho$ we notice that $\{\xi_t\}=\{\xi_{c,i,h}\}$. In fact $\xi_t=\xi_{c,i,h}$ for $(c,i,h)=\rho^{-1}(t)$. 
Notice that \Call{Setup}{} constructs the mapping $a(i,t)=c$  which is used to define $\rho$. Mapping $a(i,t)$ represents a random table of client assignments $c$ generated by using a probability vector $(p_1,\ldots, p_n)$ (with $c$ being selected with probability $p_c$ for the $(i,t)$-th entry). So, $\xi_t\sim {\cal D}_c$ in \Call{MainClient}{} with probability $p_c$. This means that $\xi_t\sim {\cal D}$ with ${\cal D} = \sum_{c=1}^n p_c {\cal D}_c$.

\subsection{Invariant $t_{delay}\leq t_{glob}$} \label{appinv}

We propose to use increasing sample size sequences $\{s_{i,c}\}$ such that we can replace the condition $\tau(t_{glob})=t_{delay}$ of the wait loop by 
$i = k + d$,
where $d$ is a threshold such that for all $i\geq d+1$,
\begin{equation} \tau(\sum_{j=0}^i s_j)\geq \sum_{j=i-d}^i s_j.
\label{eqS}
\end{equation}
The new wait loop guarantees that $i-k\leq d$ as an invariant of the algorithm. This implies the old invariant $t_{delay}\leq t_{glob}$ for the following reason: Let $i\geq d+1$. Since $t-\tau(t)$ is increasing,
$$t_{glob}-\tau(t_{glob}) \leq \sum_{j=0}^i s_j -\tau(\sum_{j=0}^i s_j) \leq \sum_{j=0}^i s_j - \sum_{j=i-d}^i s_j = \sum_{j=0}^{i-d-1} s_i.$$
Together with $i-d\leq k$, this implies
$$ \tau(t_{glob}) \geq t_{glob} - \sum_{j=0}^{i-d-1} s_i = s_{i-d}+\ldots + s_i -(s_{i,c}-h)-1) \geq 
s_{k}+\ldots + s_i -(s_{i,c}-h)-1)= t_{delay}.$$
This shows that the new invariant implies the old one when $i\geq d+1$.

\subsection{Increasing sample size sequences} \label{appinc}

The following lemma shows how to construct an increasing sample size sequence given a delay function $\tau$.

\begin{lem} \label{lemsample}
Let $g>1$.
Let function $\gamma(z)$  be increasing (i.e., $\gamma'(z)\geq 0$)  and $\geq 1$ for $z\geq 0$, with the additional property that 
$\gamma(z)\geq z \gamma'(z) \frac{g}{g-1}$
for $z\geq 0$. Suppose that 
$$ \tau(x) = M_1 +   (\frac{x+M_0}{\gamma(x+M_0)})^{1/g}  
\mbox{ with } 
M_0 \geq ((m+1)\frac{g-1}{g})^{g/(g-1)} \mbox{ and } M_1\geq d+1 $$
for some integer $m\geq 0$
(we can choose $M_0=0$ if $m=0$), and define
$$
S(x) =  ( \frac{x}{\omega(x)} \frac{g-1}{g} )^{1/(g-1)} \mbox{ with } \omega(x) = \gamma( (x \frac{g-1}{g} )^{g/(g-1)} ).
$$
Then $s_i = \lceil \frac{1}{d+1}S(\frac{m+i+1}{d+1}) \rceil$ satisfies property (\ref{eqS}) (i.e., (\ref{eqtausample})).
\end{lem}



\noindent
{\bf Proof.}
Since $\gamma(z)$ is increasing and $\geq 1$ for $z\geq 0$, also $\omega(x)$ is increasing and $\geq 1$ for $x\geq 0$ (notice that $g>1$). 

We also want to show that $z/\omega(z)$ is increasing for $z\geq 0$. Its derivative is equal to
$$ \frac{1}{\omega(z)}(1-\frac{z \omega'(z) }{\omega(z)})
$$
which is $\geq 0$ if $\omega(z) \geq z \omega'(z) $ (because $\omega(z)$ is positive).
The latter is equivalent (by $\omega$'s definition) to
$$
\gamma( (z \frac{g-1}{g} )^{g/(g-1)} )
\geq z  
\gamma'( (z \frac{g-1}{g} )^{g/(g-1)} )
\frac{g}{g-1} (z \frac{g-1}{g} )^{g/(g-1)-1} \frac{g-1}{g}.
$$
This is implied by $\gamma(y) \geq y \gamma'(y)\frac{g}{g-1}$ for $y= (z \frac{g-1}{g} )^{g/(g-1)}$. From our assumptions on $\gamma(z)$ we infer that this is true. 

Since $x/\omega(x)$ is increasing, also $S(x)$ is increasing for $x\geq 0$. This implies
\begin{eqnarray*}
\sum_{j=0}^i s_j 
&=& \sum_{j=0}^i \lceil \frac{1}{d+1} S(\frac{m+j+1}{d+1}) \rceil
\geq \sum_{j=m}^{i+m} \frac{1}{d+1}S(\frac{j+1}{d+1})  \\
&=& 
\sum_{j=0}^{i+m} \frac{1}{d+1}S(\frac{j+1}{d+1})  - \sum_{j=0}^{m-1} S(j+1) \\
&\geq &\int_{z=0}^{m+i+1} \frac{1}{d+1}S(\frac{z}{d+1}) dz -
\int_{z=0}^{m+1} \frac{1}{d+1}S(\frac{z}{d+1}) dz \\
&=&
\int_{z=0}^{(m+i+1)/(d+1)} S(z) dz -
\int_{z=0}^{(m+1)/(d+1)} S(z) dz.
\end{eqnarray*}

Since $\omega(z)\geq 1$,
$$ \int_{z=0}^{(m+1)/(d+1)} S(z) dz \leq
\int_{z=0}^{(m+1)/(d+1)} (z \frac{g-1}{g})^{1/(g-1)} dz = (\frac{m+1}{d+1} \cdot \frac{g-1}{g})^{g/(g-1)}\leq M_0.
$$
Let $x=(m+i+1)/(d+1)$ 
Then,
$$ \sum_{j=0}^i s_j  \geq \int_{z=0}^{x} S(z) dz -M_0.$$

Because $\omega(z)$ is increasing for $z\geq 0$ we have
$$\int_{z=0}^{x} S(z) dz
\geq \frac{1}{\omega(x)^{1/(g-1)}} \cdot
\int_{z=0}^{x} (z \frac{g-1}{g})^{1/(g-1)} dz
=
\frac{1}{\omega(x)^{1/(g-1)}} \cdot
(x \frac{g-1}{g})^{g/(g-1)}.
$$ 
Since $x/\gamma(x)$ is increasing (its derivative is equal to $\frac{1}{\gamma(x)}(1-\frac{x \gamma'(x)}{\gamma(x)})$ and is positive by our assumptions on $\gamma(x)$), also $\tau(x)$ is increasing and we infer 
$$ \tau( \sum_{j=0}^i s_j ) \geq
\tau( \int_{z=0}^{x} S(z) dz -M_0)
\geq \tau(\frac{1}{\omega(x)^{1/(g-1)}} \cdot
(x \frac{g-1}{g})^{g/(g-1)} -M_0).$$
By $\tau$'s definition,
the right hand side is equal to
\begin{equation} M_1+
\frac{(\frac{1}{\omega(x)^{1/(g-1)}}(x\frac{g-1}{g})^{g/(g-1)} )^{1/g}}
{\gamma( \frac{1}{\omega(x)^{1/(g-1)}} \cdot
(x \frac{g-1}{g})^{g/(g-1)} )^{1/g} } 
. \label{eqd}
\end{equation}

Since $S(x)$ is increasing,
\begin{eqnarray*}
\sum_{j=i-d}^i s_j &=& \sum_{j=i-d}^i \lceil \frac{1}{d+1} S(\frac{m+j+1}{d+1})\rceil 
\leq 
(d+1) \lceil \frac{1}{d+1} S(\frac{m+i+1}{d+1})\rceil  \\
&\leq& d+1 + S(\frac{m+i+1}{d+1})=d+1+S(x).
\end{eqnarray*}
Since $M_1\geq d+1$, the right hand side is at most (\ref{eqd}) 
if
$$
\frac{(\frac{1}{\omega(x)^{1/(g-1)}}(x\frac{g-1}{g})^{g/(g-1)})^{1/g}}
{\gamma( \frac{1}{\omega(x)^{1/(g-1)}} \cdot
(x \frac{g-1}{g})^{g/(g-1)})^{1/g} }
\geq ( \frac{x}{\omega(x)} \frac{g-1}{g} )^{1/(g-1)} = S(x).
$$
After raising to the power $g$ and reordering terms this is equivalent to
%
%
%
$$ \omega(x)  \geq \gamma( \frac{1}{\omega(x)^{1/(g-1)}} \cdot
(x \frac{g-1}{g})^{g/(g-1)} ).$$
Since $\omega(x)\geq 1$ and $\gamma(z)$ is increasing, this is implied by
$$ \omega(x) \geq 
\gamma( 
(x \frac{g-1}{g})^{g/(g-1)} ).$$
%
%
Since this is true by the definition of $\omega(x)$, we conclude inequality  (\ref{eqS}).


\subsection{Diminishing step size sequences} \label{appstep}

We assume the sample size sequence $s_i=\lceil \frac{1}{d+1} S(\frac{m+i+1}{d+1})\rceil$ developed in the previous subsection.

\begin{lem} \label{lemstep}
We assume the sample size sequence $s_i=\lceil \frac{1}{d+1} S(\frac{m+i+1}{d+1})\rceil$ of Lemma \ref{lemsample}.
Let $\{E_t\}$ be a constant or increasing sequence with $E_t\geq 1$. 
For $i\geq 0$, we define
$$ \bar{E}_i = E_{\sum_{j=0}^i s_j}. $$
We define $\bar{E}_{-1}=E_0$ and 
we assume for $i\geq 0$ that
$$ \bar{E}_{i} \leq 2 \bar{E}_{i-1} \mbox{ and } 
s_0-1\leq E_0
.$$

Let $q\geq 0$, and $a_0$ and $a_1$ be constants
such that 
$$ a_1= 
a_0 \cdot \max\{ 3, (1 + \frac{m+2}{m+1})^{1/(g-1)} \}^q.
$$
For $g\geq 2$ and $q\leq 1$ this implies
$a_0\leq a_1 \leq 3\cdot a_0$.

Then there exists a diminishing step size sequence $\eta_t = \alpha_t / (t+E_t)^q$ with $a_0\leq \alpha_t \leq a_1$ such that
$$\eta_t = \frac{\alpha_t}{(t+E_t)^q} = \frac{a_0}{((\sum_{j=0}^{i-1} s_j) +\bar{E}_{i-1})^q} = \bar{\eta}_i
$$ 
for $t\in \{(\sum_{j=0}^{i-1} s_j), \ldots, (\sum_{j=0}^{i-1} s_j) +s_i -1 \}$.
\end{lem}


\noindent
{\bf Proof.} 
For $i\geq 1$, we first show a relation between $s_i$ and $s_{i-1}$ (notice that $\omega$ is increasing):
\begin{eqnarray}
s_i-1 &\leq&
\frac{1}{d+1} S(\frac{m+i+1}{d+1})
=
\frac{1}{d+1} (\frac{1}{\omega(\frac{m+i+1}{d+1})}\frac{m+i+1}{d+1} \frac{g-1}{g} )^{1/(g-1)} \nonumber \\
&\leq &
\frac{1}{d+1} (\frac{1}{\omega(\frac{m+i}{d+1})}\frac{m+i+1}{d+1} \frac{g-1}{g} )^{1/(g-1)} \nonumber \\
&=&
(\frac{m+i+1}{m+i})^{1/(g-1)} \frac{1}{d+1} (\frac{1}{\omega(\frac{m+i}{d+1})}\frac{m+i}{d+1} \frac{g-1}{g} )^{1/(g-1)} \nonumber \\
&=&
(\frac{m+i+1}{m+i})^{1/(g-1)}
\frac{1}{d+1} S(\frac{m+i}{d+1})
\leq 
(\frac{m+2}{m+1})^{1/(g-1)} s_{i-1} \label{sieq}
\end{eqnarray}
For $i\geq 1$ and $t\in \{(\sum_{j=0}^{i-1} s_j), \ldots, (\sum_{j=0}^{i-1} s_j) +s_i -1 \}$, we are now able to derive a bound
\begin{eqnarray*}
\alpha_t &=& a_0 \frac{ (t+E_t)^q }{((\sum_{j=0}^{i-1} s_j) +\bar{E}_{i-1})^{q}}
\leq 
a_0 (\frac{ (\sum_{j=0}^{i} s_j)-1+\bar{E}_{i} }{(\sum_{j=0}^{i-1} s_j) +\bar{E}_{i-1}})^q \\
&\leq &
a_0 (\frac{ (\sum_{j=0}^{i} s_j)-1+2\bar{E}_{i-1} }{(\sum_{j=0}^{i-1} s_j) +\bar{E}_{i-1}})^q
=
a_0 (1 +\frac{ s_i-1+\bar{E}_{i-1}  }{(\sum_{j=0}^{i-1} s_j) +\bar{E}_{i-1}})^q \\
&\leq &
a_0 (1 +\frac{ (1+1/(m+1))^{1/(g-1)} s_{i-1}+\bar{E}_{i-1}  }{(\sum_{j=0}^{i-1} s_j) +\bar{E}_{i-1}})^q \\
&\leq&
a_0 (1 +\frac{ (1+1/(m+1))^{1/(g-1)} (\sum_{j=0}^{i-1} s_j)+\bar{E}_{i-1}  }{(\sum_{j=0}^{i-1} s_j) +\bar{E}_{i-1}})^q \\
&\leq&
a_0 (1 +(1+1/(m+1))^{1/(g-1)})^q
= 
a_0 (1 + (\frac{m+2}{m+1})^{1/(g-1)})^q.
\end{eqnarray*}

Notice that for $g\geq 2$ and $q\leq 1$, this bound is at most
$3\cdot a_0$.

We still need to analyse the case $i=0$. This gives the bound 
$$
\alpha_t = a_0 \frac{ (t+E_t)^q }{((\sum_{j=0}^{i-1} s_j) +E_0)^{q}}
=
a_0 \frac{ (t+E_t)^q }{E_0^{q}}
\leq 
a_0 \frac{ (s_0-1+\bar{E}_0)^q }{E_0^{q}}
\leq a_0 \cdot 3^q.
$$

The step size for iteration $t$ is equal to
$$\eta_t = \frac{\alpha_t}{(t+E_t)^q} = \frac{a_0}{((\sum_{j=0}^{i-1} s_j) +E_{i-1})^q}.$$


\section{Analysis general recursion}


The optimization problem for training many Machine Learning (ML) models using a training set $\{\xi_i\}_{i=1}^m$ of $m$ samples can be formulated as a finite-sum minimization problem as follows
\begin{equation}\label{eq:finite_sum_main}
\min_{w \in \mathbb{R}^d} \left\{ F(w) = \frac{1}{m}
\sum_{i=1}^m f(w; \xi_i) \right\}.
\end{equation}
The objective is to minimize a loss function with respect to model parameters $w$. This problem is known as empirical risk minimization and it covers a wide range of convex and non-convex problems from the ML domain, including, but not limited to, logistic regression, multi-kernel learning, conditional random fields and neural networks.
We are interested in solving the following more general stochastic optimization problem with respect to some distribution $\mathcal{D}$:
\begin{align}
\min_{w \in \mathbb{R}^d} \left\{ F(w) = \mathbb{E}_{\xi \sim \mathcal{D}} [ f(w;\xi) ] \right\},  \label{main_prob_expected_risk}  
\end{align}
where $F$ has a Lipschitz continuous gradient and $f$ has a \emph{finite lower bound} for every $\xi$.

The general form \eqref{main_prob_expected_risk} can be solved by using SGD as described in Algorithm \ref{sgd_algorithm}.
 Thanks to its simplicity in implementation and efficiency in dealing with large scale datasets, stochastic gradient descent, originally introduced in \citep{RM1951}, has become the method of choice for solving not only (\ref{eq:finite_sum_main}) when $m$ is large but also (\ref{main_prob_expected_risk}).

\begin{algorithm}[ht]
  \caption{Stochastic Gradient Descent (SGD) Method}
  \label{sgd_algorithm}
\begin{algorithmic}[1]
  \State {\bfseries Initialize:} $w_0$
  \State {\bfseries Iterate:}
  \For{$t=0,1,2,\dots$}
  \State Choose a step size (i.e., learning rate) $\eta_t>0$. 
  \State Generate a random variable $\xi_t$.
  \State Compute a stochastic gradient $\nabla f(w_{t};\xi_{t}).$
  \State Update the new iterate $w_{t+1} = w_{t} - \eta_t \nabla f(w_{t};\xi_{t})$.
  \EndFor
\end{algorithmic}
\end{algorithm}

\subsection{The Hogwild! algorithm}
\label{app-rec}

To speed up SGD, an asynchronous SGD known as Hogwild! was introduced in~\citep{Hogwild}. 
Here, multiple computing threads work together and update shared memory in asynchronous fashion. The shared memory stores the most recently computed weight as a result of the SGD iterations computed by each of the computing threads.  Writes to and reads from vector positions in shared memory can be inconsistent. As a result a computing thread may start reading positions of the current weight vector from shared memory while these positions are updated by other computing threads out-of-order. Only writes to and reads from shared memory positions are considered atomic. This means that, when a computing thread reads the `current' weight vector from shared memory,  this weight vector is a mix of partial updates to the weight vector from other computing threads that executed previous SGD iterations. 
 
\citep{nguyen2018sgd,nguyen2018new}
introduce a general recursion for $w_t$.
The recursion explains which positions in $w_t$ should be updated in order to compute $w_{t+1}$. Since $w_t$ is stored in shared memory and is being updated in a possibly non-consistent way by multiple cores who each perform recursions, the shared memory will contain a vector $w$ whose entries represent a mix of updates. That is, before performing the computation of a recursion, a computing thread will first read  $w$ from shared memory, however, while reading $w$ from shared memory, the entries in $w$ are being updated out of order. The final vector $\hat{w}_t$ read by the computing thread represents an aggregate of a mix of updates in previous iterations.

The general recursion (parts of text extracted from \citep{nguyen2018new}) is defined as follows: For $t\geq 0$,
\begin{equation}
 w_{t+1} = w_t - \eta_t d_{\xi_t}  S^{\xi_t}_{u_t} \nabla f(\hat{w}_t;\xi_t),\label{eqwM}
 \end{equation}
 where
 \begin{itemize}
 \item $\hat{w}_t$ represents the vector used in computing the gradient $\nabla f(\hat{w}_t;\xi_t)$ and whose entries have been read (one by one)  from  an aggregate of a mix of  previous updates that led to $w_{j}$, $j\leq t$, and
 \item the $S^{\xi_t}_{u_t}$ are diagonal 0/1-matrices with the property that there exist real numbers $d_\xi$ satisfying
\begin{equation} d_\xi \mathbb{E}[S^\xi_u | \xi] = D_\xi, \label{eq:SexpM} \end{equation}
where the expectation is taken over $u$ and $D_\xi$ is the diagonal 0/1 matrix whose $1$-entries correspond to the non-zero positions in $\nabla f(w;\xi)$ in the following sense: The $i$-th entry of $D_\xi$'s diagonal is equal to 1 if and only if there exists a $w$ such that the $i$-th position of $\nabla f(w;\xi)$ is non-zero. 
\end{itemize}

The role of matrix $S^{\xi_t}_{u_t}$ is that it filters which positions of gradient $\nabla f(\hat{w}_t;\xi_t)$ play a role in (\ref{eqwM}) and need to be computed. Notice that $D_\xi$ represents the support of $\nabla f(w;\xi)$; by $|D_\xi|$ we denote the number of 1s in $D_\xi$, i.e., $|D_\xi|$ equals the size of the support of $\nabla f(w;\xi)$.

We restrict ourselves to choosing (i.e., fixing a-priori) {\em non-empty} matrices  $S^\xi_u$ that ``partition'' $D_\xi$ in $D$ approximately ``equally sized'' $S^\xi_u$: 
$$ \sum_u S^\xi_u = D_\xi, $$
where each matrix $S^\xi_u$ has either $\lfloor |D_\xi|/D \rfloor$ or $\lceil |D_\xi|/D \rceil$ ones on its diagonal. We uniformly choose one of the matrices $S^{\xi_t}_{u_t}$ in (\ref{eqwM}), hence, $d_\xi$ equals the number of matrices $S^\xi_u$, see (\ref{eq:SexpM}).

In order to explain recursion (\ref{eqwM}) we  consider two special cases. For $D=\bar{\Delta}$, where 
$$ \bar{\Delta} = \max_\xi \{ |D_\xi|\}$$
represents the maximum number of non-zero positions in any gradient computation $f(w;\xi)$, we have that for all $\xi$, there are exactly $|D_\xi|$ diagonal matrices $S^\xi_u$ with a single 1 representing each of the elements in $D_\xi$. Since  $p_\xi(u)= 1/|D_\xi|$ is the uniform distribution, we have $\mathbb{E}[S^\xi_u | \xi] = D_\xi / |D_\xi|$, hence, $d_\xi = |D_\xi|$. This gives the recursion
\begin{equation}
 w_{t+1} = w_t - \eta_t |D_\xi|  [ \nabla f(\hat{w}_t;\xi_t)]_{u_t},\label{eqwM1}
 \end{equation}
 where $ [ \nabla f(\hat{w}_t;\xi_t)]_{u_t}$ denotes the $u_t$-th position of $\nabla f(\hat{w}_t;\xi_t)$ and where $u_t$ is a uniformly selected position that corresponds to a non-zero entry in  $\nabla f(\hat{w}_t;\xi_t)$.
 
At the other extreme, for $D=1$, we have exactly one matrix $S^\xi_1=D_\xi$ for each $\xi$, and we have $d_\xi=1$. This gives the recursion
\begin{equation}
 w_{t+1} = w_t - \eta_t  \nabla f(\hat{w}_t;\xi_t).\label{eqwM2}
 \end{equation}
Recursion (\ref{eqwM2}) represents Hogwild!. In a single-thread setting where updates are done in a fully consistent way, i.e. $\hat{w}_t=w_t$, yields SGD.

 
 Algorithm \ref{HogWildAlgorithm} gives the pseudo code corresponding to recursion (\ref{eqwM}) with our choice of sets $S^\xi_u$ (for parameter $D$).
 
 \begin{algorithm}
\caption{Hogwild! general recursion}
\label{HogWildAlgorithm}
\begin{algorithmic}[1]

   \State {\bf Input:} $w_{0} \in \mathbb{R}^d$
   \For{$t=0,1,2,\dotsc$ {\bf in parallel}} 
    
  \State read each position of shared memory $w$
  denoted by $\hat w_t$  {\bf (each position read is atomic)}
  \State draw a random sample $\xi_t$ and a random ``filter'' $S^{\xi_t}_{u_t}$
  \For{positions $h$ where $S^{\xi_t}_{u_t}$ has a 1 on its diagonal}
   \State compute $g_h$ as the gradient $\nabla f(\hat{w}_t;\xi_t)$ at position $h$
   \State add $\eta_t d_{\xi_t} g_h$ to the entry at position $h$ of $w$ in shared memory {\bf (each position update is atomic)}
   \EndFor
   \EndFor
\end{algorithmic}
\end{algorithm}

In order to use Algorithm \ref{HogWildAlgorithm} in federated learning, we use the following reinterpretation of shared memory and computing threads.
Clients in FL represent the different computing threads. The server plays the role of shared memory where locally  computed weight vectors at the clients are `aggregated' as in (\ref{eqwM}). Each Client$_j$  first receives each of the entries of the weight vector $w$ currently stored at the server. This is done by successive atomic reads from $w$ as described above. Next the collected entries form together a vector $\hat{w}_t$ and the gradient $\nabla f(\hat{w}_t;\xi_t)$ for some sample $\xi_t \sim {\mathcal D}$ is computed. The gradient multiplied by a step size $\eta_t$ and correction factor $d_{\xi_t}$ is subtracted from the weight vector stored at the server. This is done by a series of successive atomic writes. Each of the clients (just like the computing threads above) work in parallel continuously updating entries in the weight vector $w$ stored at the server. 

In the context of different computing threads atomically reading and writing entries of vector $w$ from shared memory, we define the amount of asynchronous behavior by parameter $\tau$ as in \citep{nguyen2018new}:

\begin{defn} 
We say that weight vector $w$ stored at the server is {\em consistent with delay $\tau$}  with respect to recursion (\ref{eqwM}) if, for all $t$, vector $\hat{w}_t$ includes the aggregate of the updates up to and including those made during the $(t-\tau)$-th iteration (where (\ref{eqwM}) defines the $(t+1)$-st iteration). Each position read from shared memory is atomic and each position update to shared memory is atomic (in that these cannot be interrupted by another update to the same position).
\end{defn}

Even though this (original) definition does not consider $\tau$ as a function of the iteration count $t$, the next subsections do summarize how $\tau$ can depend as a function on $t$.

\subsection{Convergence rate for strongly convex problems}

We shortly explain known results on the convergence rate 
for strongly convex problems. Specifically, we have the following assumptions:

\begin{ass}[$L$-smooth]
\label{ass_smooth}
$f(w;\xi)$ is $L$-smooth for every realization of $\xi$, i.e., there exists a constant $L > 0$ such that, $\forall w,w' \in \mathbb{R}^d$, 
\begin{align*}
\| \nabla f(w;\xi) - \nabla f(w';\xi) \| \leq L \| w - w' \|. \label{eq:Lsmooth_basic}
\end{align*} 
\end{ass}

\begin{ass}\label{ass_convex}
$f(w;\xi)$ is convex for every realization of $\xi$, i.e., $\forall w,w' \in \mathbb{R}^d$, 
\begin{gather*}
f(w;\xi)  - f(w';\xi) \geq \langle \nabla f(w';\xi),(w - w') \rangle.
\end{gather*}
\end{ass}

\begin{ass}[$\mu$-strongly convex]
\label{ass_stronglyconvex}
The objective function $F: \mathbb{R}^d \to \mathbb{R}$ is a $\mu$-strongly convex, i.e., there exists a constant $\mu > 0$ such that $\forall w,w' \in \mathbb{R}^d$, 
\begin{gather*}
F(w) - F(w') \geq \langle \nabla F(w'), (w - w') \rangle + \frac{\mu}{2}\|w - w'\|^2. \label{eq:stronglyconvex_00}
\end{gather*}
\end{ass}

Being strongly convex implies that $F$ has a global minimum $w_{*}$. For $w_{*}$ we assume:

\begin{ass}[Finite $\sigma$]
\label{ass_finitesigma}
Let $N = 2 \mathbb{E}[ \|\nabla f(w_{*}; \xi)\|^2 ]$ where $w_{*} = \arg \min_w F(w)$. We require $N < \infty$. 
\end{ass}

Notice that we do not assume the bounded gradient assumption which assumes $\mathbb{E}[ \|\nabla f(w_{*}; \xi)\|^2 ]$ for all $w\in \mathbb{R}^d $ (not only $w=w_{*}$ as in Assumption \ref{ass_finitesigma}) and is in conflict with assuming strong convexity as explained in \citep{nguyen2018sgd,nguyen2018new}.

\subsubsection{Constant step sizes} \label{appconst}
Algorithm~\ref{HogWildAlgorithm} for $D=1$ corresponds to Hogwild! with recursion (\ref{eqwM2}). For finite-sum problems,
\citep{Leblond2018}
 proves for 
  constant step sizes $\eta_t=\eta=\frac{a}{L}$ with delay 
\begin{equation}
\tau\leq \frac{1}{\eta\mu}
\label{tau2}
\end{equation}
  and parameter 
  $a\leq (5(1+2\tau \sqrt{\Delta})\sqrt{1+\frac{\mu}{2L}\min\{\frac{1}{\sqrt{\Delta}},\tau\}})^{-1}$ as a function of $\tau$, where $\Delta$ measures  sparsity according to Definition 7 in~\citep{Leblond2018}, 
%
that the convergence rate $\mathbb{E}[\|\hat{w}_t - w_*\|^2]$ is at most
$$
    \mathbb{E}[\|\hat{w}_t - w_*\|^2]\leq 2 (1-\rho)^t \|w_0-w_*\|^2 + b,
    $$
where $\rho=\frac{a L}{\mu}$ and $b=(\frac{4\eta(C_1+\tau C_2)}{\mu} + 2\eta^2C_1\tau )N$ for 
$C_1= 1 + \sqrt{\Delta}\tau$ and 
$C_2= \sqrt{\Delta} + \eta \mu C_1$.

With a fixed learning rate $\eta_t = \eta$, SGD and Hogwild! may provide fast initial improvement, after which it oscillates within a region containing a solution \citep{BottouCN18,CheeT18}. The upper bound on the convergence rate shows that convergence is to within some range of the optimal value; we have $\mathbb{E}[\|\hat{w}_t - w_*\|^2]= \mathcal{O}(\eta)$.
 Hence, SGD and Hogwild! can fail to converge to a solution. It is known that the behavior of SGD is strongly dependent on the chosen learning rate and on the variance of the stochastic gradients. To overcome this issue, there are two main lines of research that have been proposed in literature: variance reduction methods 
 and diminishing learning rate schemes. 
 These algorithms guarantee to converge to the optimal value.   
 
 Upper bound (\ref{tau2}) allows us to set the maximal sample size $s=s_i$  in terms the amount of asynchronous behavior allowed by $d$ in (\ref{eqS}). This allows an informed decision on how to reduce the number of broadcast messages as much as possible. We need $\tau=(d+1) s \leq \frac{1}{\eta \mu}$, i.e., $s\leq \frac{1}{\eta \mu (d+1)}$ which is typically large.

\subsubsection{Diminishing step sizes} \label{app-sc-dim}
Algorithm~\ref{HogWildAlgorithm} for diminishing step sizes $\eta_t = \frac{\alpha_t}{\mu(t+E)}$ with $4\leq \alpha_t \leq\alpha$ and $E = \max\{ 2\tau, \frac{4 L \alpha D}{\mu}\}$ has the following properties, see \citep{nguyen2018sgd,nguyen2018new}: 
The expected number of single vector entry updates after $t$ iterations is equal to $t' = t \bar{\Delta}_D /D$ with  $\bar{\Delta}_D = D \cdot \mathbb{E}[\lceil |D_\xi|/D \rceil]$ and both convergence rates $ \mathbb{E}[\|\hat{w}_{t} - w_* \|^2]$ and $\mathbb{E}[\|w_{t} - w_* \|^2]$ are at most
$$\frac{4\alpha^2D N}{\mu^2} \frac{t}{(t + E - 1)^2} + O\left(\frac{\ln t}{(t+E-1)^{2}}\right).$$
 In terms of the expected number $t'$ of  single vector entry updates after $t$ iterations, both $\mathbb{E}[\|\hat{w}_{t} - w_* \|^2]$ and $\mathbb{E}[\|w_{t} - w_* \|^2]$ are at most
$$\frac{4\alpha^2 \bar{\Delta}_D N}{\mu^2} \frac{1}{t'} + O\left(\frac{\ln t'}{t'^{2}}\right).$$

The leading term of the upper bound on the convergence rate is independent of delay $\tau$. 
%
In fact we may allow $\tau$ to be a monotonic increasing function of $t$ such that for $t$ large enough,
\begin{equation}
\frac{2 L \alpha D}{\mu}\leq \tau(t)\leq \sqrt{\frac{t}{\ln t}  \cdot \left( 1 - \frac{1}{\ln t} \right)}. \label{tau1}
\end{equation}
This will make 
$$E = \max\{ 2\tau(t), \frac{4 L \alpha D}{\mu}\}$$
also a function of $t$. In terms of asymptotic in $t$, if  $12\leq \alpha_t\leq \alpha$, then the leading term of the convergence rate for such $\tau$ as a function of $t$ does not change while the second order terms increase to $O(\frac{1}{t\ln t})$. It turns out that if $\tau(t)$ is set to meet the upper bound in (\ref{tau1}), then for large enough 
$$t\geq T_0 =  \exp[ 2\sqrt{\Delta}(1+\frac{(L+\mu)\alpha}{\mu})]$$
the constants of all the asymptotic higher order terms that contain $\tau(t)$ are such that the concrete values of all these terms are at most the leading term of the convergence rate.\footnote{For completeness, the leading term is also independent of $\|w_0-w_*\|^2$ -- it turns out that for $t \geq T_1 = \frac{\mu^2}{\alpha^2 N D}\|w_0-w_*\|^2$ the higher order term that contains $\|w_0-w_*\|^2$ is at most the leading term.}
 So, for the worst-case delay function $\tau(t)$ we have for $t\geq T_0$ that the convergence rate is at most 2 times 
 $\frac{4\alpha^2D N}{\mu^2} \frac{t}{(t + E - 1)^2}$. In practice, we expect to see this behavior for much smaller $T_0$. Also, if $\tau(t)$ is a much less increasing function in $t$, we expect to be able to prove a much smaller $T_0$.
The upper bounds on the convergence rates above are tight as shown  in \citep{nguyen2019tight}.
 
If $t$ is also large enough such that $2\tau(t)\geq \frac{4L\alpha D}{\mu}$ (this already happens for relatively small $t$), then $E$ as a function of $t$ is equal to $2\tau(t)$. Thus gives a convergence rate of 
$$\leq 2\frac{4\alpha^2D N}{\mu^2} \frac{t}{(t + 2\tau(t) - 1)^2}.$$

%
%

The results  enumerated above are summarized for $D=1$ in Lemma \ref{lemsc}:

\begin{lem} \label{lemsc} \citep{nguyen2018sgd,nguyen2018new,nguyen2019tight}
Let $f$ be $L$-smooth, convex, and let the objective function $F(w)=\mathbb{E}_{\xi\sim {\cal D}}[f(w;\xi)]$ be $\mu$-strongly convex with finite $N = 2 \mathbb{E}[ \|\nabla f(w_{*}; \xi)\|^2 ]$ where $w_{*} = \arg \min_w F(w)$.
Then 
there exists a 
step size sequence $\{\eta_t = \frac{\alpha_t}{\mu (t+2\max \{\tau(t), 2L\alpha /\mu\})} \}$ with $12\leq \alpha_t\leq \alpha$ and delay function $\tau(t)$ with $\tau(t)\leq \sqrt{\frac{t}{\ln t}  \cdot \left( 1 - \frac{1}{\ln t} \right)}$ for $t$ large enough such that there exists a constant $T_0$ with the property that for $t\geq T_0$, the convergence rates $ \mathbb{E}[\|\hat{w}_{t} - w_* \|^2]$ and $\mathbb{E}[\|w_{t} - w_* \|^2]$ are at most  $\frac{8\alpha^2 N}{\mu^2} \frac{t}{(t + 2\max \{\tau(t), 2L\alpha /\mu\} - 1)^2}$. These convergence rates are optimal up to a small constant factor (not depending on the dimension of the model).
\end{lem}

The conditions of Lemma \ref{lemsample} are satisfied for $g=2$ and $\gamma(z) = 4\ln(z)$. Let 
$$ M_1=\max \{d+1, 2L\alpha /\mu, 
\frac{1}{2}\lceil \frac{m+1}{16(d+1)^2} \frac{1}{ \ln(\frac{m+1}{2(d+1)})} \rceil
\}$$ 
and 
$$M_0=((m+1) \frac{g-1}{g})^{g/(g-1)} = \frac{(m+1)^2}{4}.
$$
This defines
$$\tau(t) = M_1 + (\frac{t+M_0}{\gamma(t+M_0)})^{1/g}=M_1+ \sqrt{\frac{t+M_0}{ 4 \ln(t+M_0)}}.$$
For $t\geq \max\{M_0, e^2\}$,
the square root is 
$$\leq \sqrt{\frac{t}{\ln t}  \cdot \left( 1 - \frac{1}{\ln t} \right)}$$
because $1-1/\ln t \geq 1/2$ and $(t+M_0)/\ln(t+M_0) \leq (t+M_0)/\ln t \leq 2t/\ln t$. Similar calculations show that $\tau(t)$ satisfies the bound in Lemma \ref{lemsc} for $t$ large enough.
%
%
Hence, we can apply Lemma \ref{lemsc}. Here,  notice that $\tau(t)\geq 2L\alpha/\mu$, hence, $E_t = 2\max \{\tau(t), 2L\alpha/\mu\}= 2\tau(t)$. 

We use Lemma \ref{lemsample} to obtain a concrete sample size sequence $\{s_i\}$.
This leads to an increasing sample size sequence defined by
$$ s_i = \lceil \frac{m+i+1}{16(d+1)^2} \frac{1}{ \ln(\frac{m+i+1}{2(d+1)})} \rceil =
O(\frac{i}{\ln i}).
$$
(For example, $s_0=50$  corresponds to $m=2(d+1)\cdot 1450$.)

We use Lemma \ref{lemstep} to find the round step size sequence $\{\bar{\eta}_i\}$. Notice that we need to choose $a_0=12$ and $\alpha=a_1\leq 3\cdot a_0=36$ in Lemma  \ref{lemstep}. Also notice that $E_0=2\tau(0)\geq s_0$.

The inequality $\bar{E}_{i+1} \leq 2 \bar{E}_i$ follows from (remember $E_t=2\tau(t)$)
$$ \sqrt{\frac{\sum_{j=0}^{i} s_j +M_0}{4 \ln ( \sum_{j=0}^{i} s_j +M_0) }}
\leq 2 \sqrt{\frac{\sum_{j=0}^{i-1} s_j +M_0}{4 \ln ( \sum_{j=0}^{i-1} s_j +M_0) }}.
$$
Notice that (\ref{sieq}) for $g=2$  implies $s_{i} -1 \leq 2s_{i-1}$. Since $s_{i-1}\geq 1$ we have $s_i \leq 3 s_{i-1}$. We derive
\begin{eqnarray*}
\sqrt{\frac{\sum_{j=0}^{i} s_j +M_0}{4 \ln ( \sum_{j=0}^{i} s_j +M_0) }}
&\leq & 
\sqrt{\frac{\sum_{j=0}^{i} s_j +M_0}{4 \ln ( \sum_{j=0}^{i-1} s_j +M_0) }}
\leq
\sqrt{\frac{\sum_{j=0}^{i-1} s_j +3 s_{i-1}+ M_0}{4 \ln ( \sum_{j=0}^{i-1} s_j +M_0) }} \\
&\leq &
\sqrt{\frac{4(\sum_{j=0}^{i-1} s_j + M_0)}{4 \ln ( \sum_{j=0}^{i-1} s_j +M_0) }}
=
2 \sqrt{\frac{\sum_{j=0}^{i-1} s_j +M_0}{4 \ln ( \sum_{j=0}^{i-1} s_j +M_0) }}.
\end{eqnarray*}
It remains to show that $\bar{E}_1\leq 2 E_0=2 \bar{E}_{-1}$. This follows from a similar derivation as the one above if we can show that $s_0\leq 3 M_0$. The latter is indeed true because $\frac{1}{16(d+1)^2}\leq \frac{1}{16} \leq \frac{3}{4} \leq 3 \frac{m+1}{4}$.




Now we are ready to apply Lemma \ref{lemstep} from which we obtain  diminishing round step size sequence defined by
$$
\bar{\eta}_i = \frac{12}{\mu} \cdot \frac{1}
{\sum_{j=0}^{i-1} s_j + 
2M_1 
+
\sqrt{\frac{(m+1)^2/4+\sum_{j=0}^{i-1} s_j}{\ln((m+1)^2/4+ \sum_{j=0}^{i-1} s_j)}}}
%
= O(\frac{\ln i}{i^2}).
$$
By Lemma \ref{lemsc} there exists a constant $T_0$ such that, for $t\geq T_0$, the convergence rates $ \mathbb{E}[\|\hat{w}_{t} - w_* \|^2]$ and $\mathbb{E}[\|w_{t} - w_* \|^2]$ are at most  
$$\frac{8\cdot 36^2 \cdot N}{\mu^2} \frac{t}{(t + 
2M_1
+
\sqrt{\frac{(m+1)^2/4+t}{\ln((m+1)^2/4+ t)}} -1)^2}
= \frac{8\cdot 36^2 \cdot N}{\mu^2} \frac{1}{t} ( 1- O(1/t)).$$
The best  convergence rate is at least $\frac{1}{2} \frac{N}{\mu^2} \frac{1}{t} (1- O((\ln t)/t))$ as shown  in \citep{nguyen2019tight}. This means that we are at least within a dimension less factor $\leq 2 \cdot 8 \cdot 36^2 = 20736$ of the best possible convergence rate.

All the above proves Theorem \ref{thmsc}:


\begin{thm} \label{thmsc}
Let $f$ be $L$-smooth, convex, and let the objective function $F(w)=\mathbb{E}_{\xi\sim {\cal D}}[f(w;\xi)]$ be $\mu$-strongly convex with finite $N = 2 \mathbb{E}[ \|\nabla f(w_{*}; \xi)\|^2 ]$ where $w_{*} = \arg \min_w F(w)$. Let $\{s_i\}$ be the increasing sample size sequence defined by
$$ 
s_i = \lceil \frac{m+i+1}{16(d+1)^2} \frac{1}{ \ln(\frac{m+i+1}{2(d+1)})} \rceil =
\Theta(\frac{i}{\ln i}).
%
$$
and let $\{\bar{\eta}_i\}$ be the diminishing round step size sequence defined by
$$
\bar{\eta}_i = \frac{12}{\mu} \cdot \frac{1}
{\sum_{j=0}^{i-1} s_j + 
2M_1
+
\sqrt{\frac{(m+1)^2/4+\sum_{j=0}^{i-1} s_j}{\ln((m+1)^2/4+ \sum_{j=0}^{i-1} s_j)}}}
%
= O(\frac{\ln i}{i^2}),
$$
where
$$ M_1=\max \{d+1, 2L\alpha /\mu, 
\frac{1}{2}\lceil \frac{m+1}{16(d+1)^2} \frac{1}{ \ln(\frac{m+1}{2(d+1)})} \rceil
\}.$$
Then, there exists a constant $T_0$ such that, for $t\geq T_0$, the convergence rates $ \mathbb{E}[\|\hat{w}_{t} - w_* \|^2]$ and $\mathbb{E}[\|w_{t} - w_* \|^2]$ are at most  
$$\frac{8\cdot 36^2 \cdot N}{\mu^2} \frac{t}{(t + 
2M_1 +
\sqrt{\frac{(m+1)^2/4+t}{\ln((m+1)^2/4+ t)}} -1)^2}
= \frac{8\cdot 36^2 \cdot N}{\mu^2} \frac{1}{t}(1- O(1/t)).$$
This is optimal up to a constant factor $\leq 16\cdot 36^2$ (independent of any parameters like $L$, $\mu$, sparsity, or dimension of the model).
\end{thm}

A derivative  of this theorem is stated in the main body as Theorem \ref{thmsc2}.





\subsection{Plain convex and non-convex problems} \label{app-nonconvex}

For an objective function $F$ (as defined in (\ref{eqObj})), we are generally interested in the expected convergence rate
\begin{equation*}
Y^{(F)}_t =  \mathbb{E}[ F(w_t)-F_{*}],
\end{equation*}
 where $F_{*} = F(w_{*})$ for a global minimum $w_{*}$ (and the expectation is over the randomness used in the recursive computation of $w_t$ by the probabilistic optimization algorithm of our choice). 
This implicitly assumes that there exists a global minimum $w_{*}$, i.e., $\mathcal{W}^*=\{ w_{*} \in \mathbb{R}^d \ : \ \forall_{w\in \mathbb{R}^d} \ F(w_{*})\leq F(w)\} $ defined as the set  of all $w_{*}$ that minimize $F(\cdot)$ is non-empty. 
Notice that $\mathcal{W}^*$ can have multiple global minima even for convex problems. 

For convex problems a more suitable definition for $Y_t$ is the averaged expected convergence rate defined as 
\begin{equation*}
Y^{(A)}_t = \frac{1}{t+1} \sum_{i=t+1}^{2t} \mathbb{E}[ F(w_t)-F_{*} ]
\end{equation*}
and  for strongly convex objective functions we may use 
\begin{equation*}
Y^{(w)}_t = \mathbb{E}[ \mbox{inf} \{ \Vert w_t - w_{*}\Vert ^2 \ : \ w_{*} \in \mathcal{W}^* \} ].
\end{equation*}
$\omega$-Convex objective functions define a range of functions between plain convex and strong convex for which both convergence rate definitions can be computed for diminishing step sizes~\citep{van2019characterization}. From \citep{van2019characterization} we have that an objective function with 'curvature' $h\in [0,1]$ (where $h=0$ represents plain convex and $h=1$ represents strong convex) achieves convergence rates $Y_t^{(w)}=O(t^{-h/(2-h)})$ and $Y_t^{(A)}=O(t^{-1/(2-h)})$ for diminishing step sizes $\eta_t = O(t^{-1/(2-h)})$. For this reason, when we study plain convex problems, we use  diminishing step size sequence $O(t^{-1/2})$ and we experiment with different increasing sample size sequences to determine into what extent asynchronous SGD or Hogwild! is robust against delays. Since strong convex problems have best convergence and therefore best robustness against delays, we expect a suitable increasing sample size sequence $O(i^p)$ for some $0\leq p\leq 1$.

For non-convex (e.g., DNN) problems we generally use the averaged expected squared norm of the objective function gradients:
 \begin{equation*}
 Y^{(\nabla)}_t=\frac{1}{t+1} \sum_{j=0}^t  \mathbb{E}[\Vert \nabla F(w_j) \Vert ^2].
 \end{equation*}
This type of convergence rate analyses convergence to any (good or bad)  local minimum and does not exclude saddle-points. 
There may not even exist a global minimum (i.e., ${\cal W}^*$ is empty) in that some entries in the weight vector $w_t$ may tend to $\pm \infty$ -- nevertheless, there may still exist a value $F_{*} = \mbox{sup} \{ F_{low} \ :  F_{low}\leq F(w), \  \forall{w\in \mathbb{R}^d}\}$ which can be thought of as the value of the global minimum if we include limits to infinite points. In practice a diminishing step size sequence of $O(t^{-1/2})$ (as in the plain convex case)  gives good results and this is what is used in our experiments.

\section{Differential privacy proofs}
\label{appDP}

\subsection{Definitions}

We base our proofs on the framework and theory presented in \citep{abadi2016deep}. In order to be on the same page we repeat and cite word for word their definitions:

For neighboring databases $d$ and $d'$, a
mechanism ${\cal M}$, auxiliary input $\texttt{aux}$, and an outcome $o$,
define the privacy loss at $o$ as
$$ c(o;{\cal M}, \texttt{aux}, d, d')
= \ln
\frac{Pr[{\cal M}(\texttt{aux}, d) = o]}
{Pr[{\cal M}(\texttt{aux}, d') = o]}.
$$
For a given mechanism {\cal M}, we define the $\lambda$-th moment
$\alpha_{{\cal M}}(\lambda ; \texttt{aux}, d, d')$ as the log of the moment generating function evaluated at the value $\lambda$:
$$
\alpha_{{\cal M}}(\lambda ; \texttt{aux}, d, d')
=
\ln \textbf{E}_{o\sim {\cal M}(\texttt{aux},d)}[\exp(\lambda \cdot c(o;{\cal M}, \texttt{aux}, d, d'))].
$$
We define
$$ \alpha_{{\cal M}}(\lambda) = \max_{\texttt{aux},d,d'} 
\alpha_{{\cal M}}(\lambda ; \texttt{aux}, d, d')
$$
where the maximum is taken over all possible $\texttt{aux}$ and all
the neighboring databases $d$ and $d'$.


We first take Lemma 3 from \citep{abadi2016deep} and make explicit their order term $O(q^3\lambda^3/\sigma^3)$ with $q=s_{i,c}$ and $\sigma=\sigma_i$ in our notation. The lemma considers as mechanism ${\cal M}$ the $i$-th round of gradient updates and we abbreviate $\alpha_{{\cal M}}(\lambda)$ by $\alpha_i(\lambda)$. The auxiliary input of the mechanism at round $i$ includes all the output of the mechanisms of previous rounds (as in \citep{abadi2016deep}).

To be precise, 
$${\cal M}(\texttt{aux},D) = \sum_{h=0}^{s_{i,c}-1} \nabla f(\hat{w}_h,\xi_h) +{\cal N}(0,C^2\sigma_i^2\textbf{I})$$ where $\hat{w}_{h+1}=\hat{w}_h-\eta_h \nabla f(\hat{w}_h,\xi_h)$ with $\hat{w}_h$ overwritten by a new broadcast global model $\hat{v}$ minus $\eta_h$ times ${\cal M}$'s updates computed so far if $\hat{v}$ just arrived (see \cal{ISRReceive}), and where $\xi_h$ are drawn from training data $D={\cal D}_c$. In order for ${\cal M}$ to be able to compute its output, it needs to know the global models received  in round $i$ and it needs to know the starting local model $\hat{w}_0$. To make sure ${\cal M}$ has all this information,
$\texttt{aux}$ represents  the collection of all outputs generated by the mechanisms of previous rounds $<i$ together with the global models received in round $i$ itself. The previous outputs represent updates, which when multiplied with the appropriate step sizes and subtracted from the default global model $\hat{v}_0$ yields the starting local model $\hat{w}_0$ of round $i$. 

In the next subsection we will use the framework of \citep{abadi2016deep} and apply its composition theory to derive bounds on the privacy budget $(\epsilon,\delta)$ for the whole computation consisting of $T$ rounds that reveal the outputs of the mechanisms for these $T$ rounds as described above.




We remind the reader that $s_{i,c}/N_c$ is the probability to select a sample from a sample set (batch) of size $s_{i,c}$ out of a training data set ${\cal D}_c$ of size $N_c=|{\cal D}_c|$;  $\sigma_i$ corresponds to the ${\cal N}(0,C^2\sigma_i^2 \textbf{I})$ noise added to the batch gradient computation in round $i$ (see the mechanism described above).

\begin{lem} \label{lemthree} 
Assume a constant $r_0<1$ 
such that $s_{i,c}/N_c\leq r_0/\sigma_i$.
Suppose that $\lambda$ is a positive integer with
$$ \lambda \leq \sigma_i^2 \ln \frac{N_c}{s_{i,c} \sigma_i}$$
and define
$$
U_0(\lambda) = \frac{2  \sqrt{\lambda r_0/\sigma_i}     
}
{
 \sigma_i -r_0
} \mbox{ and }
U_1(\lambda) = \frac{ 2e \sqrt{\lambda r_0/\sigma_i}   
}
{
 (\sigma_i-r_0) \sigma_i
}.
$$
Suppose $U_0(\lambda)\leq u_0<1$ and $U_1(\lambda)\leq u_1<1$ for some constants $u_0$ and $u_1$.
Define
\begin{eqnarray*}
r &=& r_0 \cdot 2^3 (\frac{1}{1-u_0} + \frac{ 1}{1-u_1} \frac{e^3}{\sigma_i^3})\exp(3/\sigma_i^2).
\end{eqnarray*}

Then,
\begin{equation*}
\alpha_i(\lambda) \leq  \frac{s_{i,c}^2 \lambda (\lambda+1)}{N_c(N_c-s_{i,c}) \sigma_i^2} +\frac{r}{r_0} \cdot \frac{s_{i,c}^3\lambda^2(\lambda+1)}{N_c (N_c-s_{i,c})^2\sigma_i^3}.
\end{equation*}
\end{lem}





\vspace{.5cm}

\noindent
{\bf Proof.} 
The start of the proof of Lemma 3 in \citep{abadi2016deep} implicitly uses the proof of Theorem A.1 in \citep{dwork2014algorithmic}, which up to formula (A.2) shows how the 1-dimensional case translates into a privacy loss that corresponds to the 1-dimensional problem defined by $\mu_0$ and $\mu_1$ in the proof of Lemma 3 in \citep{abadi2016deep}, and which shows at the end of the proof of Theorem A.1 (p. 268 \citep{dwork2014algorithmic}) how the multi-dimensional problem transforms into the 1-dimensional problem. In Theorem A.1, $f(D)+{\cal N}(0,\sigma^2\textbf{I})$ represents the general (random) mechanism ${\cal M}(D)$, which for Lemma 3 in  \citep{abadi2016deep}'s notation 
should be interpreted as the batch computation ${\cal M}(D)=\sum_{i\in J} f(d_i) +{\cal N}(0,\sigma^2\textbf{I})$ for a random sample $J$. 

In our scenario the batch computation recursively adds gradients $\nabla f(\hat{w},\xi)$ and is multi-dimensional. However, within the batch computation $\hat{w}$ is modified. In our notation we have ${\cal M}(D) = \sum_{h=0}^{s_{i,c}-1} \nabla f(\hat{w}_h,\xi_h) +{\cal N}(0,C^2\sigma_i^2\textbf{I})$ as described above.
This means that $\nabla f(.)$ is not only a function of $\xi$ but is also a function of prior $\xi$'s within the batch.

Nevertheless Lemma 3 in \citep{abadi2016deep} still applies because the proof of Theorem A.1 in \citep{dwork2014algorithmic} applies to any randomized mechanism, also to our recursively defined batch computation. This translates our privacy loss to the 1-dimensional problem defined by $\mu_0 \sim {\cal N}(0,C^2\sigma^2)$ and $\mu_1 \sim {\cal N}(0,C^2\sigma^2)$ for $\|\nabla f(.,.)\|_2\leq C$ as in the proof of Lemma 3 (which after normalization with respect to $C$ gives the formulation of Lemma 3 in \citep{abadi2016deep} for $C=1$). 

The remainder of the proof of Lemma 3 analyses $\mu_0$ and the mix $\mu=(1-q)\mu_0+q\mu_1$ leading to  bounds for the expectations (3) and (4) in \citep{abadi2016deep} which only depend on $\mu_0$ and $\mu_1$. Here, $q$ is the probability of having a special data sample $\xi$ (written as $d_n$ in the proof of Lemma 3) in the batch. In our algorithm $q=s_{i,c}/N_c$. So, we may adopt the statement of Lemma 3 and conclude for the $i$-th batch computation
\begin{equation*}
\alpha_i(\lambda) \leq  \frac{s_{i,c}^2 \lambda (\lambda+1)}{N_c(N_c-s_{i,c}) \sigma_i^2} +O(\frac{s_{i,c}^3\lambda^3}{N_c^3\sigma_i^3}).
\end{equation*}

In order to find an exact expression for the higher order term we look into the details of Lemma 3 of \citep{abadi2016deep}.
It computes an upper bound for the binomial tail
$$ 
\sum_{t=3}^{\lambda+1} {\lambda +1 \choose t} \mathbb{E}_{z\sim \nu_1}[((\nu_0(z)-\nu_1(z))/\nu_1(z))^t],
$$
where
\begin{eqnarray*}
&& \mathbb{E}_{z\sim \nu_1}[((\nu_0(z)-\nu_1(z))/\nu_1(z))^t]
\\
&\leq &
\frac{(2q)^t(t-1)!!}{2 (1-q)^{t-1} \sigma^t} +
\frac{q^t}{(1-q)^t\sigma^{2t}} +
 \frac{(2q)^t \exp((t^2-t)/(2\sigma^2)) (\sigma^t (t-1)!! + t^t)
}
{
2 (1-q)^{t-1} \sigma^{2t}
} \\
&=&
 \frac{(2q)^t (t-1)!!  ( 1+ \exp((t^2-t)/(2\sigma^2))  )
}
{
2 (1-q)^{t-1} \sigma^{t}
} 
+
 \frac{ q^t (1+ (1-q) 2^t\exp((t^2-t)/(2\sigma^2)) t^t )
}
{
2 (1-q)^{t} \sigma^{2t}
}.
\end{eqnarray*}
%
%
%

After multiplying with the upper bound for
$$ {\lambda +1 \choose t} \leq \frac{\lambda+1}{\lambda} \frac{\lambda^{t}}{t!}$$
and noticing that $(t-1)!!/t!\leq 1$ and $t^t/t! \leq e^t$
we get the addition of the following two terms
\begin{eqnarray*}
&& \frac{\lambda+1}{\lambda}\frac{\lambda^t (2q)^t   ( \frac{(t-2)!!}{t!}  + \exp((t^2-t)/(2\sigma^2))  )
}
{
2 (1-q)^{t-1} \sigma^{t}
} 
\\
&& \hspace{2cm} +
\frac{\lambda+1}{\lambda} \frac{ \lambda^t q^t (\frac{1}{t!}+ (1-q) 2^t\exp((t^2-t)/(2\sigma^2)) e^t )
}
{
2 (1-q)^{t} \sigma^{2t}
}.
\end{eqnarray*}
This is equal to
\begin{eqnarray*}
&& \frac{\lambda+1}{\lambda} \frac{\lambda^t (2q)^t   ( \frac{(t-1)!!}{t!}+ (\exp((t-1)/(2\sigma^2)))^t  )
}
{
2 (1-q)^{t-1} \sigma^{t}
} \\
&& \hspace{2cm} +
\frac{\lambda+1}{\lambda} \frac{ \lambda^t q^t (\frac{(1-q)^{-1}}{t!}+  (2\exp(1+(t-1)/(2\sigma^2)))^t )
}
{
2 (1-q)^{t-1} \sigma^{2t}
}.
\end{eqnarray*}
We drop the $O((t-1)!!/t!)$ and $O(1/t!)$ terms
in order to coarsely bound this as\footnote{In order to have a proper, but less tight, bound we can multiply the coarse bound by $1+(1-q)^{-1}\leq 1+ 1/(1-r_0/\sigma)
= 1+\sigma/(\sigma-r_0)
= 2 + r_0/(\sigma-r_0)$.}
\begin{eqnarray}
&&
(1-q)\frac{\lambda+1}{\lambda}(\frac{\lambda 2q    \exp((t-1)/(2\sigma^2))  
}
{
 (1-q) \sigma
})^t \nonumber
\\
&& \hspace{2cm} +
(1-q) \frac{\lambda+1}{\lambda}(\frac{ \lambda q   2\exp(1+(t-1)/(2\sigma^2)) 
}
{
 (1-q) \sigma^{2}
})^t. \label{eqterms}
\end{eqnarray}
We notice that by using $t\leq \lambda+1$, $\lambda/\sigma^2 \leq \ln (1/(q\lambda))$ (assumption), and $q=s_{i,c}/N_c \leq r_0/\sigma$ we obtain
\begin{eqnarray*} 
\frac{\lambda 2q    \exp((t-1)/(2\sigma^2))  
}
{
 (1-q) \sigma
} &\leq &
\frac{\lambda 2q    \exp(\lambda/(2\sigma^2))  
}
{
 (1-q) \sigma
}
\leq 
\frac{2  \sqrt{\lambda q}     
}
{
 (1-q) \sigma
} = 
\frac{2  \sqrt{\lambda r_0/\sigma}     
}
{
 \sigma -r_0
} =U_0(\lambda)
\end{eqnarray*}
and
\begin{eqnarray*}
\frac{ \lambda q   2\exp(1+(t-1)/(2\sigma^2)) 
}
{
 (1-q) \sigma^{2}
}
&\leq &
\frac{ \lambda q   2 e\exp(\lambda/(2\sigma^2)) 
}
{
 (1-q) \sigma^{2}
}
\leq
\frac{ 2e \sqrt{\lambda q}   
}
{
 (1-q) \sigma^{2}
} = 
\frac{ 2e \sqrt{\lambda r_0/\sigma}   
}
{
 (\sigma -r_0) \sigma
} =U_1(\lambda).
\end{eqnarray*}



Together with our assumption on $U_0(\lambda)$ and $U_1(\lambda)$, this means that the binomial tail is upper bounded by the $t=3$ terms in (\ref{eqterms}) times
$$ \sum_{j=0}^\infty U_0(\lambda)^j = \frac{1}
{1-U_0(\lambda)}\leq \frac{1}{1-u_0} \mbox{ and } \sum_{j=0}^\infty U_1(\lambda)^j = \frac{1}{1-U_1(\lambda)}\leq \frac{1}{1-u_1}$$
respectively.
This yields the upper bound
\begin{eqnarray*}
&& \frac{1}
{1-u_0} (1-q)\frac{\lambda+1}{\lambda}(\frac{\lambda 2q    \exp(1/\sigma^2)  
}
{
 (1-q) \sigma
})^3 \\
&& \hspace{2cm} +
 \frac{1}{1-u_1}
(1-q)\frac{\lambda+1}{\lambda}
(\frac{ \lambda q   2\exp(1+1/\sigma^2) 
}
{
 (1-q) \sigma^{2}
})^3 \\
&\leq &
( \frac{1}
{1-u_0}
  2^3 \exp(3/\sigma^2) 
+  \frac{1}{1-u_1}
\frac{ 2^3 \exp(3+3/\sigma^2)}{\sigma^3}
)
\cdot 
\frac{ \lambda^2 (\lambda+1) q^3}{(1-q)^2\sigma^3} .
\end{eqnarray*}
By the definition of $r$, we obtain the bound 
%
\begin{eqnarray*}
&\leq &
\frac{r}{r_0}
\cdot
\frac{\lambda^2(1+\lambda) q^3}{(1-q)^2\sigma^3},
\end{eqnarray*}
which finalizes the proof.

\subsection{Proof of Theorem \ref{thm1}}
\label{app-dp1}

The proof of our first theorem follows the line of thinking if the proof of Theorem 1 in \citep{abadi2016deep}. Our theorem applies to varying sample/batch sizes and for this reason introduces moments $\hat{S}_j$. Our theorem explicitly defines the constant used in the lower bound of $\sigma$ -- this is important for proving our second (main) theorem in the next subsection.

%
Theorem \ref{thm1} assumes $\sigma = \sigma_i$ for all rounds $i$; 
constant $r_0\leq 1/e$ 
such that $s_{i,c}/N_c\leq r_0/\sigma$;  constant
\begin{eqnarray}
r &=& r_0 \cdot 2^3 (\frac{1}{1-u_0} + \frac{ 1}{1-u_1} \frac{e^3}{\sigma^3})\exp(3/\sigma^2), \label{constr}
\end{eqnarray}
where
$$ u_0 = \frac{2  \sqrt{r_0\sigma}     
}
{
 \sigma -r_0
} \mbox{ and } u_1 = \frac{ 2e \sqrt{r_0 \sigma} 
}
{
 (\sigma-r_0) \sigma
}
$$
are both assumed $<1$. 


For $j=1,2,3$ we define\footnote{$s_{i,c}^j$ denotes the $j$-th power $(s_{i,c})^j$.}
$$\hat{S}_j = \frac{1}{T} \sum_{i=1}^T \frac{s_{i,c}^j}{N_c(N_c-s_{i,c})^{j-1}}
\mbox{ with } \frac{\hat{S}_1\hat{S}_3}{\hat{S}_2^2}\leq \rho, \
 \frac{\hat{S}_1^2}{\hat{S}_2}\leq \hat{\rho}.
$$
Based on these constants we define
$$c(x) = \min \{ \frac{\sqrt{2r\rho x+1} -1}{r \rho x}, \frac{2}{\hat{\rho} x}  
\}.
$$

Let $\epsilon = c_1 T \hat{S}_1^2$.
We want to prove Algorithm \ref{alg:DP} is $(\epsilon, \delta)$-differentially private if
$$\sigma \geq \frac{2}{\sqrt{c_0}} \frac{\sqrt{ \hat{S}_2 T \ln (1/\delta) }}{\epsilon}  \mbox{ where } c_0=c(c_1).
$$

\vspace{.5cm}

\noindent
{\bf Proof.} 
For $j=1,2,3$, we define 
$$ S_j = \sum_{i=1}^T \frac{s_{i,c}^j}{N_c (N_c-s_{i,c})^{j-1}\sigma_i^j} \mbox{ and } 
S'_j = \frac{1}{T} \sum_{i=1}^T \frac{s_{i,c}^j \sigma_i^j}{N_c (N_c-s_{i,c})^{j-1}}.
$$
(Notice that $S'_1\leq r_0$.)
Translating Lemma \ref{lemthree} in this notation yields (we will verify the requirement/assumptions of Lemma \ref{lemthree} on the fly below)
$$\sum_{i=1}^T \alpha_i(\lambda) \leq S_2 \lambda (\lambda+1) + \frac{r}{r_0} S_3 \lambda^2(\lambda+1).$$

The composition Theorem 2 in \citep{abadi2016deep} shows that  our algorithm for client $c$ is $(\epsilon, \delta)$-differentially private for
\begin{eqnarray*}
\delta &\geq & \min_{\lambda}\exp(\sum_{i=1}^T \alpha_i(\lambda)-\lambda \epsilon),
\end{eqnarray*}
where $T$ indicates the total number of batch computations and the minimum is over positive integers $\lambda$.
Similar to their proof we choose $\lambda$ such that
$$ S_2 \lambda (\lambda+1) + \frac{r}{r_0} S_3 \lambda^2(\lambda+1) -\lambda \epsilon \leq -\lambda \epsilon/2$$
(this implies that we can choose $\delta$ as small as $\exp(-\lambda\epsilon/2)$).
%
%
 This leads to the inequality
 $$ S_2 (\lambda+1) + \frac{r}{r_0} S_3 \lambda (1+\lambda) \leq \epsilon/2,$$
 which is implied by
 $$ (\lambda+1) ( 1+ \frac{r}{r_0} \frac{S_3}{S_2} \lambda) \leq \frac{\epsilon}{2S_2}.
 $$
 This is in turn implied by
 $$ \lambda +1 \leq c_0 \frac{\epsilon}{2S_2} $$
 if
 $$
 c_0 \frac{\epsilon}{2S_2} 
 ( 1+ \frac{r}{r_0} \frac{S_3}{S_2} c_0 \frac{\epsilon}{2S_2} ) \leq \frac{\epsilon}{2S_2},
 $$
 or equivalently,
 $$
 c_0 
 ( 1+ \frac{r}{2r_0} \cdot c_0 \cdot \frac{S_3}{S_2^2}\epsilon ) \leq 1.
 $$
 We use
 $$\epsilon = c_1\cdot T \hat{S}_1^2 = c_1 \cdot S_1 S'_1
 $$
 (for constant $\sigma_i=\sigma$).
 This translates our requirements into
 $$ \lambda +1 \leq \frac{c_0c_1}{2} \frac{S_1S'_1}{S_2} \mbox{ and } c_0(1+\frac{r}{2r_0}\cdot c_0 c_1 \frac{S_1S_3}{S_2^2}S'_1) \leq 1. $$
 
 Since we assume
 $$\frac{S_1S_3}{S_2^2}= \frac{\hat{S}_1\hat{S}_3}{\hat{S}_2^2}\leq \rho $$
 and since we know that $S'_1\leq r_0$,
 the  requirement $c_0(1+\frac{r}{2r_0}\cdot c_0 c_1 \frac{S_1S_3}{S_2^2}S'_1) \leq 1$ is implied by 
 $$ c_0(1+\frac{r \rho}{2} \cdot c_0 c_1 ) \leq 1, $$
 or equivalently
 $$ c_1 \leq \frac{1-c_0}{\frac{r \rho}{2}  c_0^2}.$$
 That is, any
 $$\epsilon \leq \frac{1-c_0}{\frac{r \rho}{2}  c_0^2} S_1S'_1$$
 has a fitting constant $c_1$ that satisfies the conditions analysed so far.
 

 Also notice that for constant $\sigma_i=\sigma$ we have $S'_1=S_1\sigma^2/T$. Together with 
 $$\frac{S_1^2}{S_2}= \frac{\hat{S}_1^2}{\hat{S}_2} T\leq \hat{\rho} T
 $$
 we obtain
 $$ \lambda +1 \leq 
 \frac{c_0c_1}{2} \frac{S_1S'_1}{S_2}
 \leq 
 \frac{c_0c_1}{2} \hat{\rho}
 \sigma^2.
$$
Generally, if $c_1\leq 2/(\hat{\rho}c_0)$, then (a) the condition $\lambda \leq \sigma_i^2 \ln \frac{N_c}{s_{i,c}\sigma_i}$ is satisfied (by assumption, $\frac{N_c}{s_{i,c}\sigma_i}\geq 1/r_0\geq e$), and (b) $\lambda\leq \sigma^2$ which implies that, for our choice of $u_0$ and $u_1$ in this theorem, $U_0(\lambda)\leq u_0$ and $U_1(\lambda)\leq u_1$ as defined in Lemma \ref{lemthree}. This implies that Lemma \ref{lemthree} is indeed applicable.

For the above reasons we strengthen the requirement on $\epsilon$ to
$$\epsilon \leq \min \{\frac{1-c_0}{\frac{r \rho}{2}  c_0^2}, \frac{2}{\hat{\rho} c_0} \} \cdot S_1S'_1.$$
For constant $\sigma_i=\sigma$, we have
$$S_1S'_1= T \hat{S}_1^2,$$
hence
$$ \epsilon \leq \min \{\frac{1-c_0}{\frac{r \rho}{2} \rho c_0^2}, \frac{2}{\hat{\rho} c_0} \} \cdot T\hat{S}_1^2.$$


 The condition $\exp(-\lambda \epsilon /2) \leq \delta $ reduces to
$$ \ln(1/\delta) \leq \frac{\lambda \epsilon}{2} \leq \frac{c_0 c_1^2}{4} 
\frac{S_1^2 {S'}_1^2}{S_2} = \frac{c_0}{4S_2}\epsilon^2.
$$
For constant $\sigma_i=\sigma$ we have $S_2=\hat{S}_2 T/\sigma^2$.
This leads to an inequality for $\sigma$:
$$
\sigma \geq  \frac{2}{\sqrt{c_0}} \frac{\sqrt{\hat{S}_2} \sqrt{T \ln (1/\delta)}}{\epsilon}.
$$

Notice that this corresponds exactly to Theorem 1 in \citep{abadi2016deep} where all $s_{i,c}$ are constant implying $\sqrt{\hat{S}_2}=q$. We are interested in a slightly different formulation:

 Given
$$ c_1= \min \{\frac{1-c_0}{\frac{r\rho}{2} c_0^2}, \frac{2}{\hat{\rho} c_0} \} $$
what is the maximum possible $c_0$ (which minimizes $\sigma$ implying more fast convergence to an accurate solution). We need to satisfy $c_0\leq 2/(\hat{\rho}c_1)$ and
$$ \frac{r\rho}{2} c_1 c_0^2 +c_0 -1 \leq 0,$$
that is,
$$( c_0 +1/(r\rho c_1))^2 \leq 1/(\frac{r\rho}{2} c_1)+ 1/(r\rho c_1)^2,$$
or
$$ c_0 \leq \sqrt{1/(\frac{r\rho}{2} c_1) + 1/(r\rho c_1)^2} - 1/(r \rho c_1) =\frac{\sqrt{2 r\rho c_1+1} -1}{r \rho c_1}
.$$
We have
$$ c_0 = \min \{ \frac{\sqrt{2 r\rho c_1+1} -1}{r \rho c_1}, 2/(\hat{\rho} c_1)  
\} = c(c_1).
$$
This finishes the proof.

\subsection{Proof of Theorem \ref{thmmain}} \label{app-main}


Below we prove a more detailed theorem of which Theorem \ref{thmmain} can be regarded as an immediate corollary. 
The discussion after next theorem's statement  analyses the aggregated added noise. Its conclusion is added to Theorem \ref{thmmain}.

\begin{thm} \label{thm-mainp} Let $q_i=\frac{s_{i,c}}{N_c}
= q \cdot (i+m)^p$ with $m> 0$ and $p\geq 0$. We define $s=s_{0,c}/m^p=N_c q$.
Let $T$ be the total number of rounds and let $K$, defined by $K/N_c = \sum_{i=1}^T q_i$, be the total number of grad recursions/computations across all rounds. We define
$$\alpha \geq q\cdot (T+m)^p \mbox{ and } \gamma \geq m/T.$$
We assume 
\begin{equation} r_0=\sigma \alpha \mbox{ with } r_0\leq 1/e, \ u_0 = \frac{2  \sqrt{r_0\sigma}     
}
{
 \sigma -r_0
}<1 \mbox{ and } u_1 = \frac{ 2e \sqrt{r_0 \sigma} 
}
{
 (\sigma-r_0) \sigma
}<1 .\label{condr0}
\end{equation}

Through $\gamma$, values $K$ and $T$ are related to one another by
$$\frac{1}{1+\gamma}( (p+1) \frac{K}{s})^{1/(1+p)} \leq T \leq ( (p+1) \frac{K}{s})^{1/(1+p)}.$$
We define
$$ K^* = \frac{(r_0/\sigma)^{(1+p)/p}}{p+1}  (\frac{N_c}{s})^{1/p} N_c (1+\gamma)^{1+p} .
$$
Value $K^*$ has the property that $\alpha=r_0/\sigma \geq q \cdot (T+m)^p$ is implied by $K\leq K^*$.

We define
\begin{eqnarray*} A(p,r_0,\sigma) &=& (p+1)^{-p/(1+2p)}
(\frac{r(2p+1)^2}{3p+1} \frac{(1+\gamma)^{3(1+2p)}}{(1-\alpha)^2})^{(1+p)/(1+2p)}, \mbox{ and } \\
B(p,r_0,\sigma) &=& 
A(p,r_0,\sigma)
\cdot (1+\gamma)^{-(1+p)(3+4p)/(1+2p)} (\frac{2}{(2r\frac{(p+1)(2p+1)(1+\gamma)^{2p}}{(3p+1)(1-\alpha)^2}+1)^2-1})^{\frac{1+p}{1+2p}},
\end{eqnarray*}
where $r$ is defined based on formula (\ref{constr}) based on values $r_0$, $u_0$ and $u_1$ with their additional dependency on $\sigma$, and $p$. We do not specifically mention the dependency on $\gamma$, but do notice that $\alpha$ is expressed in $r_0$ and $\sigma$.
The larger $r_0$, the larger $r$ and therefore the smaller $B(p,r_0,\sigma)$ and the larger $A(p,r_0,\sigma)$. For $r_0\downarrow 0$, also $r\downarrow 0$ and $B(p,r_0,\sigma)$ increases and converges to a limit. We have an iterative procedure that can be used to compute $r_0(\sigma)$ as a function of $\sigma\geq 1.137$ such that conditions (\ref{condr0}) are satisfied and
\begin{eqnarray*}
B(p,r_0(\sigma),\sigma) &=& \frac{1}{1+p} \cdot (\frac{\sqrt{3}-1}{2} (2p+1))^{(1+p)/(1+2p)}, \mbox{ and } \\
A(p,r_0(\sigma),\sigma) &=& B(p,r_0(\sigma),\sigma)  \cdot (1+\gamma)^{(3+4p)(1+p)/(1+2p)}.
\end{eqnarray*}

Let $\epsilon= \hat{c}_1 q^2 K/s$.

{\bf [Case 1:]} We define
$$K^-=B(p,r_0,\sigma) \epsilon^{(1+p)/(1+2p)} ( \frac{N_c}{ s})^{1/(1+2p)}
N_c.$$
For $K\leq K^-$, 
Alg. \ref{alg:DP} is $(\epsilon, \delta)$-differentially private if
$$\sigma \geq \sqrt{2 \hat{c}_1 }   q \frac{\sqrt{  \frac{K}{s} \ln (1/\delta) }}{\epsilon}
 \frac{(1+\gamma)^{2+3p}}{\sqrt{1-\alpha}}
 = \sqrt{\frac{2\ln (1/\delta)}{\epsilon}}
 \frac{(1+\gamma)^{2+3p}}{\sqrt{1-\alpha}}.$$
 In this case, we can choose
 \begin{eqnarray*}
 \alpha &= & (1+\gamma)^p ((p+1) B_{\alpha=0}(p,r_0,\sigma))^{p/(1+p)} (1-\alpha)^{-2p/(1+2p)} \epsilon^{p/(1+2p)} \frac{s}{N_c} \\
 &=& O(\epsilon^{p/(1+2p)} \frac{s}{N_c}) = O((\frac{\epsilon^{1/(1+2p)}}{m})^p \frac{s_{0,c}}{N_c}), \mbox{ and }\\
 \gamma &= &  m (1+\gamma) (p+1)^{-1/(1+p)} (\frac{s}{K})^{1/(1+p)}
= O(m (\frac{s}{K})^{1/(1+p)}) = O( (\frac{s_{0,c}m}{K})^{1/(1+p)}).
 \end{eqnarray*}
 This shows that, for large $N_c$ and $K$ relative to $s_{0,c}m$, we can neglect $\alpha$ and $\gamma$ terms in our expression
 for the lower bound for $\sigma$.
 
 
 
{\bf  [Case 2:]} We define
$$K^+=A(p,r_0,\sigma) \epsilon^{(1+p)/(1+2p)} ( \frac{N_c}{ s})^{1/(1+2p)}
N_c.$$
 Suppose that
 $$ r\geq \frac{\sqrt{3}-1}{2} \frac{(3p+1)}{(p+1)(2p+1)} \frac{(1-\alpha)^2}{(1+\gamma)^{2p}}.$$
Then, for $K\geq K^+$, 
Alg. \ref{alg:DP} is $(\epsilon, \delta)$-differentially private if
$$\sigma \sqrt{T} \geq 
\sqrt{\frac{K}{s}} \cdot \frac{2\sqrt{2}}{\sqrt{1-\alpha}} \cdot
\frac{p+1}{\sqrt{2p+1}} 
 q 
\frac{\sqrt{ \frac{K}{s} \ln (1/\delta)}}
{\epsilon} 
(1+\gamma)^{(1+2p)/2}.$$
This condition is equivalent to
\begin{equation}
\sigma \geq (K/K^+)^{(1+2p)/(2+2p)}  \cdot \frac{2}{(1-\alpha)^{3/2}} \cdot  \sqrt{ r \frac{(p+1)(2p+1)}{3p+1}} \cdot \sqrt{\frac{2\ln( 1/\delta)}{\epsilon}} (1+\gamma)^{2(1+2p)}.
\label{LBk}
\end{equation}

In case 2, we can choose
\begin{eqnarray*}
\gamma &=&m (1+\gamma) (p+1)^{-1/(1+p)}
A_{\gamma=0,\alpha=0}(p,r_0,\sigma)^{-1/(1+p)} \epsilon^{-1/(1+2p)} ( \frac{s}{N_c})^{2/(1+2p)} \\
&=& O( m ( \frac{s}{\sqrt{\epsilon} N_c})^{2/(1+2p)}
)=O((\frac{s_{0,c} \sqrt{m}}{\sqrt{\epsilon} N_c})^{2/(1+2p)}),
\end{eqnarray*}
which, for large $N_c$ relative to $s_{0,c}\sqrt{m/\epsilon}$, can be neglected in our expression\footnote{Notice that neglecting the $\gamma$ term in $A(p)$ leads to only a difference of $O(1/N_c^{1/(1+2p)})$ epochs in $K^+$.} for 
the lower bound for $\sigma$.


For $r_0(\sigma)$, (\ref{LBk}) reduces to
\begin{equation} \sigma \geq
(K/K^+)^{(1+2p)/(2+2p)}  \cdot \frac{1.21}{\sqrt{1-r_0/\sigma}}   \cdot \sqrt{\frac{2\ln( 1/\delta)}{\epsilon}} (1+\gamma)^{2+3p}. \label{KK+}
\end{equation}
\end{thm}

We will present the proof  of the theorem (as a collection of calculations) in Section \ref{app:prth6}. In Section \ref{app:app} we will use the theorem to calculate parameter settings for different configurations of $\sigma$, $N_c$, $s_{0,c}$, etc.

We remark that the theorem allows a slightly different formulation in the following form: Fix $K$.
Let $p^-$ and $p^+$ with $s^-=s_{0,c}/m^{p^-}$ and $s^+=s_{0,c}/m^{p^+}$   be such that 
\begin{eqnarray*}
K &= &
B(p^-,r_0,\sigma) \epsilon^{(1+p^-)/(1+2p^-)} ( \frac{N_c}{ s^-})^{1/(1+2p^-)}
N_c
\end{eqnarray*}
and
\begin{eqnarray*}
K &= &
A(p^+,r_0,\sigma) \epsilon^{(1+p^+)/(1+2p^+)} ( \frac{N_c }{ s^+})^{1/(1+2p^+)}
N_c.
\end{eqnarray*}
We say $p^-$ realizes $K=K^-$ and $p^+$ realizes $K=K^+$.
Then $p\leq p^-$ corresponds to case 1 and $p\geq p^+$ corresponds to case 2.
We notice that $p^+-p^- = O( 1/\ln N_c)$. 

For $r_0(\sigma)$ and $p\leq 1$, function $A(p,r_0(\sigma),\sigma)$ is a small factor $(1+\gamma)^{(3+4p)(1+p)/(1+2p)}\leq (1+\gamma)^{14/3}$  (where $\gamma=m/T$) larger compared to $B(p,r_0(\sigma),\sigma)$. This means that we have an order $O( \ln(1+m/T) / \ln N_c)$ difference between $p^+$ and $p^-$. For small $m/T$, we have $p^-\approx p^+$.

The theorem shows that as a function of $p\leq p^-$, we have (using the lower bound for $T$ and the lower bound for $\sigma$ in case 1)
$$\sqrt{T}\sigma \geq ((p+1)\frac{K}{s})^{1/(2+2p)}
\frac{1}{\sqrt{1-r_0/\sigma}} \sqrt{\frac{2\ln(1/\delta)}{\epsilon}} (1+\gamma)^{(5+6p)/2}.
$$
At the transition from $p^-$ to $p^+$, this lower bound changes, see (\ref{KK+}) of case 2 with $K=K^+$,
$$\sqrt{T}\sigma \geq ((p+1)\frac{K}{s})^{1/(2+2p)}
\frac{1.21}{\sqrt{1-r_0/\sigma}} \sqrt{\frac{2\ln(1/\delta)}{\epsilon}} (1+\gamma)^{(5+6p)/2}.
$$
From this moment on case 2 shows a lower bound on $\sigma \sqrt{T}$ which only varies with $\frac{p+1}{\sqrt{2p+1}}(1+\gamma)^{(1+2p)/2}$ as a function of $p$ (notice that (\ref{KK+}) is equivalent with this lower bound for $r_0(\sigma)$).
We conclude that for $\gamma=m/T$ small, the smallest possible aggregated noise $\sqrt{T}\sigma$ for which we can prove $(\epsilon,\delta)$-differential privacy varies with $((p+1) \frac{K}{qN_c})^{1/(1+2p)}$ as a function of $p<p^-$, next $\sqrt{T}\sigma$ jumps by a factor $1.21$ when transitioning from $p^-\leq p\leq p^+$ (noting that $p^+-p^- = O(\ln(1+m/T) / \ln N_c) = O(m/(T \ln N_c))$ is small as well), after which $\sqrt{T}\sigma$ varies with $\frac{p+1}{\sqrt{2p+1}}$ as a function of $p$. This observation is added to the theorem in the main body.

\subsubsection{Proof of Theorem \ref{thm-mainp}} \label{app:prth6}

Below we apply Theorem \ref{thm1}:

 \vspace{.5cm}

\noindent
{\bf Discussion $\alpha$ and $\gamma$.}
Quantities $\alpha$ and $\gamma$ satisfy
$$q_i=\frac{s_{i,c}}{N_c} = q\cdot (i+m)^p \leq q\cdot (T+m)^p \leq \alpha \mbox{ and } m/T\leq \gamma.$$
In order to apply Theorem \ref{thm1} we require
\begin{equation} \alpha \leq r_0/\sigma \mbox{ with } r_0\leq 1/e, \ u_0 = \frac{2  \sqrt{r_0\sigma}     
}
{
 \sigma -r_0
}<1 \mbox{ and } u_1 = \frac{ 2e \sqrt{r_0 \sigma} 
}
{
 (\sigma-r_0) \sigma
}<1 . \label{cond}
\end{equation}
Notice that we have these conditions stated as assumptions.


For $\alpha$ we derive
\begin{eqnarray*}
q(T+m)^p &\leq & q T^p (1+\gamma)^p 
\leq  \frac{s}{N_c} ((p+1) \frac{K}{s} )^{p/(1+p)} (1+\gamma)^p.
\end{eqnarray*}
If 
$$ \frac{s}{N_c} ((p+1) \frac{K}{s} )^{p/(1+p)} (1+\gamma)^p \leq \alpha,$$
then $q(T+m)^p\leq \alpha$. 
%
The inequality on $\alpha$ is equivalent to
$$ K
\leq \frac{\alpha^{(1+p)/p}}{p+1}  (\frac{N_c}{s})^{1/p} N_c (1+\gamma)^{1+p} =\frac{(r_0/\sigma)a^{(1+p)/p}}{p+1}  (\frac{N_c}{s})^{1/p} N_c (1+\gamma)^{1+p}= K^*.
$$
This allows $K$ to be $O((N_c/s)^{1/p})$ epochs (each epoch has size $N_c$); this may occur in case 2 where $K=\Omega((N_c/s)^{1/(1+2p)})$ epochs, see the definition for $K^+$ (notice that $1/p\geq 1/(1+2p)$). We conclude that if we choose $r_0$ small enough such that $K\leq K^*$, then condition $q\cdot (T+m)^p\leq \alpha$ is implied.



\vspace{.5cm}

\noindent
{\bf Analysis $\hat{S}_j$.}
We can use $\alpha$ and $\gamma$ to derive the bounds
$$ 1 \leq \frac{N_c}{N_c-s_{i,c}} \leq (1-\alpha)^{-1} \mbox{ and }
1\leq \frac{T+m}{T} = 1+ \gamma.
$$


We have 
 $$ \hat{S}_j = \frac{1}{T} \sum_{i=1}^T q_i^j (\frac{N_c}{N_c-s_{i,c}})^{j-1}\leq 
 \frac{1}{T} \sum_{i=1}^T \frac{q^j (i+m)^{pj}}{(1-\alpha)^{j-1}} \leq \frac{q^j (T+m)^{pj+1}}{T(pj+1)(1-\alpha)^{j-1}}
 \leq 
 \frac{q^j T^{pj}}{pj+1}
 \cdot \frac{(1+\gamma)^{1+pj}}{(1-\alpha)^{j-1}}$$
 and
 $$ \hat{S}_j \geq \frac{1}{T} \sum_{i=1}^T q_i^j \geq 
 \frac{1}{T} \sum_{i=1}^T q^j i^{pj}
\approx
 \frac{q^j T^{pj}}{pj+1}.
 $$
 This can be used to obtain the upper bounds
 \begin{eqnarray*} 
 \frac{\hat{S}_1\hat{S}_3}{\hat{S}_2^2} &\leq &
 \frac{ (2p+1)^2 }{(p+1)(3p+1)} \frac{(1+\gamma)^{2+4p}}{(1-\alpha)^2}= \rho,  \\
 \frac{\hat{S}_1^2}{\hat{S}_2} &\leq & \frac{2p+1}{(p+1)^2} (1+\gamma)^{2+2p} =\hat{\rho} .
 \end{eqnarray*}
 
  \vspace{.5cm}

\noindent
{\bf Dependencies among $K$, $T$, and $\epsilon$.}
 Since $K$ is the total number of grad recursions/computations, we have
 $$ \frac{K}{N_c} = \sum_{i=1}^T q_i = \hat{S}_1 T \leq \frac{q (T(1+\gamma))^{p+1}}{p+1} \mbox{ and }
 \frac{K}{N_c}  \geq \frac{q T^{p+1}}{p+1}.
 $$
 Hence,
 $$ \frac{1}{1+\gamma}( \frac{p+1}{q} \frac{K}{N_c})^{1/(1+p)} \leq T \leq ( \frac{p+1}{q} \frac{K}{N_c})^{1/(1+p)}.$$
  Notice that $N_c q=s_{0,c}/m^p$, which we  denote by $s$. By substituting $q=s/N_c$ we obtain
  \begin{equation} \frac{1}{1+\gamma}( (p+1) \frac{K}{s})^{1/(1+p)} \leq T \leq ( (p+1) \frac{K}{s})^{1/(1+p)}.
  \label{eqT}
  \end{equation}

 For fixed $\epsilon$ we define $c_1$ (as a function of $p$) by
 $$ \epsilon = c_1 T \hat{S}_1^2.$$
 Our bounds on $\hat{S}_1$ and $T$ give
 \begin{equation} \epsilon = c_1 T \hat{S}_1^2 \geq \frac{c_1}{(p+1)^2} T^{1+2p} q^2 
 \geq  
 c_1 (\frac{q}{p+1})^{1-p/(1+p)} (K/N_c)^{1+p/(1+p)} (1+\gamma)^{-(1+2p)} \nonumber
 \end{equation}
 and
 \begin{equation} \epsilon = c_1 T \hat{S}_1^2 \leq 
 c_1 (\frac{q}{p+1})^{1-p/(1+p)} (K/N_c)^{1+p/(1+p)} (1+\gamma)^{2+2p}. \nonumber
 \end{equation}
 We substitute $q=s/N_c$ in the bounds above 
 and reorder terms in order to get bounds on $c_1$:
 \begin{equation}
 c_1
\leq (p+1)^{1-p/(1+p)}  \frac{\epsilon N_c^2}{s^2}
(\frac{s}{K})^{1+p/(1+p)}(1+\gamma)^{1+2p} \label{upc1}
\end{equation}
and
 \begin{equation} c_1
\geq (p+1)^{1-p/(1+p)}  \frac{\epsilon N_c^2}{s^2}
(\frac{s}{K})^{1+p/(1+p)}(1+\gamma)^{-(2+2p)}. \label{lowc1}
\end{equation}
 
 
 

 
 
 
 
 \vspace{.5cm}

\noindent
{\bf Analysis $c(x)$.}
 For the lower bound on $\sigma$ we need
 $c_0=c(c_1)$ as defined in Theorem \ref{thm1}, see also (\ref{constr}). For 
 \begin{eqnarray}
 x&\geq & \frac{(2r\rho/\hat{\rho}+1)^2-1}{2r\rho}= 
 \frac{(2r\frac{(p+1)(2p+1)(1+\gamma)^{2p}}{(3p+1)(1-\alpha)^2}+1)^2-1}{2} \frac{1}{r\rho} \label{eqxtau} \\
 &=& 
 \frac{(2r\frac{(p+1)(2p+1)(1+\gamma)^{2p}}{(3p+1)(1-\alpha)^2}+1)^2-1}{2} \cdot 
\frac{(p+1)(3p+1)(1-\alpha)^2}{r(2p+1)^2(1+\gamma)^{2+4p}}
=
 \tau \nonumber
 \end{eqnarray}
 we have $c(x)= 2/(\hat{\rho} x)$. 
See (\ref{lowc1}), we have $c_1\geq \tau$ if 
 $$
 c_1\geq (p+1)^{1-p/(1+p)}  \frac{\epsilon N_c^2}{s^2}
(\frac{s}{K})^{1+p/(1+p)} (1+\gamma)^{-(2+2p)}\geq \tau,
$$
that is,
\begin{eqnarray}
K &\leq& (p+1)^{1/(1+2p)} ( \frac{\epsilon N_c^2}{\tau s^2})^{(1+p)/(1+2p)}
s (1+\gamma)^{-2(1+p)^2/(1+2p)} \nonumber \\
&=&
(1+\gamma)^{-2(1+p)^2/(1+2p)} (p+1)^{1/(1+2p)} ( \frac{\epsilon}{\tau })^{(1+p)/(1+2p)}
( \frac{N_c}{ s})^{1/(1+2p)}
N_c.  \label{eqKu}
\end{eqnarray}

In a similar way we can analyse the upper bound $c_1\leq 1/(r\rho)$. As one can verify from (\ref{eqxtau}), $1/(r\rho)\leq \tau$ if 
\begin{equation} r\geq \frac{\sqrt{3}-1}{2} \frac{(3p+1)}{(p+1)(2p+1)} \frac{(1-\alpha)^2}{(1+\gamma)^{2p}}
= 0.37 \cdot \frac{(3p+1)}{(p+1)(2p+1)}
\frac{(1-\alpha)^2}{(1+\gamma)^{2p}}. 
\label{ineqr}
\end{equation}
See (\ref{upc1}), inequality $c_1\leq 1/(r\rho)$  is implied by
$$ 
c_1\leq  (p+1)^{1-p/(1+p)}  \frac{\epsilon N_c^2}{s^2}
(\frac{s}{K})^{1+p/(1+p)} (1+\gamma)^{1+2p} \leq
\frac{1}{r\rho} $$
(which implies $c_1\leq \tau$ if (\ref{ineqr}) is true). This condition translates to
\begin{eqnarray}
K &\geq& (p+1)^{1/(1+2p)} ( \frac{\epsilon N_c^2}{(r\rho)^{-1} s^2})^{(1+p)/(1+2p)}
s (1+\gamma)^{1+p} \nonumber \\
&=&
(1+\gamma)^{p+1} (p+1)^{1/(1+2p)} ( \frac{\epsilon}{(r\rho)^{-1} })^{(1+p)/(1+2p)}
( \frac{N_c}{ s})^{1/(1+2p)}
N_c.  \nonumber 
\end{eqnarray}
This lower bound is equivalent to
\begin{eqnarray}
K&\geq &
A(p,r_0,\sigma) \epsilon^{(1+p)/(1+2p)} ( \frac{N_c}{ s})^{1/(1+2p)}
N_c = K^+, \label{eqKA}
\end{eqnarray}
where $A(p,r_0,\sigma)$ is defined as
\begin{eqnarray*}
A(p,r_0,\sigma)&=&(1+\gamma)^{1+p}\frac{(p+1)^{1/(1+2p)}}
{(1/(r\rho))^{(1+p)/(1+2p)}}\\
&=&
(1+\gamma)^{1+p} (p+1)^{-p/(1+2p)}
(\frac{r(2p+1)^2}{3p+1} \frac{(1+\gamma)^{2+4p}}{(1-\alpha)^2})^{(1+p)/(1+2p)} \\
&=& 
(p+1)^{-p/(1+2p)}
(\frac{r(2p+1)^2}{3p+1} \frac{(1+\gamma)^{3(1+2p)}}{(1-\alpha)^2})^{(1+p)/(1+2p)}.
\end{eqnarray*} 

In a similar way we rewrite inequality (\ref{eqKu}) as
\begin{eqnarray}
K&\leq &
B(p,r_0,\sigma) \epsilon^{(1+p)/(1+2p)} ( \frac{N_c}{ s})^{1/(1+2p)}
N_c =K^-, \label{eqKB}
\end{eqnarray}
by defining $B(p,r_0,\sigma)$ as
\begin{eqnarray*}
B(p,r_0,\sigma)&=& (1+\gamma)^{-2(1+p)^2/(1+2p)}
\frac{(p+1)^{1/(1+p)}}
{\tau^{(1+p)/(1+2p)}} \\
&=&
A(p,r_0,\sigma)
\cdot (1+\gamma)^{-(1+p)(3+4p)/(1+2p)} (\frac{2}{(2r\frac{(p+1)(2p+1)(1+\gamma)^{2p}}{(3p+1)(1-\alpha)^2}+1)^2-1})^{\frac{1+p}{1+2p}} \\
&=&
A(p,r_0,\sigma)
\cdot  (\frac{2(1+\gamma)^{-(3+4p)}}{(2r\frac{(p+1)(2p+1)(1+\gamma)^{2p}}{(3p+1)(1-\alpha)^2}+1)^2-1})^{(1+p)/(1+2p)}.
\end{eqnarray*}


Summarizing
$$ [ (\ref{eqKB}) \Rightarrow c_1\geq \tau ] \mbox{ and } [ (\ref{eqKA},\ref{ineqr}) \Rightarrow c_1\leq 1/(r\rho) \leq  \tau ].$$
Furthermore $c_1\geq \tau$ implies $c_0=c(c_1)=2/(\hat{\rho} c_1)$ and $c_1\leq \tau$ implies 
$$c_0=c(c_1)=\frac{\sqrt{2r\rho c_1 +1}-1}{r\rho c_1} .
$$

\vspace{.5cm}

\noindent
{\bf Analysis $A(p,r_0,\sigma)$, $B(p,r_0,\sigma)$, and function $r_0(\sigma)$.}
The larger $r_0$, the larger $r$ and therefore the smaller $B(p,r_0,\sigma)$ and the larger $A(p,r_0,\sigma)$. For $r_0\downarrow 0$, also $r\downarrow 0$, $\alpha \downarrow 0$, and $B(p,r_0,\sigma)$ increases and converges to a limit:
 $$ B(p,r_0\downarrow 0,\sigma) = \frac{(p+1/2)^{(1+p)/(1+2p)}}{p+1}$$
 independent of $\gamma$.
However, this limit cannot be reached as it violates the condition on $\alpha$ because $\alpha$ tends to 0. 

It is possible for $r_0$ to get close to $0$? First, we want $B(p)$ to be close to the maximum limit stated above.
Second, we want no gap between $K^-$ and $K^+$, that is, $K^-\approx K^+$. Third, we want to satisfy (\ref{ineqr}) so that we can use this choice of $r_0$ in both case 1 and case 2. We analyse
$$ r= \frac{\sqrt{3}-1}{2} \frac{(3p+1)}{(p+1)(2p+1)} \frac{(1-r_0/\sigma)^2}{(1+\gamma)^{2p}},
$$
which satisfies (\ref{ineqr})  by its definition.

For this value of $r_0$, we obtain
$$ B(p,r_0,\sigma) = \frac{1}{1+p} \cdot (\frac{\sqrt{3}-1}{2} (2p+1))^{(1+p)/(1+2p)}$$
independent of $\gamma$. This value is equal to the maximum limit times $(\sqrt{3}-1)^{(1+p)/(1+2p)}$ which is in $[0.73,0.81]$ for $0\leq p\leq 1$.

We have
$$A(p,r_0,\sigma)=B(p,r_0,\sigma) \cdot (1+\gamma)^{(3+4p)(1+p)/(1+2p)},$$
which is just slightly larger depending on the value of $\gamma$ for case 2, which is only $O(m(\frac{s}{\sqrt{\epsilon} N_c})^{2/(1+2p)}$. This closes the gap between $K^-$ and $K^+$ tightly.


The chosen value $r$ needs to satisfy its definition, see (\ref{constr}): $r_0\leq 1/e$ with
\begin{eqnarray*}
r &=& r_0 \cdot 2^3 (\frac{1}{1-u_0} + \frac{ 1}{1-u_1} \frac{e^3}{\sigma^3})\exp(3/\sigma^2), \nonumber 
\end{eqnarray*}
where
$$ u_0 = \frac{2  \sqrt{r_0\sigma}     
}
{
 \sigma -r_0
} < 1 \mbox{ and } u_1 = \frac{ 2e \sqrt{r_0 \sigma} 
}
{
 (\sigma-r_0) \sigma
} <1.
$$
The value for $r_0$ can be computed as a function of $\sigma$ in an iterative way starting with $r_0=0$:
$$ r_0^{new}
= \frac{\frac{\sqrt{3}-1}{2} \frac{(3p+1)}{(p+1)(2p+1)} (1-r_0^{old}/\sigma)^2}{2^3 (\frac{ \sigma -r_0^{old}}{ \sigma -r_0^{old}-2  \sqrt{r_0^{old}\sigma}} + \frac{ (\sigma-r_0^{old}) \sigma}{(\sigma-r_0^{old}) \sigma-2e \sqrt{r_0^{old} \sigma} } \frac{e^3}{\sigma^3})\exp(3/\sigma^2)}.$$
Initially, for $r_0=0$, we have $u_0<1$, $u_1<1$, and $r_0\leq (\sqrt{3}-1)/16$.
If $u_0$ and $u_1$ are both $<1$ for $r_0^{old}$, then  $r_0^{new}\leq (\sqrt{3}-1)/16$.  If $r^{old}_0\leq (\sqrt{3}-1)/16$, then $u_0<1$ and $u_1<1$ are implied by 
$$\sigma - (\sqrt{3}-1)/16>  2\sqrt{(\sqrt{3}-1)/16} \sqrt{\sigma} \mbox{ and } (\sigma - (\sqrt{3}-1)/16) \sigma > 2e\sqrt{(\sqrt{3}-1)/16} \sqrt{\sigma}.$$
This is true for $\sigma\geq 1.137$. Then, by induction in the number of iterations, we have for each iterate that $u_0<1$, $u_1<1$, hence, each $r_0>0$, and each $r_0\leq (\sqrt{3}-1)/16<1/e$. This implies that the procedure converges to a value $r_0$ which satisfies all necessary requirements.


For example, for $\sigma=3$ and $p=1$, we obtain $r_0=0.0110$.
For $\sigma=5$ and $p=1$ we obtain $0.0202$.
We denote this function by $r_0(\sigma)$.

 \vspace{.5cm}

\noindent
{\bf Lower bound on $\sigma$ -- Case 1:}
Theorem \ref{thm1} states that $(\epsilon,\delta)$-differentially privacy is implied by the lower bound
 \begin{eqnarray*}
 \sigma &\geq & \frac{2}{\sqrt{c_0}} \frac{\sqrt{ \hat{S}_2 T \ln (1/\delta) }}{\epsilon}
 \end{eqnarray*}
 if condition (\ref{cond}) is satisfied.
 
 We first assume (\ref{eqKB}) implying $c_0=2/(\hat{\rho}c_1)$. By using the upper bounds on $T$, $\hat{S}_2$, and $c_1$ and substituting our definition for $\hat{\rho}$, we derive
 \begin{eqnarray*}
 && \sqrt{2\hat{\rho}} \sqrt{c_1}  \frac{\sqrt{ \hat{S}_2 T \ln (1/\delta) }}{\epsilon} \\
 &\leq & \sqrt{2\hat{\rho}} \sqrt{c_1} \frac{\sqrt{ \frac{q^2 T^{2p}}{2p+1}\frac{(1+\gamma)^{1+2p}}{(1-\alpha)} T \ln (1/\delta) }}{\epsilon} \\
 &\leq &
 \sqrt{2\hat{\rho}} \sqrt{\frac{\epsilon N_c^2}{s K} \frac{(p+1)^{1-p/(1+p)}}{2p+1} } (\frac{K}{s})^{-p/(2+2p)} q \frac{\sqrt{  ((p+1) \frac{K}{s})^{(1+2p)/(1+p)} \ln (1/\delta) }}{\epsilon} \frac{(1+\gamma)^{1+2p}}{\sqrt{1-\alpha}} \\
&=& 
 \sqrt{2\hat{\rho}} \sqrt{\frac{\epsilon N_c^2}{s K}} \frac{p+1}{\sqrt{2p+1}}   q \frac{\sqrt{  \frac{K}{s} \ln (1/\delta) }}{\epsilon} \frac{(1+\gamma)^{1+2p}}{\sqrt{1-\alpha}} \\
 &=& 
 \sqrt{2 \frac{\epsilon N_c^2}{s K} }   q \frac{\sqrt{  \frac{K}{s} \ln (1/\delta) }}{\epsilon}
 \frac{(1+\gamma)^{2+3p}}{\sqrt{1-\alpha}}.
\end{eqnarray*}
We conclude that the lower bound on $\sigma$ is implied by
\begin{equation} \sigma \geq 
\sqrt{2 \frac{\epsilon N_c^2}{s K} }   q \frac{\sqrt{  \frac{K}{s} \ln (1/\delta) }}{\epsilon}
\frac{(1+\gamma)^{2+3p}}{\sqrt{1-\alpha}}. \label{lowcase1}
\end{equation}

We notice that $\alpha$ depends on $T$ and $p$. By using our upper bounds on $T$ and $K$ we derive 
\begin{eqnarray*}
q(T+m)^p &\leq& q T^p (1+\gamma)^p 
\leq  \frac{s}{N_c} ((p+1) \frac{K}{s} )^{p/(1+p)} (1+\gamma)^p \\
&\leq &
(1+\gamma)^p ((p+1) B(p,r_0,\sigma))^{p/(1+p)} \epsilon^{p/(1+2p)} (\frac{N_c}{s})^{-1} \\
&\leq &
(1+\gamma)^p ((p+1) B_{\alpha=0}(p,r_0,\sigma))^{p/(1+p)} (1-\alpha)^{-2p/(1+2p)} \epsilon^{p/(1+2p)} \frac{s}{N_c}
\end{eqnarray*}
(here, $\alpha=0$ in $B$'s definition means that we pull out the $(1-\alpha)^{-2(1+p)/(1+2p)}$ term from $A$'s part).
Hence, if
$$ 
(1+\gamma)^p ((p+1) B_{\alpha=0}(p,r_0,\sigma))^{p/(1+p)}  \epsilon^{p/(1+2p)} \frac{s}{N_c}
\leq \alpha (1-\alpha)^{2p/(1+2p)},
$$
then $q(T+m)^p\leq \alpha$. 

Similarly,
$$ \frac{m}{T}
\leq m (1+\gamma) (p+1)^{-1/(1+p)} (\frac{s}{K})^{1/(1+p)}
$$
and if 
$$ m(p+1)^{-1/(1+p)} (\frac{s}{K})^{1/(1+p)}\leq \gamma/(1+\gamma),$$
then $m/T\leq \gamma$.

The above proves 
that we can choose
$$ \alpha = O(\epsilon^{p/(1+2p)} \frac{s}{N_c}) \mbox{ and } \gamma = O(m (\frac{s}{K})^{1/(1+p)}).$$
This implies
$$
\frac{(1+\gamma)^{2+3p}}{\sqrt{1-\alpha}}= 1 + O(\epsilon^{p/(1+2p)} \frac{s}{N_c}) + O(m (\frac{s}{K})^{1/(1+p)}).
$$


We can interpret (\ref{lowcase1}) in two ways.
The theorem simply defines $\hat{c}_1$ as the solution $\epsilon = \hat{c}_1 q^2 K/s= \hat{c}_1  s K/ N_c^2$.
Substituting this into the above quantity gives
$$
 \sigma \geq  \sqrt{2 \hat{c}_1 }   q \frac{\sqrt{  \frac{K}{s} \ln (1/\delta) }}{\epsilon}
\frac{(1+\gamma)^{2+3p}}{\sqrt{1-\alpha}}.
 $$

In order to give some insight on how $\hat{c}_1$ relates to $c_1$ we observe the following:
Let $c'_1=c_1$ for $p=0$. The bounds on $c_1$ for $p=0$ yield
$$
\frac{\epsilon N_c^2}{s K} (1+\gamma)^{-2} \leq c'_1 \leq \frac{\epsilon N_c^2}{s K} (1+\gamma)
$$
and this proves
$$  c'_1 (1+\gamma)^{-1} \leq \hat{c}_1 \leq c'_1 (1+\gamma)^2,
$$
hence $\hat{c}_1=c'_1 \cdot (1+ O(m (\frac{s}{K})^{1/(1+p)}))$.
 
 The second interpretation is that we can cancel terms by substituting $q=s/N_c$ and this yields
 $$
 \sigma\geq \sqrt{2 \frac{\epsilon N_c^2}{s K} }   q \frac{\sqrt{  \frac{K}{s} \ln (1/\delta) }}{\epsilon}
 \frac{(1+\gamma)^{2+3p}}{\sqrt{1-\alpha}}=
 \sqrt{\frac{2\ln (1/\delta)}{\epsilon}}
 \frac{(1+\gamma)^{2+3p}}{\sqrt{1-\alpha}}.
 $$
 
 This proves the first part of the theorem (case 1) if we can show that the conditions for Theorem \ref{thm1}, that is (\ref{cond}) with $r_0/\sigma = \alpha \geq q\cdot (T+m)^p$, are satisfied. Since case 1 allows $\alpha$ to be small, also $r_0$ can be small which automatically implies $u_0<1$ and $u_1<1$. In our derivations for the lower bound, $r$ of (\ref{constr}) plays no role because it will also be small (implying the higher order term in Lemma \ref{lemthree} being small).
 
We do note that $r_0$ and $r$ play a role in the definition of $K^-$. 
Taking the limit $r_0 \to 0$ implies $r\to 0$ and $\alpha \to  0$, and we have (as we saw before)
 $$ B(p,r_0\downarrow 0,\sigma) = \frac{(p+1/2)^{(1+p)/(1+2p)}}{p+1}$$
 independent of $\gamma$.
However, this limit cannot be reached as it violates the condition on $\alpha$. 
It is possible for $r_0$ to get close to $0$ and we suggest to use $r_0(\sigma)$.

 
 




\vspace{.5cm}

\noindent
{\bf Lower bound on $\sigma$ -- Case 2:}
The second part of the theorem considers the case (\ref{eqKA},\ref{ineqr}).
From  $c_1\leq \tau$ we infer that
$$c_0=c(c_1)=\frac{\sqrt{2r\rho c_1 +1}-1}{r\rho c_1} 
$$
and, since $c_1\leq 1/(r\rho)$ and $x/2\geq \sqrt{1+x}-1 \geq x/2 -x^2/8$ for $x\geq 0$,
$$ 1/2\leq 1-r\rho c_1/2 \leq c_0\leq 1.$$



The lower bound on $\sigma$ is satisfied if 
$$\sigma \sqrt{T} \geq 
\frac{2}{\sqrt{c_0}} T \frac{\sqrt{ \hat{S}_2  \ln (1/\delta) }}{\epsilon}.$$

By using $1/2\leq c_0$ together with upper bounds on $T$ and $\hat{S}_2$, we derive
\begin{eqnarray*}
&& \frac{2}{\sqrt{c_0}} T \frac{\sqrt{ \hat{S}_2  \ln (1/\delta) }}{\epsilon} \\
&\leq&
2\sqrt{2} \cdot T \frac{\sqrt{ \hat{S}_2  \ln (1/\delta) }}{\epsilon} \\
&\leq&
2\sqrt{2} \cdot
\frac{p+1}{\sqrt{2p+1}} 
\frac{K}{s} q 
\frac{\sqrt{ \ln (1/\delta)}}
{\epsilon} 
\frac{(1+\gamma)^{(1+2p)/2}}
{\sqrt{1-\alpha}} \\
&=& \sqrt{\frac{K}{s}} \cdot 2\sqrt{2} \cdot
\frac{p+1}{\sqrt{2p+1}} 
 q 
\frac{\sqrt{ \frac{K}{s} \ln (1/\delta)}}
{\epsilon} 
\frac{(1+\gamma)^{(1+2p)/2}}
{\sqrt{1-\alpha}}.
\end{eqnarray*}
We conclude that the lower bound on $\sigma$ is implied by
\begin{equation} \sigma \sqrt{T} \geq 
\sqrt{\frac{K}{s}} \cdot \frac{2\sqrt{2}}{\sqrt{1-\alpha}} \cdot
\frac{p+1}{\sqrt{2p+1}} 
 q 
\frac{\sqrt{ \frac{K}{s} \ln (1/\delta)}}
{\epsilon} 
(1+\gamma)^{(1+2p)/2}. \label{eqc2}
\end{equation}


We notice that $\alpha$ depends on $T$ and $p$. This time, by using our upper bound on $T$ and lower bound on $K$, we derive 
\begin{eqnarray*}
\frac{m}{T}
&\leq & m (1+\gamma) (p+1)^{-1/(1+p)} (\frac{s}{K})^{1/(1+p)} \\
&\leq&
m (1+\gamma) (p+1)^{-1/(1+p)}
A(p)^{-1/(1+p)} \epsilon^{-1/(1+2p)} ( \frac{s}{N_c})^{2/(1+2p)} \\
&\leq &
m (1+\gamma) (p+1)^{-1/(1+p)}
A_{\gamma=0,\alpha=0}(p)^{-1/(1+p)} \epsilon^{-1/(1+2p)} ( \frac{s}{N_c})^{2/(1+2p)} .
\end{eqnarray*}
If
$$ m  (p+1)^{-1/(1+p)}
A_{\gamma,\alpha=0,0}(p)^{-1/(1+p)} \epsilon^{-1/(1+2p)} ( \frac{s}{N_c})^{2/(1+2p)}\leq \gamma/(1+\gamma),$$
then $m/T\leq \gamma$, so,  we can choose
$$\gamma  = O( m ( \frac{s}{\sqrt{\epsilon} N_c})^{2/(1+2p)}
).$$

The best upper bound for $\alpha$ is the one that leads to $K\leq K^*$ derived earlier. The value of $\alpha$ will not be small as in case 1. We need to assume $r_0$ to be such that $K\leq K^*$.

A second interpretation for lower bound (\ref{eqc2}) is derived by 
$$
K = k\cdot K^+ =  k \cdot A(p) (\epsilon^{1+p} N_c/s)^{1/(1+2p)} N_c
$$
for some $k\geq 1$. 
We have
\begin{eqnarray} \sigma &\geq & \frac{Kq}{s\sqrt{T}} \cdot \frac{2}{\sqrt{1-\alpha}} \cdot \frac{(p+1)}{\sqrt{2p+1}} \frac{\sqrt{2\ln( 1/\delta)}}{\epsilon} (1+\gamma)^{(1+2p)/2} \nonumber \\
&=&
\frac{K}{N_cK^{1/(2+2p)}} (\frac{s}{1+p})^{1/(2+2p)} \cdot
\frac{2}{\sqrt{1-\alpha}} \cdot \frac{(p+1)}{\sqrt{2p+1}} \frac{\sqrt{2\ln( 1/\delta)}}{\epsilon} (1+\gamma)^{(1+2p)/2}  \nonumber \\
&=&
\frac{K^{(1+2p)/(2+2p)}}{N_c} (\frac{s}{1+p})^{1/(2+2p)} \cdot
\frac{2}{\sqrt{1-\alpha}} \cdot
\frac{(p+1)}{\sqrt{2p+1}} \frac{\sqrt{2\ln( 1/\delta)}}{\epsilon} (1+\gamma)^{(1+2p)/2}  \nonumber \\
&=& k^{(1+2p)/(2+2p)} A(p)^{(1+2p)/(2+2p)} 
(\epsilon^{1+p} N_c/s)^{1/(2+2p)} \frac{N_c^{(1+2p)/(2+2p)}}{N_c} \cdot  \nonumber \\
&& \cdot (\frac{s}{1+p})^{1/(2+2p)}
\cdot \frac{2}{\sqrt{1-\alpha}} \cdot
\frac{(p+1)}{\sqrt{2p+1}} \frac{\sqrt{2\ln( 1/\delta)}}{\epsilon} (1+\gamma)^{(1+2p)/2}  \nonumber \\
&=& k^{(1+2p)/(2+2p)} A(p)^{(1+2p)/(2+2p)} (\frac{1}{1+p})^{1/(2+2p)} \cdot \nonumber 
\\
&& \cdot
\frac{2}{\sqrt{1-\alpha}} \cdot
\frac{(p+1)}{\sqrt{2p+1}} \sqrt{\frac{2\ln( 1/\delta)}{\epsilon}} (1+\gamma)^{(1+2p)/2}  \nonumber \\
&=&
k^{(1+2p)/(2+2p)}  \cdot \frac{2}{(1-\alpha)^{3/2}} \cdot  \sqrt{ r \frac{(p+1)(2p+1)}{3p+1}} \cdot \sqrt{\frac{2\ln( 1/\delta)}{\epsilon}} (1+\gamma)^{2(1+2p)}.\label{interpretation}
\end{eqnarray}

Lower bound (\ref{interpretation})  for $\sigma$ seems to even increase in $\sqrt{k}$ for  the constant sample size sequence case $p=0$;  but this is not true because for $p=0$ we have a very large $K^+$ as a function of $p$ and we are always in case 1 
(because  the theorem essentially assumes $K$ does not need to be very large $\geq K^+$ in order to achieve good accuracy).

We suggest to choose $r_0(\sigma)$: This makes the gap between $K^-$ and $K^+$ very small, just $O(\gamma)$.
Also it makes (\ref{interpretation}) as tight as possible: We obtain the lower bound
$$\sigma \geq 
k^{(1+2p)/(2+2p)}  \cdot \frac{2}{ \sqrt{1-r_0/\sigma}} \cdot  \sqrt{\frac{\sqrt{3}-1}{2} } \cdot \sqrt{\frac{2\ln( 1/\delta)}{\epsilon}} (1+\gamma)^{2+3p}.
$$


For example, for $\sigma=3$ and $p=1$, we obtain $r_0=0.0110$ giving the bound
$$\sigma \geq 
k^{(1+2p)/(2+2p)}  \cdot 1.21 \cdot \sqrt{\frac{2\ln( 1/\delta)}{\epsilon}} (1+\gamma)^5.
$$
Since $\sigma=3$, we need to make $\epsilon$ a factor $1.21^2 \cdot k^{(1+2p)/(1+p)}$ larger compared to case 1 with $\sigma=3$. For example, $k=2$ gives a factor $4.14$ implying that $\epsilon=2$ increases to $\epsilon\approx 8$ (still within the range considered in \citep{abadi2016deep}).

For $\sigma=5$ and $p=1$ we obtain $0.0202$ which also gives the bound
$$\sigma \geq 
k^{(1+2p)/(2+2p)}  \cdot 1.21 \cdot \sqrt{\frac{2\ln( 1/\delta)}{\epsilon}} (1+\gamma)^5.
$$
If we want to achieve differential privacy with $3=\sqrt{\frac{2\ln( 1/\delta)}{\epsilon}}$, which leads to $\sigma=3$ in case 1, then
$5\geq k^{(1+2p)/(2+2p)}  \cdot 1.21 \cdot 3$ implying $k\leq 1.53$.

\subsubsection{Parameter selection} \label{app:app}

We  consider $0\leq p\leq 1$, $\delta$ preferably smaller than $1/|N_c|$, and $\epsilon$ values like 2, 4, 8 as used in literature (e.g., \citep{abadi2016deep}); this makes the lower bound on $\sigma$ not 'too' large for the algorithm to converge to a solution with good/proper test accuracy.\footnote{See p. 25 of \citep{dwork2014algorithmic} for a discussion on 'large' $\epsilon$. Merely being able to state an $(\epsilon,\delta)$-differentially privacy guarantee offers an understanding of the existence of worst-case scenarios (certain specifically construed adjacent training data sets $d$ and $d'$ with a set of outputs $S$) which in themselves are unlikely (have a small probability) 
to occur in practice. See p. 18 of \citep{dwork2014algorithmic} for a discussion on the size of $\delta$.}




We use the notation of Theorem \ref{thmmain} and define 
$$B = \frac{1}{1+p} \cdot (\frac{\sqrt{3}-1}{2}(2p+1))^{\frac{1+p}{1+2p}}, \ A = B\cdot (1+\frac{m}{T})^{\frac{(3+4p)(1+p)}{(1+2p)}}, \mbox{ and } D=\frac{(r_0(\sigma)/\sigma)^{\frac{1+p}{p}}}{p+1}(1+\frac{m}{T})^{1+p},
$$
$$ K^-= B \cdot \epsilon^{\frac{1+p}{1+2p}} q^{-\frac{1}{1+2p}} N_c, \ \
K^+ = A \cdot \epsilon^{\frac{1+p}{1+2p}} q^{-\frac{1}{1+2p}} N_c, \ \mbox{ and } K^*=D\cdot q^{-\frac{1}{p}} N_c.
$$
For fixed $p$, we have, for $K\leq K^*$,
\begin{eqnarray*}
\sigma &\geq& \sqrt{\frac{2\ln (1/\delta)}{\epsilon}} \frac{(1+\frac{m}{T})^{2+3p}}{\sqrt{1-r_0(\sigma)/\sigma}} \hspace{3.7cm} \mbox{ if } K\leq K^-, \\
\sigma &\geq& (\frac{K}{K^+})^\frac{1+2p}{2+2p}\cdot 1.21 \cdot  \sqrt{\frac{2\ln (1/\delta)}{\epsilon}}
\frac{(1+\frac{m}{T})^{2+3p}}{\sqrt{1-r_0(\sigma)/\sigma}}\hspace{1cm}
\mbox{ if } K^+\leq K.
\end{eqnarray*}

We assume we know the initial sample size $s_{0,c}$,  data set size $N_c$, value for $p$, and differential privacy budget ${\cal B}= \sqrt{2\ln (1/\delta)}/\sqrt{\epsilon}$ for some $\epsilon$ which we wish to achieve. We estimate a total number of grad computations $K$ needed for an anticipated $\sigma \geq {\cal B}$ based on the lower bound of $\sigma$ for $K\leq K^-$.
We apply Theorem \ref{thmmain} to find concrete parameter settings: We compute $r_0(\sigma)$. Next (case 1 -- aka $K\leq K^-$), we compute $q$ small enough such that $K\leq K^-$ and $K\leq K^*$. Given the initial sample size $s_{0,c}$ and data set size $N_c$ we compute $m$. This yields a formula for $q_i$. We compute $T$ (notice that $K/N_c = \sum_{i=0..T} q_i$) and check the value for $m/T$. Factor $m/T$ indicates how much alike the sample size sequence is to the constant sample size sequence; we will need to correct our calculations based on the calculated $m/T$ and possibly iterate. This leads to a lower bound on $\sigma$ from which we obtain an upper bound on ${\cal B}$; if acceptable, then we are done. If not, then we try a larger $\sigma$, recompute $r_0(\sigma)$, and again start with (case 1) as explained above. At the end we have a parameter setting for which we can compute the overall reduction in communication rounds and reduction in aggregated added DP noise -- this is compared to parameter settings found for other values of $p$ (in particular, $p=0$) that achieve ${\cal B}$. The sample size sequence is computed as $s_{i,c}=\lceil N_c q_i\rceil$. 

After all these investigations, we may also decide to (case 2 -- aka $K\geq K^+$) compute $q$ small enough such that $K\leq K^*$ and $K\leq k\cdot K^+$ for some small enough factor $k>1$. Based on $k$ and ${\cal B}$  we anticipate $\sigma$ based on the lower bound of $\sigma$ for $K\geq K^+$. As in case 1, we use $q$ to compute $m$ and $T$. We again check factor $m/T$, correct calculations, possibly iterate, and finally leading to a lower bound on $\sigma$ which gives an upper bound on ${\cal B}$. We may need to try a larger $\sigma$, recompute $r_0(\sigma)$ and repeat (case 2) until satisfied. Our general theory 
allows other larger settings for $r_0$ (other than $r_0=r_0(\sigma)$), especially in case 2, this may lead to improved results.

\noindent
{\bf Example 1.}
As an example, let
$$ s_{0,c}=16, \ N_c=50,000, \ {\cal B}=\sqrt{\frac{2 \ln (1/\delta)}{\epsilon}}.$$
Here, we assume $\epsilon=2$ as in \citep{abadi2016deep}. We assume $p=1$ and compute 
$$ B= 0.5321797270231777, \ A=0.5321797270231777 \cdot (1+\frac{m}{T})^{14/3}, \ D = 0.5 \cdot (r_0(\sigma)/\sigma)^2 (1+ \frac{m}{T})^2.$$
This yields (using $\epsilon=2$)
$$ K^- =0.8447826585127415 \cdot q^{-1/3} N_c, \
$$
$$
K^+ = 0.8447826585127415 \cdot (1+\frac{m}{T})^{14/3} \cdot q^{-1/3} N_c,
$$
$$
K^* = 
D\cdot q^{-1} N_c.
$$
We may try $\sigma=3$ which gives $r_0(\sigma)=0.0110$.
Suppose we will want $K=100$ epochs. 
For case 1 we compute $q$ such that $K\leq K^-$, that is,
$$ 100 \leq 0.8447826585127415\cdot q^{-1/3},
\mbox{ hence, }  q\leq 0.0000006028856829700258. 
$$
We also check $K\leq K^*$ and find
$$100\leq 0.5 \cdot (0.0110/3)^2 \cdot q^{-1}, \mbox{ hence, } q\leq 0.0000000672222222222222.$$
We set $q$ to meet the latter inequality.
We know that
$$ s_{0,c} = N_c q m^p = N_c q m, $$
Hence, $$m= 16/(50,000 \cdot 0.0000000672222222222222)= 4760.330578512398.$$
We use 
$$ T\approx ((p+1)\frac{K}{N_c q})^{1/(p+1)}, \mbox{ hence, } T\approx 54546.$$
This gives 
$$m/T = 0.08727185455418175.$$
Notice that the value $m/T$ does not affect the upper bound on $q$ as this is limited by $K\leq K^*$ and $K^-$ only increases for larger $m/T$.
The above  leads to the bound
$$ 3 \geq {\cal B} 
\frac{(1+\frac{m}{T})^{2+3p}}{\sqrt{1-r_0(\sigma)/\sigma}}
= {\cal B} \frac{(1+0.08727185455418175)^5}{\sqrt{1-0.0110/3}} = 1.5222584705194226 \cdot {\cal B}, $$
which gives the maximum value
$$ {\cal B} = 3/1.5222584705194226 = 1.9707559905883425.$$

For $\epsilon=2$, this gives 
$$ \delta = 0.020570872204539764.$$
Notice that $N_c\cdot \delta =1029$, which is too large since we want $\delta$ to be smaller than $1/(100*N_c)$. 
In this example, we may increase $\epsilon$ to $\epsilon=8$ (also within the range of typical values, see \citep{abadi2016deep}). We need to redo the calculations: The changed $\epsilon$ only affects $K^-$ (making it larger), which will again lead to a much higher upper bound on $q$
compared to what $K\leq K^*$ gives. This means that the same parameter setting will be calculated and we obtain for $\epsilon=6$,
$$ \delta = 0.0000000634577338911708
$$
and we are fine since $N_c\delta =0.00317288669455854$, less than $1$.

The resulting sample size sequence is
$$ s_{i,c}= N_c q (i+m) =0.0039733713991774566 \cdot (i + 4026.8070594438323) = 16 + 0.00397\cdot i,$$
that is, we round this number  to the nearest larger integer. This means that the $T=50168$ rounds increment the sample size every $1/0.0039733713991774566=252$ rounds.

Notice that compared to a constant sample size sequence with $K/s_{0,c}= 100\cdot 50,000 /16=312500$ and ${\cal B}= 1.98$ for fair comparison, we have a factor $312500/50168=6.23$ reduction in communication rounds and the aggregated added noise is reduced from
$\sqrt{312500}\cdot 1.98=1107$ to $\sqrt{50168}\cdot 3=672$.


\noindent
{\bf Example 2.}
As a second example, let
$$ s_{0,c}=16, \ N_c=10,000, \ {\cal B}=\sqrt{\frac{2 \ln (1/\delta)}{\epsilon}}.$$
Here, we assume $\epsilon=2$ as in \citep{abadi2016deep}. We assume $p=1$ and compute 
$$ B= 0.5321797270231777,$$
$$A=0.5321797270231777 \cdot (1+\frac{m}{T})^{14/3}, $$
$$D = 0.5 \cdot (r_0(\sigma)/\sigma)^2 (1+ \frac{m}{T})^2.$$
This yields (with $\epsilon=2$)
$$ K^- =0.8447826585127415 \cdot q^{-1/3} N_c,$$
$$ K^+ = 0.8447826585127415 \cdot (1+\frac{m}{T})^{14/3} \cdot q^{-1/3} N_c, $$
$$K^* =
D \cdot q^{-1} N_c.
$$
We try $\sigma=5$ which gives $r_0(\sigma)=0.0202$.
Suppose we will want $K=10$ epochs (assuming a 'good' data set). 
We   compute $q$ such that $K\leq K^*$, that is,
$$ 10 \leq 0.5 \cdot (0.0202/5)^2 \cdot q^{-1},$$
Hence, 
$$q\leq 0.0000008160800000000002.$$
We also compute, as in the first example,  
$$10\leq 0.845\cdot q^{-1/3}, $$
Hence, 
$$q\leq 0.0006033511249999998.$$
So, we are again in case 1 now with $q=0.0000008160800000000002$.
We know that
$$ s_{0,c} = N_c q m^p = N_c q m, \mbox{ hence, } m= 16/(10,000 \cdot 0.0000008160800000000002)= 1960.5920988138412.$$
We use 
$$ T\approx ((p+1)\frac{K}{N_c q})^{1/(p+1)}, \mbox{ hence, } T\approx 4951.$$
Notice that the value $m/T$ does not affect the upper bound on $q$ as this is limited by $K\leq K^*$ and $K^-$ only increases for larger $m/T$.
The above 
 gives 
$$m/T = 0.39599921204076777.$$
This leads to the bound
$$ 5 \geq {\cal B} 
\frac{(1+\frac{m}{T})^{2+3p}}{\sqrt{1-r_0(\sigma)/\sigma}}
= {\cal B} \frac{(1+0.40)^5}{\sqrt{1-0.0202/5}} = 5.31257308281 \cdot {\cal B}, $$
which gives a bad bound on ${\cal B}$ and we need to look for a larger $r_0$ so that $K\leq K^*$ will relax as the deciding constraint for $q$ and a larger $q$ can be chosen.

\noindent
{\bf Example 3.}
As a third example, let
$$ s_{0,c}=16, \ N_c=10,000, \ {\cal B}=\sqrt{\frac{2 \ln (1/\delta)}{\epsilon}}.$$
Here, we assume $\epsilon=1$ and $p=1$.


We want to increase $r_0$ so that the constraint $K\leq K^*$ leads to a higher upper bound on $q$. This will help to make the ratio $m/T$ small with as consequence that more privacy budget is available, we see a larger reduction in communication rounds, and  reduced aggregated added noise.

In this example we have $\sigma=8$ and we will use
$r_0=1/e$ such that $u_1=0.15275204077456322$ and $u_0=0.4495546831835495$.
We compute $r=5.7460446671129635$ by using (\ref{constr}).
These values yield
$$ K^*= 0.0010573069002860367 (1+\frac{m}{T})^2\cdot q^{-1} N_c $$
and 
$$
K^- = 0.1368988621622339 \cdot 1^{2/3} \cdot q^{-1/3} 
N_c = 0.1368988621622339 \cdot q^{-1/3} N_c ,$$
where in the calculation for $K^-$ we neglected $\gamma=m/T$ terms and we use $\epsilon=1$; this must be verified later.
Suppose we have K = 2.5 epochs. We compute $q$ such that $K\leq K^*$, that is,
$$ 2.5 \leq 0.0010573069002860367 \cdot q^{-1},
\mbox{ hence, }  q\leq 0.0004229227601144147.
$$
We also compute $q$ such that $K\leq K^-$, as in the first example,  
$$2.5\leq 0.1368988621622339 \cdot q^{-1/3}, \mbox{ hence, } q\leq 0.00016420239582699232.$$
So, we are again in case 1 now with $q=0.00016420239582699232$.
We know that
$$ s_{0,c} = N_c q m^p = N_c q m,$$
Hence,
$$m= 16/(10,000 \cdot 0.00016420239582699232)= 9.744072197861225 .$$
We use 
$$ T\approx ((p+1)\frac{K}{N_c q})^{1/(p+1)}, \mbox{ hence, } T\approx 175.$$
This gives 
$$m/T = 0.055680412559207.$$
This can be used to verify whether our bound on $K^-$ is precise enough: 

We have $K^-=0.12819685818649007\cdot 2^{2/3} \cdot q^{-1/3} 
N_c = 0.12819685818649007 \cdot q^{-1/3} N_c$ and this gives $q\leq (0.12819685818649007/2.5)^3=0.00013483794319482493$ with $m=11.86609616024911$, $T=193$, and $m/T=0.061482363524606794$.


We need to iterate once more:
We have $K^-=0.12734354695486969\cdot 1^{2/3} \cdot q^{-1/3} 
N_c = 0.12734354695486969 \cdot q^{-1/3} N_c$ and this gives $q\leq (0.12734354695486969/2.5)^3=0.00013216327772100012$ with $m=12.106237281566509$, $T=195$, and $m/T=0.06208326811059748$. 

We have converged.

When we repeat this calculation with the new $m/T$, it turns out that $m/T$ has actually converged to its two most significant digits.

This leads to the bound
$$ 8 \geq {\cal B}
\frac{(1+\frac{m}{T})^{2+3p}}{\sqrt{1-r_0(\sigma)/\sigma}}
= {\cal B} \frac{(1+0.06208326811059748)^5}{\sqrt{1-1/(8e)}} = 1.38361486085 \cdot {\cal B}, $$
which gives 
$${\cal B} = 8/1.38361486085 = 5.78195582192962.$$
For $\epsilon=1$, this gives 
$$ \delta = 5.502343985212556\cdot 10^{-8}.$$

In the current setting we have 
$$s_{i,c} = \lceil N_c q (i+m) \rceil = 16 + \lceil 1.3216327772100012\cdot i \rceil. $$

Notice that compared to a constant sample size sequence with $K/s_{0,c}= 2.5\cdot 10,000 /16=1563$ and ${\cal B}= 5.80$ for fair comparison, we have a factor $1563/195=8.02$ reduction in communication rounds and the aggregated added noise is reduced from
$\sqrt{1563}\cdot 5.78=229$ to $\sqrt{193}\cdot 8=112$

\noindent
{\bf Example 4.}
As a fourth example, let
$$ s_{0,c}=16, \ N_c=25,000, \ {\cal B}=\sqrt{\frac{2 \ln (1/\delta)}{\epsilon}}.$$
We assume $\epsilon=2$. 
We assume $p=1$ and compute 
$$ B= 0.5321797270231777, $$
$$ A=0.5321797270231777 \cdot (1+\frac{m}{T})^{14/3}$$
$$ D = 0.5 \cdot (r_0(\sigma)/\sigma)^2 (1+ \frac{m}{T})^2.$$
This yields (with $\epsilon=2$)
$$ K^- =0.8447826585127415 \cdot q^{-1/3} N_c, $$
$$ K^+ = 0.8447826585127415 \cdot (1+\frac{m}{T})^{14/3} \cdot q^{-1/3} N_c, $$
$$ K^* = 
D\cdot q^{-1} N_c.
$$
%
%
We try $\sigma=8$ which gives $r_0(\sigma)=0.0247$.
Suppose we will want $K=5$ epochs (assuming a 'good' data set). 
We   compute $q$ such that $K\leq K^*$, that is,
$$ 5 \leq 0.5 \cdot (0.0247/8)^2 \cdot q^{-1},
\mbox{ hence, }  q\leq 0.000000953265625.
$$
We also compute $q$ such that $K\leq K^-$, as in the first example,  
$$5\leq 0.8447826585127415\cdot q^{-1/3}, \mbox{ hence, } q\leq 0.00482308546.$$
So, we are again in case 1 now with $q=0.000000953265625$.
We know that
$$ s_{0,c} = N_c q m^p = N_c q m,$$
Hence,
$$m= 16/(25,000 \cdot 0.000000953265625)= 671.376354308.$$
We use 
$$ T\approx ((p+1)\frac{K}{N_c q})^{1/(p+1)}, \mbox{ hence, } T\approx 3239.$$
This gives 
$$m/T = 0.20727889913.$$
Notice that the value $m/T$ does not affect the upper bound on $q$ as this is limited by $K\leq K^*$ and $K^-$ only increases for larger $m/T$.
The above leads to the bound
$$ 8 \geq \mathcal{B} 
\frac{(1+\frac{m}{T})^{2+3p}}{\sqrt{1-r_0(\sigma)/\sigma}}
= \mathcal{B} \frac{(1+0.21)^5}{\sqrt{1-0.0247/5}} = 2.57106713 \cdot \mathcal{B}, $$
which gives 
$$\mathcal{B} = 8/2.57106713 = 3.11154847209.$$
For $\epsilon=2$, this gives 
$$ \delta = 0.00006241319203188485.$$
Notice that $N_c\cdot \delta =1.5603298007971211$, which fits somewhat. So, the increasing sample size sequence
$$ s_{i,c} = N_c q (i+m)
= 0.023831640625 \cdot (i+ 671.376354308) = 16 + 0.023831640625\cdot i$$
together with $\sigma=8$ works well in this example.

If we have $\sigma=3$ for constant sample size sequence, then we obtain the same privacy guarantee but now with $5*25000/16=7813$ rounds. This example shows $3239$ rounds giving a reduction of about a factor $2.41$.
The aggregated noise is
$\sqrt{3239}\cdot 8 =455$
versus 
$\sqrt{7813}\cdot 3 =265$ for the constant step size sequence.
This seems to contradict our theory, however, this is because $m/T$ is considered very small in Theorem \ref{thm1} (and we neglect the role of $\gamma$ in the analysis on aggregated added noise).

\noindent
{\bf Example 5.}
We continue the previous example.
We want to increase $r_0$ so that the constraint $K\leq K^*$ leads to a higher upper bound on $q$. This will help to make the ratio $m/T$ small with as consequence that more privacy budget is available, we see a larger reduction in communication rounds, and  reduced aggregated added noise.

In this example we have $\sigma=8$ and we will use
$r_0=1/e$ such that $u_1=0.15275204077456322$ and $u_0=0.4495546831835495$.
We compute $r=5.7460446671129635$ by using (\ref{constr}).
These values yield
$$ K^*= 0.0010573069002860367  (1+\frac{m}{T})^2\cdot q^{-1} N_c
$$
and 
$$
K^- = 0.1368988621622339 \cdot 2^{2/3} \cdot q^{-1/3} 
N_c = 0.21731339780957962 \cdot q^{-1/3} N_c ,$$
where in the calculation for $K^-$ we neglected $\gamma=m/T$ terms and we use $\epsilon=2$; this must be verified later.
We   compute $q$ such that $K\leq K^*$, that is,
$$ 5 \leq 0.0010573069002860367 \cdot q^{-1},
\mbox{ hence, }  q\leq 0.00021146138005720736.
$$
We also compute $q$ such that $K\leq K^-$, as in the first example,  
$$5\leq 0.21731339780957962 \cdot q^{-1/3}, \mbox{ hence, } q\leq 0.00008210119791349613.$$
So, we are again in case 1 now with $q=0.00008210119791349613$.
We know that
$$ s_{0,c} = N_c q m^p = N_c q m, $$
Hence,
$$m= 16/(25,000 \cdot 0.00008210119791349613)= 7.795257758288984 .$$
We use 
$$ T\approx ((p+1)\frac{K}{N_c q})^{1/(p+1)}, \mbox{ hence, } T\approx 349.$$
This gives 
$$m/T = 0.022335982115441216.$$
This can be used to verify whether our bound on $K^-$ is precise enough: 

We have $K^-=0.13329256340146362\cdot 2^{2/3} \cdot q^{-1/3} 
N_c = 0.21158875536302124 \cdot q^{-1/3} N_c$ and this gives $q\leq (0.214/5)^3=0.00007578229244195698$ with $m=8.445244652504904$, $T=364$, and $m/T=0.02320122157281567$. It turns out that we have converged and that $m/T$ remains stable.
%

We need to iterate once more:
We have $K^-=0.133\cdot 2^{2/3} \cdot q^{-1/3} 
N_c = 0.21137208415301048 \cdot q^{-1/3} N_c$ and this gives $q\leq (0.21137208415301048/5)^3=0.00007554972$ with $m=8.471242187728482$, $T=364$, and $m/T=0.023272643372880444$. 
%

We have converged.

This leads to the bound
$$ 8 \geq {\cal B}
\frac{(1+\frac{m}{T})^{2+3p}}{\sqrt{1-r_0(\sigma)/\sigma}}
= {\cal B} \frac{(1+0.023272643372880444)^5}{\sqrt{1-1/(8e)}} = 1.1486274762710156 \cdot {\cal B}, $$
which gives 
$${\cal B} = 8/1.1486274762710156 = 6.964834261123335   .$$
For $\epsilon=2$, this gives 
$$ \delta = 8.56732681988984\cdot 10^{-22},$$
an overkill. 
We may get away with a smaller $\epsilon$, but this leads to a smaller $q$ and we need to repeat the above calculations.
Another option is to move to case 2 as explained at the start of this subsection.

In the current setting we have 
$$s_{i,c} = \lceil N_c q (i+m) \rceil = 16 + \lceil 1.8887430727901684\cdot i \rceil. $$
Notice that the constant sample size sequence has $5\cdot 25000/16=7813$ rounds implying a reduction factor $365/7813=0.047$, a significant improvement. The aggregated added noise is reduced from $\sqrt{7813}\cdot 6.96=615$ to $\sqrt{365}\cdot 8=153$.

\section{Experiments}
\label{app:experiment}

In this section, we provide  experiments to support our theoretical findings, i.e., the convergence of our proposed asynchronous federated learning (FL) with strong convex, plain convex and non-convex objective functions with Stochastic Gradient Descent (SGD) at the client sides. Moreover, we also show the convergence of our FL with Gaussian differential privacy  for strong convex, plain convex and non-convex objective functions. 

We introduce the settings and parameters used in the experiments in Section~\ref{subsec:hyperparameter}. Section~\ref{subsec:exp_asynFL} provides  detailed experiments for asynchronous FL with different types of objective functions (i.e., strong convex, plain convex and non-convex objective functions), different types of step size schemes (i.e., constant and diminishing step size schemes), different types of sample size sequences (i.e., constant and increasing sample sizes) and different types of data sets at the clients' sides (i.e., unbiased and biased). Section~\ref{subsec:exp_asynFL_DP} provides  detailed experiments for asynchronous FL with different Gaussian differential privacy schemes with different types of objective functions (i.e., strong convex, plain convex and non-convex objective functions), different types of step size schemes (i.e., constant and diminishing step size schemes), different types of sample size sequences (i.e., constant and increasing sample sizes) and different types of data sets at the clients' sides (i.e., unbiased and biased). All our experiments are conducted mainly on LIBSVM\footnote{https://www.csie.ntu.edu.tw/~cjlin/libsvmtools/datasets/binary.html} and MNIST data sets.

\subsection{Experiment settings}
\label{subsec:hyperparameter}

Equation~$\ref{eq_logstic_reg}$ defines the plain convex logistic regression problem. The weight $w$ of the logistic function can be learned by minimizing the log-likelihood function $J$:
\begin{equation} \label{eq_logstic_reg}
    J = - \sum_{i}^N [ y_{i} \cdot \log (\sigma_i) + (1 - y_{i}) \cdot \log (1 - \sigma_i) ], \text{ (plain convex)}
\end{equation}
where $N$ is the number of training samples $(x_i,y_i\in\{0,1\})$ and
\begin{equation} \nonumber 
    \sigma(w, x, b) = \frac{1}{1 + e^{-(w^{\mathrm{T}}x + b)}}
\end{equation}
is the sigmoid function with as parameters the weight vector $w$ and bias value $b$.
Function $J$ can be changed into a strong convex problem by adding a regularization parameter $\lambda>0$:
\begin{equation} \nonumber 
    \hat{J} = - \sum_{i}^N [ y_{i} \cdot \log (\sigma_i) + (1 - y_{i}) \cdot \log (1 - \sigma_i) ] + \frac{\lambda}{2}\norm{w}^{2}, \text{ (strongly convex).}
\end{equation}
For simulating non-convex problems, 
we choose a simple neural network (Letnet) for image classification.

The parameters used for our FL algorithms with strongly convex, plain convex and non-convex objective functions are described in Table~\ref{tbl:tbl_async_fl_parameter}. Table~\ref{tbl:tbl_async_fl_parameter_diff_privacy} provides the parameters used for our FL with Gaussian differential privacy. 

\begin{table}[H]
\caption{Asynchronous FL training parameters}
\label{tbl:tbl_async_fl_parameter}
\vskip 0.1in
\begin{center}
\begin{small}
\scalebox{0.9}{
\begin{tabular}{|l|c|c|c|c|c|}
\hline
                 & Sample size sequence & \# of clients & Diminishing step size & Regular $\lambda$         \\ \hline
Strongly convex  & $s_i = a \cdot i^c + b$             & 5                 & $\frac{\eta_0}{1 + \beta {t}}$                & $\frac{1}{N}$                  \\ \hline
Plain convex  & $s_i = a \cdot i^c + b$             & 5                 & $\frac{\eta_0}{1 + \beta {t}}$ or $\frac{\eta_0}{1 + \beta \sqrt{t}}$ &  $N/A$          \\ \hline
Non-convex    & $s_i = a \cdot i^c + b$             & 4                 & $\frac{\eta_0}{1 + \beta \sqrt{t}}$ &  $N/A$ \\ \hline
\end{tabular}
}
\end{small}
\end{center}
\vskip -0.1in
\end{table}


\begin{table}[ht!]
\caption{Asynchronous FL training parameters with differential privacy}
\label{tbl:tbl_async_fl_parameter_diff_privacy}
\vskip 0.1in
\begin{center}
\begin{small}
\scalebox{0.9}{
\begin{tabular}{|l|c|c|c|c|c|c|}
\hline
               & Sample size sequence & \# of clients & Diminishing step size & Regular $\lambda$  & Grad norm $C$  \\ \hline
Strongly convex & $s_i = s \cdot (i+m)^p$ & 5 & $\frac{\eta_0}{1 + \beta {t}}$ & $\frac{1}{N}$ & 0.1                   \\ \hline

Plain convex    & $s_i = s \cdot (i+m)^p$ & 5 & $\frac{\eta_0}{1 + \beta {t}}$ or $\frac{\eta_0}{1 + \beta \sqrt{t}}$ & $N/A$ & 0.1                                       \\ \hline

Non-convex      & $s_i = s \cdot (i+m)^p$ & 4 & $\frac{\eta_0}{1 + \beta \sqrt{t}}$ & $N/A$ & 0.1     \\ \hline
\end{tabular}
}
\end{small}
\end{center}
\vskip -0.1in
\end{table}

For the plain convex case, we can use the diminishing step size schemes $\frac{\eta_0}{1 + \beta \cdot t}$ or $\frac{\eta_0}{1 + \beta \cdot \sqrt{t}}$. In this paper, we focus our experiments for the plain convex case on $\frac{\eta_0}{1 + \beta \cdot \sqrt{t}}$. Here, $\eta_0$ is the initial step size and we perform a systematic grid search on parameter $\beta=0.001$ for strongly convex case and $\beta=0.01$ for both plain convex and non-convex cases.

\subsection{Asynchronous federated learning}
\label{subsec:exp_asynFL}

We consider the asynchronous FL with strong convex, plain convex and non-convex objective functions for different settings, i.e., different step sizes schemes (constant and diminishing step sizes), different sample sizes (constant and increasing samples sizes), different type of the data sets (biased and unbiased data sets).

\subsubsection{Asynchronous FL with constant step size scheme and constant sample sizes}
\label{subsec:constant_stepsize_samplesize_FL_simu}

The purpose of this experiment is to choose which is the best constant sample size we can use. This allows us to compare constant step size and constant sample size schemes with diminishing step size schemes that use increasing sample size sequences.


For simplicity, we set the total number of iterations at $20000$  for 5 clients, and a constant step size $\eta=0.0025$.

\begin{table}[ht!]
\caption{The accuracy of asynchronous FL with constant sample sizes}
\label{tbl_async_fl_strongly_convex_constant_sample_size}
\vskip 0.1in

\begin{center}
\begin{footnotesize}
\scalebox{0.9}{
\begin{tabular}{|c|l|cccccc|}
\hline
& Sample size          & 50      & 100     & 200     & 500     & 700     & 1000    \\
& \# of Communication rounds & 80      & 40      & 20      & 8       & 6       & 4       \\
\hline \hline
{a9a}  & strongly convex             & 0.8418 & \textbf{0.8443} & \textbf{0.8386} & 0.8333 & 0.8298 & 0.7271 \\

{} & plain convex             & 0.8409 & \textbf{0.8415} & \textbf{0.8417} & 0.8299 & 0.8346 & 0.7276 \\

\hline \hline

{covtype.binary}  & strongly convex            & 0.8429 & \textbf{0.8402} & \textbf{0.8408} & 0.8360 & 0.8278 & 0.7827 \\

{}  & plain convex               & 0.8421 & \textbf{0.8438} & \textbf{0.8404} & 0.8381 & 0.8379 & 0.7287 \\

\hline \hline

{mnist}  & non-convex               & 0.9450 & \textbf{0.9470} & \textbf{0.9180} & 0.8740 & 0.8700 & 0.7080 \\

\hline
\end{tabular}
}
\end{footnotesize}
\end{center}
\vskip -0.1in
\end{table}

The results are in Table~$\ref{tbl_async_fl_strongly_convex_constant_sample_size}$. We see that with  constant sample size $s_i=100$ or $s_i=200$, we get the best accuracy for strong convex, plain convex and non-convex cases, when compared to the other constant sample sizes. We notice that we can choose the small constant sample size $s_i=50$ as well, because this  also achieves good accuracy, however, this constant sample size requires the algorithm to run for (much) more communication rounds. In conclusion, we need to choose a suitable constant sample size, for example $s_i=100$ or $s_i=200$ to get a good accuracy and a decent number of communication rounds.

\subsubsection{Asynchronous FL with diminishing step size schemes and increasing sample sizes }
\label{subsec:samplesize_FL_simu}

To study the behaviour of asynchronous FL with different increasing sample size sequences, we conduct the following experiments for both $O(i)$ and $O(\frac{i}{\ln i})$ sequences. The purpose of this experiment is to show that these two sampling methods provide good accuracy.

\begin{figure}[ht!]
\begin{center}
\includegraphics[width=0.65\textwidth]{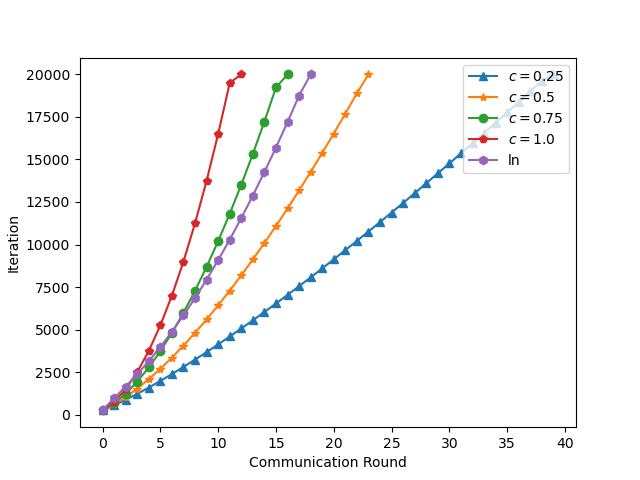}
\end{center}
\caption{Number of communication rounds by sampling methods.}
\label{fig:async_fl_sampling_method}
\end{figure}

\begin{figure}[ht!]
  \centering
  \subfloat[Strong convex.]{\includegraphics[width=0.5\textwidth]{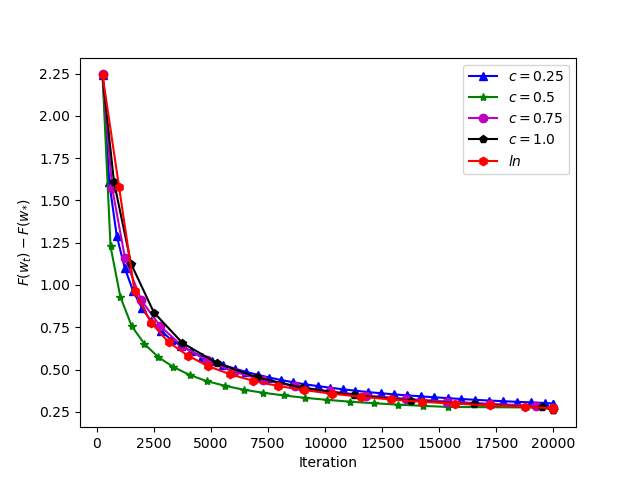}\label{fig:async_fl_a9a_sampling_method_1}}
  \hfill
  \subfloat[Plain convex.]{\includegraphics[width=0.5\textwidth]{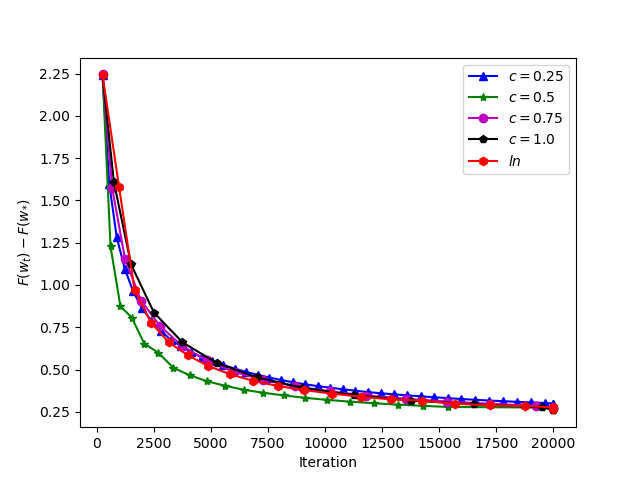}\label{fig:async_fl_plain_convex_a9a_sampling_method_2}}
  \caption{Effect of sampling methods (a9a dataset).}
  \label{fig:async_fl_a9a_sampling_method}
\end{figure}

\begin{figure}[ht!]
  \centering
  \subfloat[Strong convex.]{\includegraphics[width=0.5\textwidth]{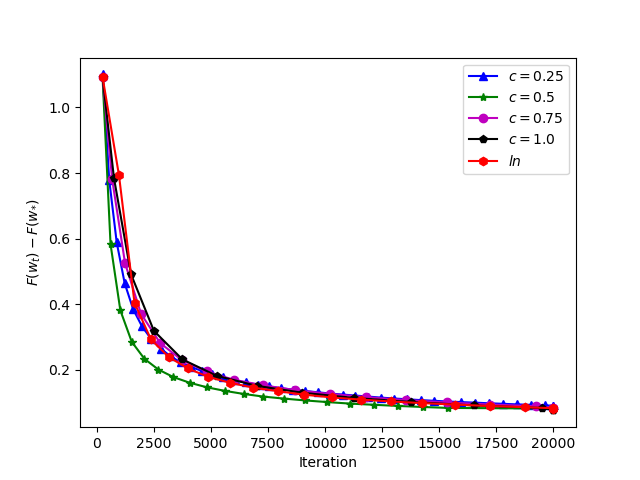}\label{fig:async_fl_covtype_binary_sampling_method_1}}
  \hfill
  \subfloat[Plain convex.]{\includegraphics[width=0.5\textwidth]{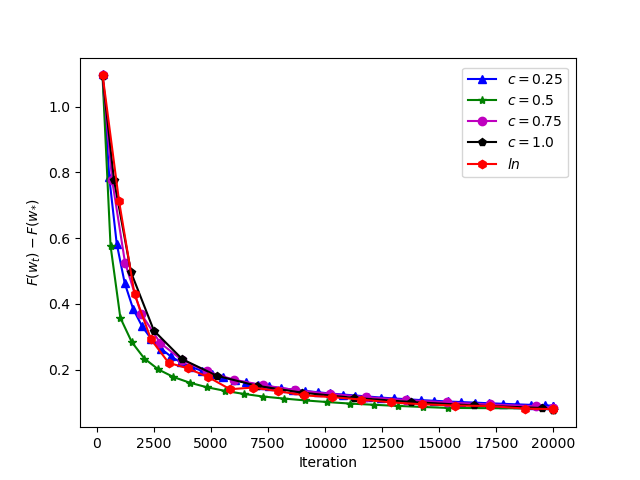}\label{fig:async_fl_plain_convex_covtype_binary_sampling_method_2}}
  \caption{Effect of sampling methods (covtype-binary dataset).}
  \label{fig:async_fl_covtype_binary_sampling_method}
\end{figure}

\textbf{Sampling method:} In this paper, we choose two ways to increase the sample sizes from one communication round to the next. Let  $s_i$ be the number of iterations that the collection of all clients will run in round $i$:
\begin{enumerate}
    \item{${O}(i)$ method: $s_i = a \cdot i^c + b$ where $c \in [0, 1]$, $a,b \geq 0$. 
    }
    
     \item{${O}(\frac{i}{\ln i})$ method: $s_i = a \cdot \frac{i}{\ln(i)} + b$ where $a, b \geq 0$.}
\end{enumerate}

For simplicity, we set $b=0, a=50$, and the total number of iterations at $20000$ for 5 clients in Figure~$\ref{fig:async_fl_sampling_method}$. We choose the diminishing step size sequence $\frac{\eta_0}{1 + \beta \cdot t}$ for the strong convex case and $\frac{\eta_0}{1 + \beta \cdot \sqrt{t}}$ for the plain convex case, with initial step size $\eta_{0} = 0.01$. From Figure~$\ref{fig:async_fl_a9a_sampling_method}$ and Figure~$\ref{fig:async_fl_covtype_binary_sampling_method}$ we infer that $O(i)$ with $c=1.0$ and $O(\frac{i}{\ln{i}})$ sample size sequences provide  fewer communication rounds while maintaining a good accuracy, when compared to other settings. (Other increasing sample size sequences, such as exponential increase and cubic increase, are not  good choices for our asynchronous FL setting.)

\subsubsection{Comparison  asynchronous FL with (constant step size, constant sample size) and (diminishing step size, increasing sample size).}
\label{subsec:stepsize_FL_simu}

In order to understand the behavior of the FL framework for different types of step sizes, we conduct the following experiments on strong convex, plain convex and non-convex problems.

\textbf{Diminishing step size scheme:} We use the following two diminishing step size schemes:
\begin{enumerate}
    \item{Diminishing step size scheme over iterations (diminishing$_1$): Each client $c$ uses $\eta_t$ for $t = N\cdot (\sum_{j=0}^{i-1} s_{j,c} + h)$ where $i \geq 0$ denotes the current round, and $h \in \{0,\ldots,  s_{i,c}-1\}$ is the current iteration which the client executes. In our experiments all clients use the same sample sizes $s_{j,c}=s_j/N$. Hence, $t = \sum_{j=0}^{i-1} s_{j} + N\cdot h$.}
    
     \item{Diminishing step size scheme over rounds (diminishing$_2$): Each client $c$ uses a round step size $\bar{\eta}_i$ for all iterations in round $i$. The round step size $\bar{\eta}_i$ is equal to $\eta_t$ for  $t = \sum_{j=0}^{i-1} s_{j}$.
     }
     
\end{enumerate}

\begin{figure}[ht!]
  \centering
  \subfloat[Strong convex.]{\includegraphics[width=0.5\textwidth]{Experiments/Stepsize/strong_convex_phishing_convergence_rate_4.png}\label{fig:async_fl_strongly_convex_phishing_convergence}}
  \hfill
  \subfloat[Plain convex.]{\includegraphics[width=0.5\textwidth]{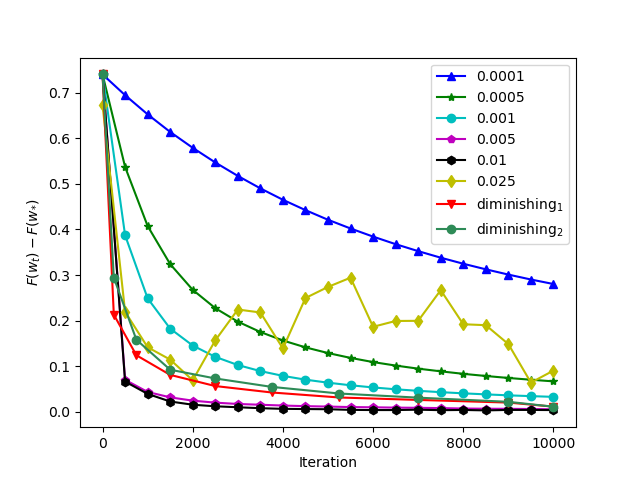}\label{fig:async_fl_plain_convex_phishing_convergence}}
  \caption{Convergence rate with different step sizes (phishing dataset)}
  \label{fig:async_fl_phishing_convergence}
\end{figure}

\begin{figure}[ht!]
  \centering
  \subfloat[Strong convex.]{\includegraphics[width=0.5\textwidth]{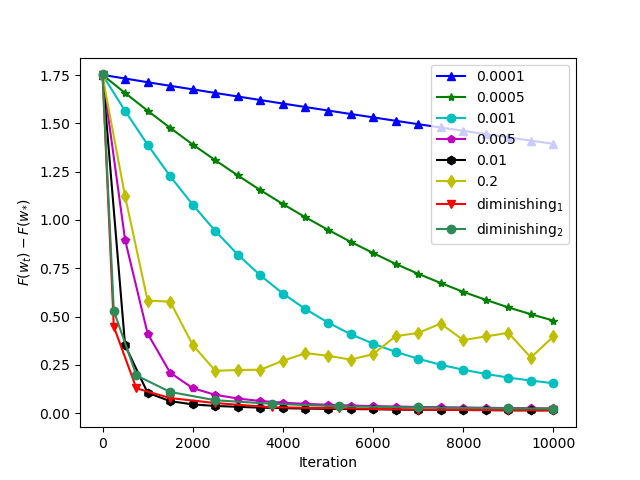}\label{fig:async_fl_strongly_convex_ijcnn1_convergence}}
  \hfill
  \subfloat[Plain convex.]{\includegraphics[width=0.5\textwidth]{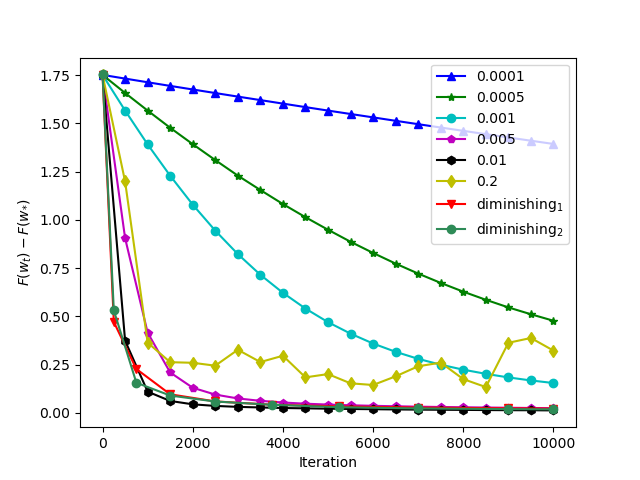}\label{fig:async_fl_plain_convex_ijcnn1_convergence}}
  \caption{Convergence rate with different step sizes (ijcnn1 dataset)}
  \label{fig:async_fl_ijcnn1_convergence}
\end{figure}

\begin{figure}[ht!]
\begin{center}
\includegraphics[width=0.65\textwidth]{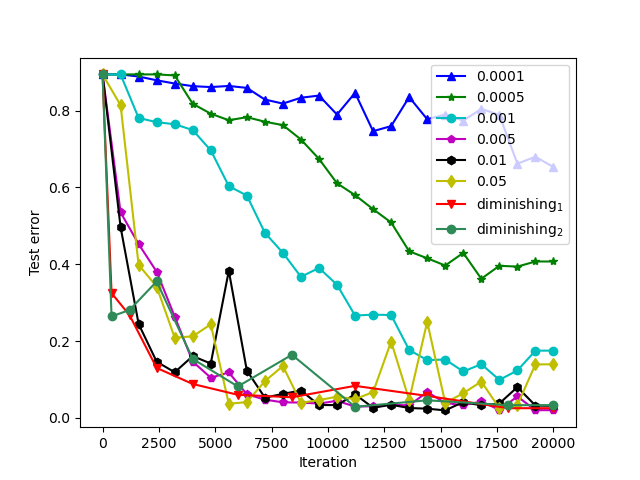}
\end{center}
\caption{Asynchronous FL in non-convex (MNIST dataset)}
\label{fig:async_fl_mnist_neural_net}
\end{figure}

\begin{figure}[ht!]
  \centering
  \subfloat[Strongly convex (real-sim dataset).]{\includegraphics[width=0.5\textwidth]{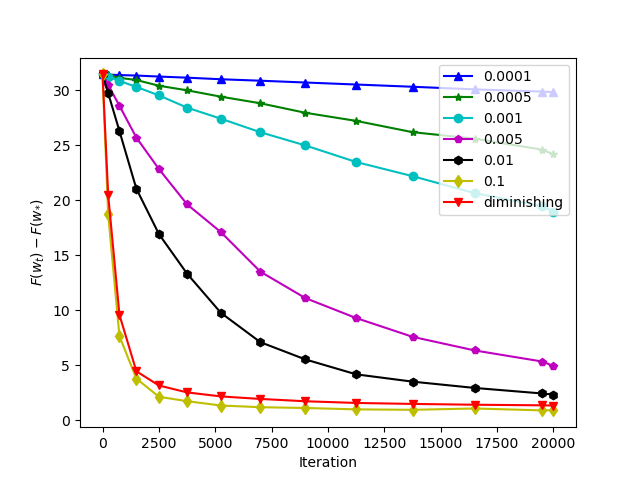}\label{fig:async_fl_stepsize_method_decay_vs_constant_real_sim}}
  \hfill
  \subfloat[Strongly convex (w8a dataset).]{\includegraphics[width=0.5\textwidth]{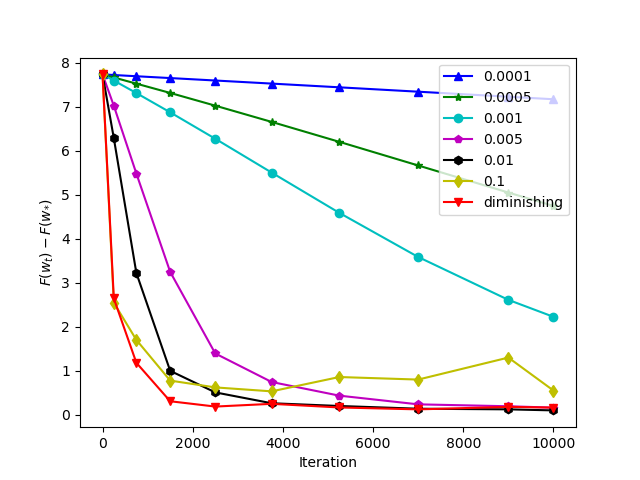}\label{fig:async_fl_stepsize_method_decay_vs_constant_w8a}}
  \caption{Asynchronous FL (linearly increase sampling) with constant and diminishing step sizes.}
  \label{fig:async_fl_stepsize_method_decay_vs_constant}
\end{figure}

The detailed setup of this experiment is described in Table~\ref{tbl:tbl_async_fl_parameter}. In terms of diminishing step size, the experiment will start with the initial step size $\eta_{0}=0.1$ and linear increasing sample size sequence over communication rounds.

\textbf{Strongly convex case:}
Figure~\ref{fig:async_fl_strongly_convex_phishing_convergence} and Figure~\ref{fig:async_fl_strongly_convex_ijcnn1_convergence}  show that our proposed asynchronous FL with diminishing step sizes and increasing sample size sequence achieves the  same or better accuracy when compared to FL with constant step sizes and constant sample sizes.
The figures depict  strong convex problems with diminishing step size scheme $\eta_t=\frac{\eta_0}{1 + \beta \cdot t}$ for an initial step size $\eta_{0}=0.1$ with a linearly increasing sample size sequence $s_i=a \cdot i^c + b$, where $c=1$ and $a, b \geq 0$ 
; diminishing$_1$ uses a more fine tuned  $\eta_t$ locally at the clients and diminishing$_2$ uses the transformation to round step sizes $\bar{\eta}_i$.
The number of communication rounds, see Figure~$\ref{fig:async_fl_phishing_convergence}$ as example, for constant step and sample sizes is $20$ rounds, while  the diminishing step size with increasing sample size setting only needs  $9$ communication rounds.


\textbf{Plain convex case:} We repeat the above experiments for plain convex problems. Figure~$\ref{fig:async_fl_plain_convex_phishing_convergence}$ and Figure~$\ref{fig:async_fl_plain_convex_ijcnn1_convergence}$  illustrate the same results for the diminishing step size scheme $\eta_{t}=\frac{\eta_0}{1 + \beta \cdot \sqrt{t}}, \eta_0=0.1$ with increase sample size sequence  $s_i=a \cdot i^c + b$, where $c=1$ and $a, b \geq 0$.

\textbf{Non-convex case:} We run the experiment with the MNIST data set using the LeNet-$5$ model. We choose a diminishing step size scheme by heuristically decreasing the step size by $\eta_{t}=\frac{\eta_0}{1 + \beta \cdot \sqrt{t}}, \eta_0=0.1$ and a heuristic sample sequence $s_i=a \cdot i^c + b$, where $c=1$ and $a, b \geq 0$. The detailed result of this experiment is illustrated in Figure~$\ref{fig:async_fl_mnist_neural_net}$.


In addition, we extend the experiments for constant step size, but linearly increase the sample size $s_i=a \cdot i^c + b$, where $c=1$ and $a, b \geq 0$ over communication rounds for strongly convex case.
The difference between constant step size scheme$-$increasing sample size sequence and diminishing step size scheme$-$increasing sample size sequence can be found in Figure~\ref{fig:async_fl_stepsize_method_decay_vs_constant}. Overall, our asynchronous FL framework with diminishing step sizes, $\eta_0=0.1$ gains good accuracy, which can only be achieved by fine tuning constant step sizes $\eta=0.01$ and $\eta=0.005$ for the two data sets. 

In conclusion, our proposed asynchronous FL with diminishing step size schemes and increasing sample size sequences  works effectively for strong convex, plain convex and non-convex problems because it can achieve the best accuracy when compared to other constant step size schemes while requiring fewer communication rounds.

\subsubsection{Asynchronous FL with biased and unbiased data sets}
\label{subsec:dataset_FL_simu}

To study the behaviour of our proposed framework towards  biased and unbiased data set, we run a simple experiment with in total 10000 iterations for 2 clients, and each client has their own data set.  The goal of this experiment is to find out whether our FL framework can work well with biased data sets. Specifically, the first client will run for a data set which contains only digit 0 while the second client runs for a data set with only digit 1. Moreover, we choose the initial step size $\eta_{0}=0.01$ and a linearly increasing sample size sequence $s_i=a \cdot i^c + b$, where $c=1$ and $a, b \geq 0$ for strongly convex and plain convex problems. For simplicity, we choose the diminishing step size scheme $\frac{\eta_0}{1 + \beta \cdot t}$ for the strong convex case and $\frac{\eta_0}{1 + \beta \cdot \sqrt{t}}$ for the plain convex case over communication rounds.

\begin{figure}[ht!]
  \centering
  \subfloat[Strong convex.]{\includegraphics[width=0.5\textwidth]{Experiments/BiasDataset/biased_strongly_convex_fashion_mnist_1_digit01.png}\label{fig:asyn_fl_biased_strongly_convex_mnist}}
  \hfill
  \subfloat[Plain convex.]{\includegraphics[width=0.5\textwidth]{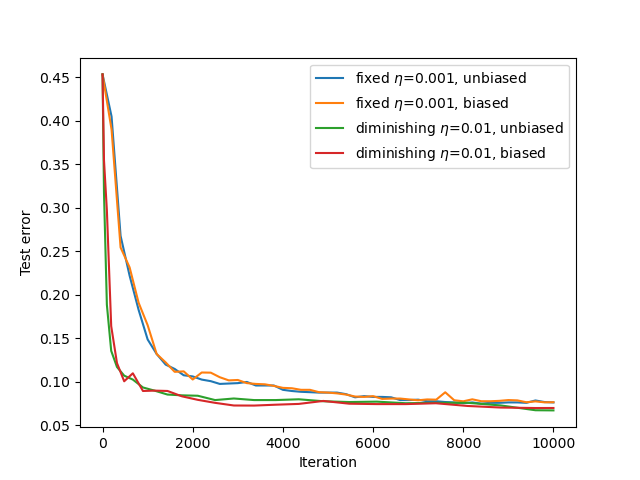}\label{fig:asyn_fl_biased_plain_convex_mnist}}
  \caption{Asynchronous FL with biased and unbiased dataset (MNIST subsets)}
  \label{fig:asyn_fl_biased_mnist}
\end{figure}

\begin{figure}[H]
\begin{center}
\includegraphics[width=0.65\textwidth]{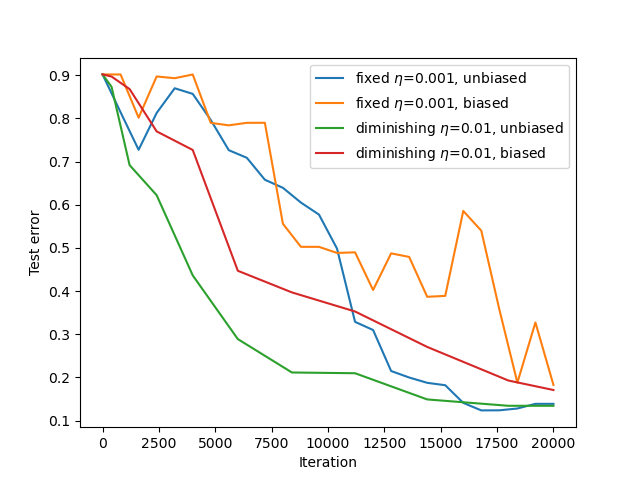}
\end{center}
\caption{Asynchronous FL with biased and unbiased dataset (MNIST dataset)}
\label{fig:async_fl_biased_non_convex_mnist}
\end{figure}

As can be seen from Figure~$\ref{fig:asyn_fl_biased_mnist}$, generally, there is no significant difference when the clients run for biased or unbiased data sets. It means that our proposed FL framework can tolerate the issue of biased data set, which is common in reality. Turning to the non-convex problem, we extend this experiment for the MNIST dataset, but the local data set in each client is separately biased, i.e, each client just has a separate subset of MNIST digits. The experiment uses the initial step size $\eta_{0}=0.01$ with diminishing step size scheme $\frac{\eta_0}{1 + \beta \cdot \sqrt{t}}$ over the communication rounds. Figure~$\ref{fig:async_fl_biased_non_convex_mnist}$ shows that while the accuracy might fluctuate during the training process, our asynchronous FL still achieves good accuracy in general.

In conclusion, our asynchronous FL framework can work well under biased data sets, i.e, this framework can tolerate the effect of biased data set, which is quite common in reality.

\subsection{Asynchronous FL with differential privacy mechanisms}
\label{subsec:exp_asynFL_DP}

To study the behaviour of our proposed FL framework when applying differential privacy (DP) to protect data from client sides, we run the experiments with constant step size-constant sample size vs. diminishing step size-linearly increasing sample size with and without DP. The goal of this experiment is that our asynchronous FL framework can achieve a good accuracy and needs fewer communication rounds between the clients and the server. 



\begin{figure}[H]
  \centering
  \subfloat[Strongly convex, $K=25000, \sigma=8.0$]{\includegraphics[width=0.5\textwidth]{Experiments/DP/strongly_convex_dp_w8a_2_case1_new.png}\label{fig:tbl_async_fl_strong_convex_dp_w8a_1}}
  \hfill
  \subfloat[Plain convex, $K=25000, \sigma=8.0$]{\includegraphics[width=0.5\textwidth]{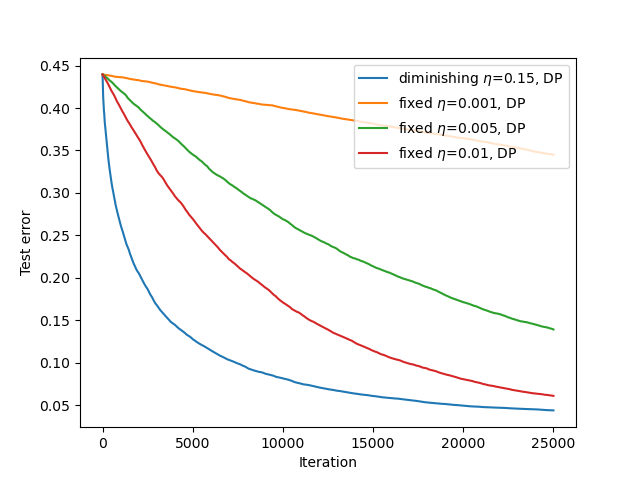}\label{fig:tbl_async_fl_plain_convex_dp_w8a_2}}
  \caption{Differential private with asynchronous FL (w8a dataset)}
  \label{fig:tbl_async_fl_convex_dp_w8a}
\end{figure}

\begin{figure}[ht!]
\begin{center}
\includegraphics[width=0.65\textwidth]{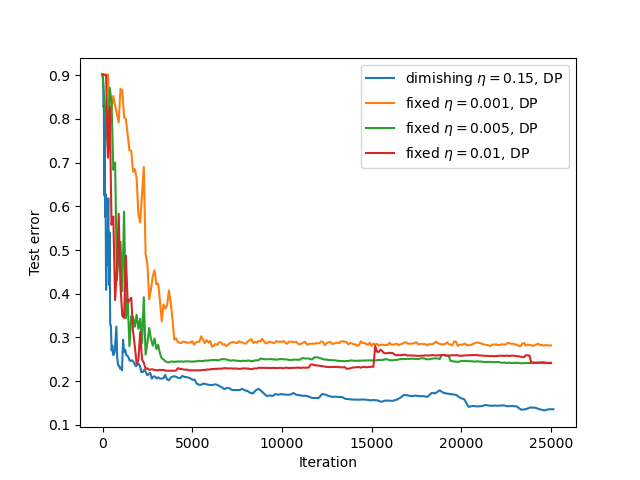}
\end{center}
\caption{Differential private with asynchronous FL (non convex, $K=25000,\sigma=8.0$, MNIST dataset)}
\label{fig:tbl_async_fl_non_convex_dp_mnist}
\end{figure}

Our experiments with DP are conducted on strong convex, plain convex and non-convex problems, where we choose increasing sample size sequences $s_{i,c} = N_c q (i+m)^p$, where 
$m \geq 0$ and $p \in [0,1]$ according to Example 3 of Supplemental Material \ref{app:app} with suitable differential privacy ($\epsilon, \delta$) = ($1.0$, $5.5 \cdot 10^{-8}$), $\sigma=8.0$. In this experiment, we set the data set size to $N_c=10000$ records for each client, $q = 0.000132, K=25000, m=12.106$ and $p=1.0$, so the increasing sample sequences become $s_{i,c} = 1.322 \cdot i + 16$
. Moreover, we have the
diminishing step size scheme $\frac{\eta_0}{1 + \beta \cdot t}$ for strong convex case and $\frac{\eta_0}{1 + \beta \cdot \sqrt{t}}$ for both plain convex and non-convex cases, with initial step size $\eta_0=0.15$. 
Figure~$\ref{fig:tbl_async_fl_convex_dp_w8a}$ shows that our FL framework 
gains  good accuracy when compared to other well-known constant step sizes like $\eta=0.01, \eta=0.001$ and constant sample sizes $s=16$ 
under differential privacy setting. We also extend the experiment to the non-convex case in Figure~$\ref{fig:tbl_async_fl_non_convex_dp_mnist}$, which shows that our FL framework works well within the DP setting.

In conclusion, this experiment shows that our asynchronous FL with diminishing step size scheme and increasing sample size sequence works well under DP, i.e, our asynchronous FL can gain differential privacy guarantees while maintaining acceptable accuracy. It improves over other parameter settings which use constant step sizes.

\end{document}